%% file: main.tex
\documentclass[10pt,twocolumn,letterpaper]{article}

\usepackage[pagenumbers]{wacv} 

\definecolor{wacvblue}{rgb}{0.21,0.49,0.74}
\usepackage[pagebackref,breaklinks,colorlinks,allcolors=wacvblue]{hyperref}

\input{packages}

\def\wacvPaperID{683} 
\def\confName{WACV}
\def\confYear{2027}

\newcommand{\schemename}{\textit{ForAug}\xspace}

\title{\schemename: Mitigating Biases in Image Classification via Controlled Image Compositions}

\author{Tobias Christian Nauen${}^{1,2}$, Brian B. Moser${}^2$, Federico Raue${}^2$, Stansislav Frolov${}^2$, Andreas Dengel${}^{1, 2}$ \\
${}^1$ RPTU University Kaiserslautern-Landau, Kaiserslautern, Germany \\
${}^2$ German Research Center for Artificial Intelligence (DFKI), Kaiserslautern, Germany \\
{\tt\small first\_second.last@dfki.de / first.last@dfki.de}
}

\begin{document}
\maketitle
\input{sec/abstract.tex}

\input{sec/intro.tex}

\input{sec/related_work.tex}

\input{sec/method.tex}

\input{sec/experiments.tex}

\input{sec/conclusion.tex}

\input{sec/acks.tex}
{
	\small
	\bibliographystyle{ieeenat_fullname}
	\bibliography{main_bib}
}

\newpage
\onecolumn
\appendix

\input{sec/appendix}

\end{document}

%% file: packages.tex
\usepackage{color}

\usepackage{amssymb}
\usepackage{amsfonts}
\usepackage{amsmath}
\usepackage[capitalize,noabbrev]{cleveref}
\usepackage{amsxtra}
\usepackage{cancel}
\usepackage{dsfont}
\usepackage{graphicx}
\usepackage{tikz}
\usepackage{tikz-qtree}
\usetikzlibrary{shapes}
\usetikzlibrary{positioning}
\usetikzlibrary{trees}
\usepackage{mathcomp}
\usepackage{mathtools}
\usepackage{multirow}
\usepackage{verbatim}
\usepackage{polynom}
\usepackage{textcomp}
\usepackage{float}
\usepackage{placeins}
\usepackage{xcolor}
\usepackage{pdflscape}
\usepackage{csquotes}
\usepackage{afterpage}
\usepackage{makecell}
\usepackage{listings}
\usepackage{url}
\usepackage{enumitem}
\usepackage{minibox}
\usepackage{algorithm}
\usepackage{algorithmic}
\usepackage{shuffle}
\usepackage{svg}
\usepackage{pifont}
\usepackage{subcaption}
\usepackage{xspace}
\usepackage{siunitx}
\usepackage{booktabs}
\usepackage{microtype}
\usepackage{footmisc}
\usepackage[export]{adjustbox}

\DeclareMathSymbol{\mlq}{\mathord}{operators}{``}
\DeclareMathSymbol{\mrq}{\mathord}{operators}{`'}

\newcommand{\N}{\mathbb N}

\renewcommand{\P}{\mathbb{P}}

\newcommand{\Z}{\mathbb{Z}}

\newcommand{\eps}{\varepsilon}

\renewcommand{\phi}{\varphi}

\newcommand{\abs}[1]{\left| #1 \right|}

\newcommand{\gtxt}[1]{\text{\textcolor{gray}{#1}}}
\definecolor{DarkGreen}{RGB}{34,149,34}
\newcommand{\grntxt}[1]{\text{\textcolor{ForestGreen}{#1}}}
\newcommand{\rdtxt}[1]{\text{\textcolor{red}{#1}}}
\newcommand{\code}[1]{\texttt{#1}}
\newcommand{\cmark}{\ding{51}}%
\newcommand{\xmark}{\ding{55}}%
\DeclareMathOperator{\mean}{mean}

%% file: sec/abstract.tex
\begin{abstract}
	Large-scale image classification datasets exhibit strong compositional biases: objects tend to be centered, appear at characteristic scales, and co-occur with class-specific context.
	By exploiting such biases, models attain high in-distribution accuracy but remain fragile under distribution shifts.
	To address this issue, we introduce \schemename, a controlled composition augmentation scheme that factorizes each training image into a \emph{foreground object} and a \emph{background} and recombines them to explicitly manipulate object position, object scale, and background identity.
	\schemename uses off-the-shelf segmentation and inpainting models to (i) extract the foreground and synthesize a neutral background, and (ii) paste the foreground onto diverse neutral backgrounds before applying standard strong augmentation policies.
	Compared to conventional augmentations and content-mixing methods, our factorization provides direct control knobs that break foreground-background correlations. %
	Across 10 architectures, \schemename improves ImageNet top-1 accuracy by up to 6 percentage points (p.p.) and yields gains of up to 7.3\,p.p.\ on fine-grained downstream datasets.
	Moreover, the same control knobs enable targeted diagnostic tests: we quantify background reliance, foreground focus, center bias, and size bias via controlled background swaps and position/scale sweeps, and show that training with \schemename substantially reduces these shortcut behaviors and significantly increases accuracy on standard distribution-shift benchmarks by up to $19$\,p.p.
	Our code and dataset are publicly available at \url{https://github.com/tobna/ForAug}.
\end{abstract}

%% file: sec/intro.tex
\section{Introduction}
\label{sec:intro}

\begin{figure}
	\centering
	\includegraphics[width=.9\columnwidth]{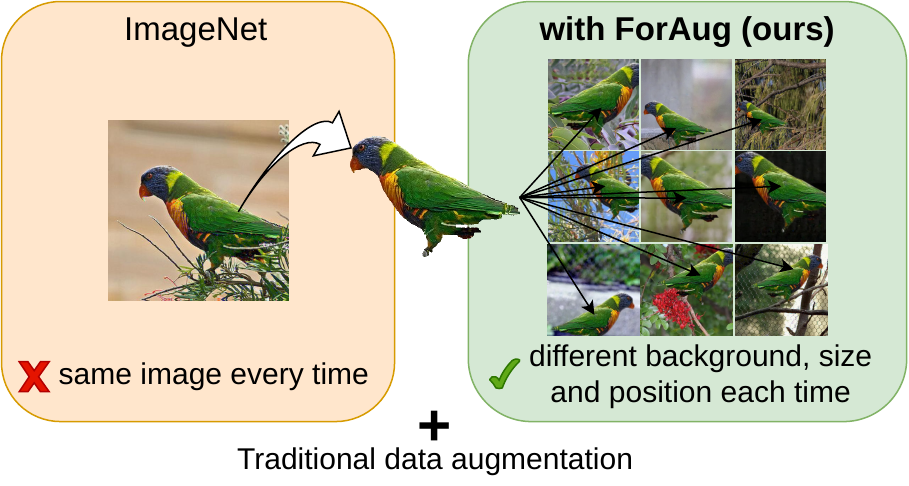}
	\caption{\schemename factorizes each training image into a foreground object and a background, then recombines them on the fly while controlling background identity, object position, and object scale.}
	\label{fig:foraug-example}
\end{figure}

Large-scale image classification is a central driver of modern computer vision: it benchmarks progress in computer vision~\cite{Khan2022,Rangel2024}, powers pretraining~\cite{Dosovitskiy2021,Liu2021,Touvron2021b}, and yields representations that transfer broadly and underpin applications like medical diagnosis~\cite{Sanderson2022,Vezakis2024}, autonomous driving~\cite{Wang2023a}, and object recognition~\cite{Carion2020,He2017,Girshick2014}.
However, classification supervision is weak in an important sense: the label does not specify \emph{how} the class-object should appear.
In ImageNet~\cite{Deng2009} for example, objects often occur at characteristic positions and scales and co-occur with correlated scene context~\cite{Fatima2025,Barbu2019}.
As a result, models rely on shortcuts like background cues, center bias, or size bias, that boost in-distribution accuracy but hurt robustness and transfer learning~\cite{Geirhos2020,Fatima2025,Barbu2019}.

Here, data augmentation is the default defense.
Standard transformations (crop/flip/color jitter) and stronger policies, such as, MixUp~\cite{Zhang2018a}/CutMix~\cite{Yun2019} and automated augmentation search~\cite{Cubuk2019,Cubuk2020} expand appearance diversity~\cite{Shorten2019,Xu2023d} %
However, they largely preserve spatial distributions and image compositions and do not offer control of object position and scale.
This constraint matters especially for Vision Transformers (ViTs)~\cite{Dosovitskiy2021}: with weaker built-in spatial inductive biases than Convolutional Neural Networks (CNNs), ViTs must learn key equivariances (e.g., translation and scale robustness) primarily from data.
Copy-paste style augmentations~\cite{Ghiasi2021,Kang2022} alter composition more aggressively by overlaying segmented objects onto other images.
These are typically designed for detection or instance segmentation and rely on dense human annotations available for these tasks~\cite{Ghiasi2021,Werman2022,Ling2022} or use unconstrained dataset images as backgrounds~\cite{Kang2022}.
As a result, they do not explicitly enforce that the pasted background is semantically neutral, creating ambiguous labels for classification.

To encode compositional invariances directly in the training data, we propose the \emph{Foreground-Background Augmentation} (\schemename), a controlled composition augmentation that \emph{explicitly factorizes each image into foreground and background, then recombines them for label-preserving, interpretable distribution shifts}.
Concretely, \schemename uses off-the-shelf segmentation and inpainting models to (i) extract a foreground and synthesize a class-consistent, semantically neutral background, and (ii) paste the foreground onto diverse neutral backgrounds while controlling its position and scale (see \Cref{fig:foraug-example}).
Unlike prior copy-paste methods that simply overlay objects onto arbitrary scenes~\cite{Ghiasi2021,Kang2022,Werman2022}, \schemename first removes and neutralizes the original background, then samples from well-defined distributions of backgrounds, object positions, and sizes.
This explicit factorization preserves a clean label for the recombined image while providing direct control over compositions, enabling us to break spurious correlations while still fitting seamlessly into modern strong augmentation pipelines. %
To ensure all gains are complementary to strong augmentation pipelines (RandAugment, MixUp, CutMix), we apply \schemename on top of these.

Empirically, \schemename yields consistent accuracy gains across architectures, improving ImageNet top-1 accuracy by up to 6\,p.p. and fine-grained downstream accuracy by up to 7.3\,p.p., and even improving transfer when ImageNet accuracy is matched.
Beyond accuracy, training with \schemename substantially improves robustness on standard distribution-shift benchmarks, where we observe gains of roughly $2-19$\,p.p. across ViT, Swin, and ResNet architectures.

Finally, \schemename{}'s control knobs enable a targeted diagnosis of shortcut reliance and robustness.
We quantify background reliance via controlled background swaps, and probe center and size biases through systematic position and scale shifts, showing that training with \schemename reduces biases.

\medskip
\noindent
\textbf{Contributions}
\begin{itemize}[topsep=0pt]
	\item \textbf{Controlled composition augmentation for classification.}
	      We introduce \schemename, a foreground-background factorization and recombination scheme for image classification that creates label-preserving training samples with explicit control over background identity, position, and scale.
	\item \textbf{Accuracy and transfer gains.}
	      Training with \schemename, in addition to standard strong augmentation pipelines, improves ImageNet top-1 accuracy by up to 6\,p.p., boosts fine-grained downstream accuracy by up to 7.3\,p.p. and increases accuracy on shifted distributions by up to $19$\,p.p.
	\item \textbf{Controlled bias diagnostics and mitigation.}
	      Using the same controls during evaluation, we measure background reliance, and position/scale biases through targeted distribution shifts.
	      \schemename systematically reduces these shortcut behaviors and reliance on dataset biases.
\end{itemize}

%% file: sec/related_work.tex
\section{Related Work}
\label{sec:related_work}

\textbf{Data Augmentation for Image Classification.}
Data augmentation is a crucial technique for improving model performance and generalization.
Traditional augmentation strategies rely on simple geometric or color-space transformations like cropping, flipping, rotation, blurring, color jittering, or random erasing~\cite{Zhong2020} to increase training data diversity without changing the semantic meaning.
With the advent of ViTs~\cite{Dosovitskiy2021}, new data augmentation operations like PatchDropout~\cite{Liu2022d} have been proposed.
Other transformations like MixUp~\cite{Zhang2018a}, CutMix~\cite{Yun2019}, or random cropping and patching~\cite{Takahashi2018} combine multiple input images.
These simple transformations are usually bundled to form more complex augmentation policies like AutoAugment~\cite{Cubuk2019}, RandAugment~\cite{Cubuk2020}, or 3-Augment~\cite{Touvron2022}. %
For a general overview of data augmentation for image classification, we refer to Shorten et al.~\cite{Shorten2019} and Xu et al.~\cite{Xu2023d}.

We advance these general augmentations by introducing \schemename to explicitly separate objects and backgrounds for image classification, allowing us to move beyond image compositions from the dataset.
Thus, \schemename unlocks performance improvements and bias reduction not possible with traditional data augmentation.

\textbf{Copy-Paste Augmentation.}
The copy-paste augmentation~\cite{Ghiasi2021}, used only for object detection~\cite{Shermaine2025,Ghiasi2021} and instance segmentation~\cite{Werman2022,Ling2022}, since ground truth object positions are available for these tasks, involves copying segmented objects from one image and pasting them onto another.
While typically human annotated segmentation masks are required to extract the foreground objects, other foreground sources have been explored, like 3D models~\cite{Hinterstoisser2019} and pretrained object-detection models for use on objects with white background~\cite{Dwibedi2017} or synthetic images~\cite{Ge2023}.
Kang et al.~\cite{Kang2022} apply copy-paste as an alternative to CutMix in classification, but they do not shift the size or position of the foregrounds and use dataset images (with object) as backgrounds.

Unlike prior copy-paste methods that overlay objects, \schemename extracts foregrounds and replaces their backgrounds with semantically neutral fills, thereby preserving label integrity while enabling controlled and diverse recombination.

\textbf{Generative data augmentation.}
Recent work uses generative models to synthesize additional training images, e.g., via GANs or diffusion models driven by text prompts or attribute labels~\cite{Lu2022,Trabucco2024,Islam2024}.
Concurrently to our work, AGA~\cite{Rahat2025} combines LLMs, diffusion models, and segmentation to generate fully synthetic backgrounds from text prompts, onto which real foregrounds are pasted.
These synthetic images are appended to the original training set.
While AGA focuses on increasing diversity via prompt-driven background synthesis,
fully synthetic, prompt-generated backgrounds are likely to change the effective background distribution, especially when generators are biased~\cite{Zverev2025,Shumailov2024,Adamkiewicz2026}.

\schemename uses generative models differently:
We apply inpainting to \emph{locally} neutralize the original object region, yielding semisynthetic backgrounds that preserve the global layout, style, and characteristics of real dataset images.
We then do online recombination of real foregrounds with these neutralized, dataset-consistent backgrounds under explicit control of object position and scale.
\schemename acts as a dynamic large-scale augmentation method while AGA is statically expanding small-scale training sets with synthetic data.
Thus, we outperform AGA while being more than $40\times$ faster at preprocessing (see Appendix A).

\textbf{Model robustness evaluation.}
Evaluating model robustness to various image variations is critical for understanding and improving model generalization.
Datasets like ImageNet-A~\cite{Hendrycks2021}, ImageNet-C~\cite{Hendrycks2019} and ImageNet-P~\cite{Hendrycks2019} introduce common corruptions and perturbations.
ImageNet-E~\cite{Li2023e} evaluates model robustness against a collection of distribution shifts.
Other datasets, such as ImageNet-D~\cite{Zhang2024f} and ImageNet-R~\cite{Hendrycks2021a}, focus on varying background, texture, and material, but rely on synthetic data.
Stylized ImageNet~\cite{Geirhos2019} investigates the impact of texture changes.
ImageNet-9~\cite{Xiao2021} explores background variations using segmented images for a 9-class subset of ImageNet with artificial backgrounds.

In contrast to these existing datasets, which are used only for evaluation, \schemename provides fine-grained control over foreground object placement, size, and background selection, enabling a precise and comprehensive analysis of specific model biases within the context of a large-scale, real-world image distribution.
As \schemename also provides controllable training data generation, it goes beyond simply measuring robustness to actively improving it during training.

%% file: sec/method.tex
\begin{figure*}[t]
	\centering
	\includegraphics[width=.85\textwidth]{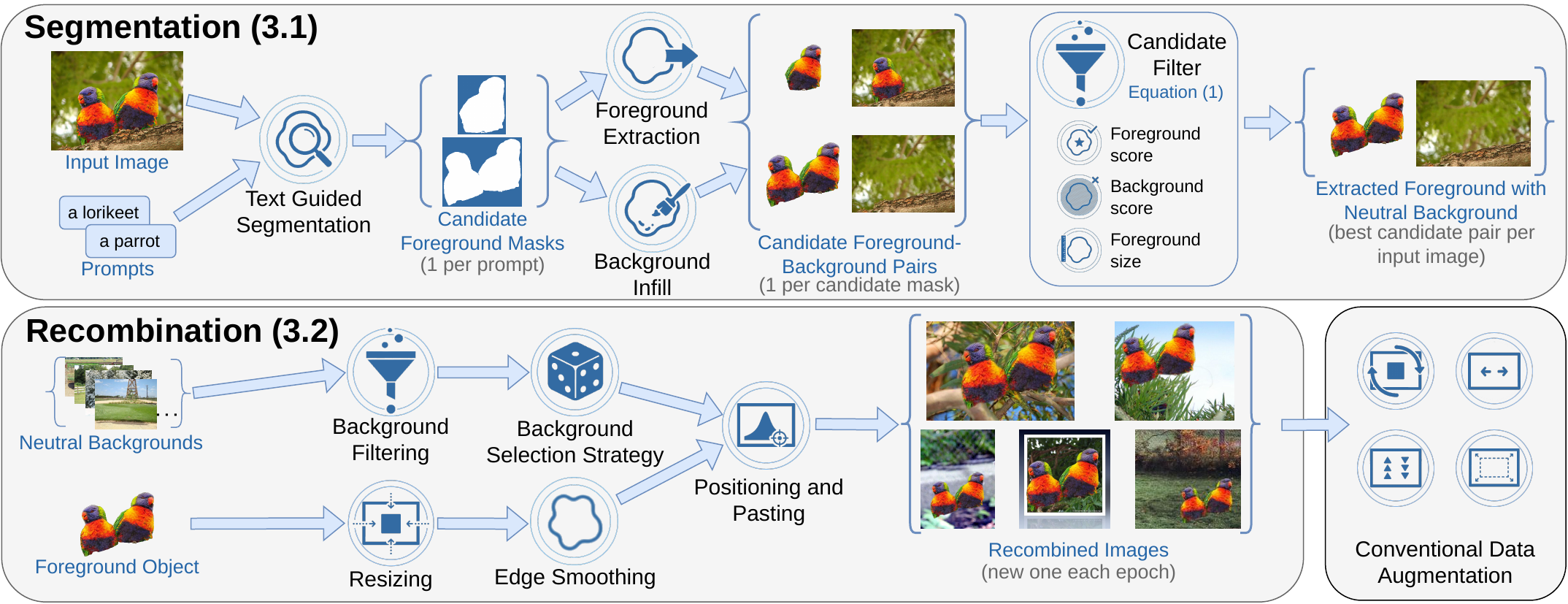}
	\caption{\textbf{Overview of \schemename.}
		We segment the foreground object and inpaint the removed region to obtain a neutral background (offline, \Cref{sec:segmentation}).
		We then paste the foreground onto a sampled background while controlling position and scale, then apply standard strong traditional augmentations (online, \Cref{sec:recombination}).}
	\label{fig:method}
\end{figure*}

\section{\schemename}
\label{sec:method}

We introduce \schemename, a data augmentation designed to enhance training by embedding spatial invariances, which ViTs would otherwise need to learn implicitly, directly into the training data.
\schemename comprises two distinct stages: Segmentation and Recombination. Both are illustrated in \Cref{fig:method}.

\subsection{Segmentation}
\label{sec:segmentation}
The offline segmentation stage produces reusable assets for recombination:
For each labeled training image, we create a pair $(\mathrm{fg},\mathrm{bg})$ consisting of (\textit{i}) a foreground cut-out $\mathrm{fg}$ with an alpha mask and (\textit{ii}) an inpainted background image $\mathrm{bg}$ where the foreground region has been removed.
This stage is computed once offline and the results are stored for the recombination stage.

\textbf{Generate candidate foreground masks.}
We obtain foreground candidates with Grounded SAM~\cite{Ren2024} (Grounding DINO~\cite{Liu2024a} + SAM~\cite{Kirillov2023}).
We leverage the dataset label by prompting the model with ``\code{a <class name>, a type of <object category>}''.
Here \code{<object category>} is the immediate WordNet hypernym of the class (e.g., ``sorrel'' $\rightarrow$ ``horse''), which improves robustness when the class name is rare or overly specific.
This can be the case with prompts like ``sorrel'' or ``guenon'', where the more general name ``horse'' or ``monkey'' is more ubiquitous.
To increase recall, we generate up to $N=3$ masks per image by iteratively moving one level up the hypernym chain (e.g., ``sorrel'' $\rightarrow$ ``horse'' $\rightarrow$ ``equine'' $\dots$).
We merge near-duplicate masks with pairwise IoU $\ge 0.9$, yielding a small set of $n_i<N$ candidate masks per image $i$.
We select the best mask per image (according to \Cref{eq:filtering-score}) in a later filtering step, described below.

\textbf{Create neutral backgrounds.}
Given a candidate mask, we remove the masked region and inpaint it using an object-removal model (LaMa~\cite{Suvorov2022} or Attentive Eraser~\cite{Sun2025}).
This produces a visually plausible, ``neutral'' candidate background that can be paired with many foregrounds.
For an image $i$ we now have $n_i$ foreground objects, extracted from $i$ by cutting out masked regions, each paired with a background where the same mask has been infilled.

\textbf{Select a high-quality pair.}
Different masks can trade off including the full object versus leaking class cues into the background.
We therefore score each candidate pair using an ensemble $E$ of six pretrained classifiers (ViT/ResNet/Swin) trained on the original dataset.
For each model $m \in E$, we compute the class scores of the ground truth class $c$, $\P[m(\mathrm{fg})=c]$ on the foreground (with gray background) and $\P[m(\mathrm{bg})=c]$ on the background and combine them with a prior $\operatorname{size}(\cdot)$ (pixel count).
Intuitively, we prefer \textcolor{Red}{(\textit{i})} foregrounds that strongly support the ground-truth class and \textcolor{Green}{(\textit{ii})} backgrounds that do \emph{not} support the ground-truth class, while \textcolor{Mulberry}{(\textit{iii})} discouraging overly large foreground regions:

\begin{align} \begin{split} \label{eq:filtering-score}
		\text{score}(\mathrm{fg}, \mathrm{bg}, c) & = \color{Red}\log \left( \frac{1}{\abs E} \sum_{m \in E} \P[m(\mathrm{fg}) = c] \right)                                                    \\
		                                          & + \color{Green}\log \left( 1 - \frac{1}{\abs E}\sum_{m \in E} \P[m(\mathrm{bg}) = c] \right)                                               \\
		                                          & + \lambda \color{Mulberry} \log \left( 1 - \abs{\frac{\operatorname{size}(\mathrm{fg})}{\operatorname{size}(\mathrm{bg})} - \eps} \right).
	\end{split} \end{align}
We run a hyperparameter search using a manually annotated subset of foreground/background variants to find the factors in \Cref{eq:filtering-score}: $\lambda = 2$ and $\eps = 0.1$.
For each image, we keep the candidate mask with the highest score.

\textbf{Filter low-quality backgrounds.}
Finally, we discard backgrounds that are heavily ($\geq 80\%$) inpainted, as they tend to look synthetic and provide little useful diversity (see \Cref{apdx:infill-ratio}).
This step filters out $10\%$ of backgrounds.

Although segmentation is the main computational overhead, it is performed once offline and reused across all training runs.
On NVIDIA H100 GPUs, the segmentation stage computes at a rate of $5 338.6 \frac{\text{img}}{\text{GPU} \times \text{h}}$ when inpainting with LaMa.
For ImageNet this comes down to just under $30$ hours on a single node.
At roughly twice the cost of a single ViT-B training run ($\approx 14$ hours), this is a modest investment that is amortized over every subsequent experiment the dataset is used in.
For details see \Cref{apdx:foraug-compute}.
The output of the segmentation stage is a collection of foreground cut-outs (with transparency) and a pool of diverse, neutral backgrounds, which we use in the online recombination stage.
For ImageNet, we provide pre-computed segmentation and infill output\footnote{\code{<URL will go here>}}.

\subsection{Recombination}
\label{sec:recombination}
In each epoch, the recombination stage generates a recombined training sample for each foreground by (\textit{i}) choosing a background, (\textit{ii}) choosing a target foreground size, (\textit{iii}) sampling a placement, and (\textit{iv}) pasting the foreground using its alpha mask.
This exposes the model to controlled changes in context and spatial layout that are largely absent from standard augmentation.

\textbf{Background sampling.}
For each foreground object, we draw a random background using one of three increasingly challenging strategies:
(\textit{i}) \textit{Original}: use the object's own inpainted background (no context shift);
(\textit{ii}) \textit{Same-class}: sample a background from the pool of backgrounds belonging to the same class (slight, but plausible context shift);
(\textit{iii}) \textit{All-classes}: sample from the pool of all backgrounds (large context shift).
These strategies trade off context diversity against semantic plausibility.
We ensure each foreground is used exactly once per epoch; backgrounds may repeat.

\textbf{Foreground scaling.}
Let $r_{\text{fg}}$ denote the relative foreground area in the source image of the foreground, and $r_{\text{bg}}$ the relative foreground area in the source image of the background. %
We compute the lower/upper size limits $(s_l, s_u)$ from these two ratios using one of two variants:
(\textit{i}) \emph{mean} sets $s_l = s_u = \text{mean}[r_\text{fg}, r_\text{bg}]$, while
(\textit{ii}) \emph{range} uses the min and max to preserve a wider scale range.
We sample the final scale from a $\pm 30\%$ interval around $(s_l, s_u)$ and resize the foreground to this scale, while keeping the aspect ratio.

\textbf{Placement and boundary smoothing.}
We paste the resized foreground at a uniformly random location within the background.
To reduce cut-and-paste artifacts, we slightly soften the alpha mask boundary by applying a Gaussian blur with $\sigma \sim \mathcal U [\frac{\sigma_{\text{max}}}{10}, \sigma_{\text{max}}]$, following the range used in modern augmentation~\cite{Touvron2022}.

\textbf{Mixing with original images.}
We optionally mix recombined samples with original dataset images.
A mixing ratio $p$ acts as the Bernoulli-probability of drawing the original image; otherwise we use its foreground and apply \schemename.
We consider constant $p$ as well as linear/cosine schedules that increase $p$ over training.
Finally, we apply standard data augmentation techniques on the resulting images.

The online recombination is CPU-parallel and does not measurably increase training time.
We find a $\approx 1\%$ increase in average step-time (see \Cref{apdx:foraug-compute}).

%% file: sec/experiments.tex
\begin{table}[t]
  \caption{\textbf{ImageNet results with different augmentation pipelines.}
      \schemename consistently improves performance in low- and mid-augmentation regimes and remains complementary to strong augmentation pipelines, with larger gains for larger models.
    }
    \label{tab:imagenet-pipelines}
    \centering
    \resizebox{\columnwidth}{!}{
      \setlength{\tabcolsep}{3pt}
      \begin{tabular}{lccccc}
        \toprule
        \multirow{2.5}{*}{Augmentation} & \multirow{2.5}{*}{MixUp} & \multirow{2.5}{*}{CutMix} & \multicolumn{3}{c}{Accuracy [\%]}                                      \\
        \cmidrule(l){4-6}
                                        &                          &                           & ViT-S                                   & ViT-B            & ViT-L           \\
        \midrule
        Basic                           & \xmark                   & \xmark                    & $71.9 \pm 0.1$                          & $69.5 \pm 0.2$   & $68.3 \pm 0.4$  \\
        Basic + \schemename             & \xmark                   & \xmark                    & $75.7 \pm 0.2$                          & $75.5 \pm 0.6$   & $73.1 \pm 1.7$  \\
                                        &                          &                           & \grntxt{$+3.8$}                         & \grntxt{$+6.0$}  & \grntxt{$+4.8$} \\
        \midrule
        RandAugment                     & \xmark                   & \xmark                    & $76.3 \pm 0.5$                          & $75.5 \pm 0.2$   & $74.7 \pm 0.4$  \\
        RandAugment + \schemename       & \xmark                   & \xmark                    & $78.0 \pm 0.1$                          & $77.8 \pm 0.1$   & $78.0 \pm 0.6$  \\
                                        &                          &                           & \grntxt{$+1.7$}                         & \grntxt{$+2.3$}  & \grntxt{$+3.3$} \\
        \midrule
        3-Augment                       & \xmark                   & \cmark                    & $79.1 \pm 0.1$                          & $77.6 \pm 0.2$   & $75.3 \pm 0.4$  \\
        3-Augment + \schemename         & \xmark                   & \cmark                    & $81.4 \pm 0.1$                          & $81.1 \pm 0.4$   & $79.8 \pm 0.1$  \\
                                        &                          &                           & \grntxt{$+2.3$}                         & \grntxt{$+3.5$}  & \grntxt{$+4.5$} \\
        \midrule
        Basic                           & \cmark                   & \cmark                    & $79.8 \pm 0.3$                          & $78.6 \pm 0.4$   & $78.1 \pm 1.6$  \\
        Basic + \schemename             & \cmark                   & \cmark                    & $79.8 \pm 0.3$                          & $81.6 \pm 0.5$   & $81.0 \pm 0.4$  \\
                                        &                          &                           & \gtxt{$\pm 0.0$}                        & \grntxt{$+3.0$}  & \grntxt{$+2.9$} \\
        \midrule
        RandAugment                     & \cmark                   & \cmark                    & $80.1 \pm 0.1$                          & $81.9 \pm 0.3$   & $79.3 \pm 2.3$  \\
        RandAugment + \schemename       & \cmark                   & \cmark                    & $80.0 \pm 0.3$                          & $81.9 \pm 0.2$   & $82.4 \pm 0.1$  \\
                                        &                          &                           & \gtxt{$-0.1$}                           & \gtxt{$\pm 0.0$} & \grntxt{$+3.1$} \\
        \bottomrule
      \end{tabular}
    }
\end{table}

\begin{table}
  \caption{\textbf{ImageNet results with different architectures.} \schemename improves the performance of most models, with a larger gain for larger models.}
    \label{tab:imagenet-results}
    \centering
    \setlength{\tabcolsep}{3pt}
    \resizebox{.7\columnwidth}{!}{
      \begin{tabular}{lccc}
        \toprule
        \multirow{2.5}{*}{Model} & \multicolumn{2}{c}{\makecell{Accuracy [\%]}} & \multirow{2.5}{*}{Delta}                   \\
        \cmidrule(lr){2-3}
                                 & Baseline                              & +\schemename           &                 \\
        \midrule
        ViT-S                    & $79.1\pm0.1$                                 & $81.4\pm0.1$             & \grntxt{$+2.3$} \\
        ViT-B                    & $77.6\pm0.2$                                 & $81.1\pm0.4$             & \grntxt{$+3.5$} \\
        ViT-L                    & $75.3\pm0.4$                                 & $79.8\pm0.1$             & \grntxt{$+4.5$} \\
        \midrule
        DeiT-S                   & $80.1 \pm 0.1$                               & $80.0\pm0.3$             & \gtxt{$-0.1$}   \\
        DeiT-B                   & $81.9 \pm 0.3$                               & $81.9\pm0.2$             & \gtxt{$\pm0.0$} \\
        DeiT-L                   & $79.3\pm2.3$                                 & $82.4\pm0.1$             & \grntxt{$+3.1$} \\
        \midrule
        Swin-Ti                  & $77.9\pm0.2$                                 & $79.7\pm0.1$             & \grntxt{$+1.8$} \\
        Swin-S                   & $79.4\pm0.1$                                 & $80.6\pm0.1$             & \grntxt{$+1.2$} \\
        \midrule
        ResNet-50                & $78.3\pm0.1$                                 & $78.8\pm0.1$             & \grntxt{$+0.5$} \\
        ResNet-101               & $79.4\pm0.1$                                 & $80.4\pm0.1$             & \grntxt{$+1.0$} \\
        \bottomrule
      \end{tabular}
    }
\end{table}

\section{Experiments}
\label{sec:experiments}

We comprehensively evaluate \schemename{} by comparing ImageNet training across 10 models and 5 augmentation pipelines.
Furthermore, we assess the impact on multiple fine-grained downstream datasets.
Finally, we exploit \schemename's control over the image distribution to quantify model behaviors and biases.
We always report the mean and standard deviation of three independent training runs.
More experiments and evaluations in the Appendix.

\subsection{Image Classification Results}

\textbf{ImageNet training.}
\Cref{tab:imagenet-pipelines} reports the effect of \schemename combined with different data augmentation pipelines:
A \emph{basic} pipeline with random-resized-crop, flip and color-jitter, the \emph{3-augment} pipeline from \cite{Touvron2022,Nauen2025} that also includes Grayscale, Solarization and gaussian-blur, as well as the widely used \emph{RandAugment}~\cite{Cubuk2020} based pipeline from DeiT~\cite{Touvron2021b}.
Additionally, we include MixUp~\cite{Zhang2018a} and CutMix~\cite{Yun2019} augmentations.
We find that the effectiveness of \schemename depends on the interplay between model capacity and baseline augmentation strength.
When the baseline augmentation is weak (\eg no MixUp/CutMix) or moderate (\emph{3-Augment}), \schemename consistently improves ImageNet accuracy, with gains increasing for larger ViT models (up to $+6.0$\,p.p.\ for ViT-B).
As the augmentation pipeline becomes stronger (\eg \emph{RandAugment} with MixUp/CutMix), ImageNet improvements diminish for smaller models, indicating that the baseline augmentation already saturates their capacity.
Importantly, even in cases where ImageNet accuracy does not improve, we consistently observe gains during downstream fine-tuning (see \Cref{tab:downstream-results}), suggesting that \schemename enhances representation quality beyond what is reflected by ImageNet accuracy.
For results of \schemename on other datasets and a comparison to AGA, see \Cref{apdx:augment-method-comparison,apdx:other-datasets}

\Cref{tab:imagenet-results} additionally compares performance across architectures.
ViT~\cite{Dosovitskiy2021}, Swin~\cite{Liu2021}, and ResNet~\cite{He2016} are trained using the \emph{3-augment} strategy, while DeiT~\cite{Touvron2021b} is trained using the \emph{RandAugment} strategy.
\schemename improves performance across all tested architectures, including the ResNet models, %
demonstrating benefits beyond Transformers.

\begin{table}[t]
  \caption{\textbf{Downstream accuracy [\%] after finetuning.} Models are pretrained on ImageNet with and without \schemename. \schemename{} is never used during finetuning. Pretraining using \schemename increases downstream accuracy; especially for Transformers.
  }
  \label{tab:downstream-results}
  \centering
  \setlength{\tabcolsep}{3pt}
    \resizebox{.95\columnwidth}{!}{\begin{tabular}{lcccccc}
        \toprule
        Model   & \schemename & Aircraft        & Cars            & Flowers         & Food            & Pets            \\
        \midrule
        \multirow{3}{*}{ViT-S}   & \xmark      & $72.4\pm1.0$    & $89.8\pm0.3$    & $94.5\pm0.2$    & $89.1\pm0.1$    & $93.8\pm0.2$    \\
        & \cmark      & $78.6\pm0.5$    & $92.2\pm0.2$    & $95.5\pm0.2$    & $89.6\pm0.1$    & $94.5\pm0.2$    \\
                &             & \grntxt{$+6.2$} & \grntxt{$+2.4$} & \grntxt{$+1.0$} & \grntxt{$+0.5$} & \grntxt{$+0.7$} \\
        \midrule
        \multirow{3}{*}{ViT-B}   & \xmark      & $71.7\pm0.5$    & $90.0\pm0.2$    & $94.8\pm0.4$    & $89.8\pm0.2$    & $94.1\pm0.4$    \\
        & \cmark      & $79.0\pm2.2$    & $93.3\pm0.1$    & $ 96.5\pm0.1$   & $90.9\pm0.1$    & $95.1\pm0.4$    \\
                &             & \grntxt{$+7.3$} & \grntxt{$+3.3$} & \grntxt{$+1.7$} & \grntxt{$+1.1$} & \grntxt{$+1.0$} \\
        \midrule
        \multirow{3}{*}{ViT-L}   & \xmark      & $72.1\pm1.0$    & $88.8\pm0.3$    & $94.4\pm0.3$    & $90.1\pm0.2$    & $94.2\pm0.4$    \\
        & \cmark      & $77.6\pm1.2$    & $89.1\pm0.2$    & $96.6\pm0.1$    & $91.3\pm0.1$    & $95.1\pm0.1$    \\
                &             & \grntxt{$+5.5$} & \grntxt{$+0.3$} & \grntxt{$+2.2$} & \grntxt{$+1.2$} & \grntxt{$+0.9$} \\
        \midrule
        & \xmark      & $75.3\pm0.4$    & $91.1\pm0.2$    & $94.8\pm0.4$    & $89.2\pm0.2$    & $92.4\pm0.2$            \\
        DeiT-S     & \cmark      & $76.8\pm0.8$    & $91.9\pm0.2$    & $95.2\pm0.3$    & $89.1\pm0.2$    & $92.3\pm0.4$            \\
                   &             & \grntxt{$+1.5$} & \grntxt{$+0.8$} & \grntxt{$+0.4$} & \gtxt{$-0.1$}   & \gtxt{$-0.1$}           \\
        \midrule
        & \xmark      & $77.0\pm1.2$    & $92.9\pm0.2$    & $96.1\pm0.2$    & $91.2\pm0.1$    & $93.3\pm0.4$            \\
        DeiT-B     & \cmark      & $79.3\pm0.3$    & $93.1\pm0.1$    & $96.4\pm0.2$    & $91.3\pm0.1$    & $93.3\pm0.1$            \\
                   &             & \grntxt{$+2.3$} & \gtxt{$+0.2$}   & \grntxt{$+0.3$} & \gtxt{$+0.1$}   & \gtxt{$\pm0.0$}         \\
        \midrule
        & \xmark      & $72.8\pm5.5$    & $92.8\pm1.0$    & $95.8\pm1.5$    & $90.5\pm2.6$    & $92.4\pm2.0$            \\
        DeiT-L     & \cmark      & $78.8\pm0.8$    & $93.8\pm0.2$    & $97.0\pm0.2$    & $92.0\pm0.2$    & $93.5\pm0.2$            \\
                   &             & \grntxt{$+6.0$} & \grntxt{$+1.0$} & \grntxt{$+1.2$} & \grntxt{$+1.5$} & \grntxt{$+1.1$}         \\
        \midrule
        \multirow{3}{*}{Swin-Ti} & \xmark      & $77.0\pm0.1$    & $91.3\pm0.6$    & $95.9\pm0.1$    & $90.0\pm0.2$    & $94.2\pm0.1$    \\
        & \cmark      & $81.1\pm0.8$    & $92.8\pm0.4$    & $96.2\pm0.1$    & $90.4\pm0.3$    & $94.8\pm0.5$    \\
                &             & \grntxt{$+4.1$} & \grntxt{$+2.5$} & \grntxt{$+0.3$} & \grntxt{$+0.4$} & \grntxt{$+0.6$} \\
        \midrule
        \multirow{3}{*}{Swin-S}  & \xmark      & $75.7\pm1.4$    & $91.0\pm0.3$    & $95.9\pm0.5$    & $91.1\pm0.2$    & $94.4\pm0.1$    \\
        & \cmark      & $81.4\pm0.2$    & $93.1\pm0.2$    & $96.3\pm0.3$    & $91.2\pm0.2$    & $94.9\pm0.3$    \\
                &             & \grntxt{$+5.7$} & \grntxt{$+2.1$} & \grntxt{$+1.4$} & \gtxt{$+0.1$}   & \grntxt{$+0.5$} \\
        \midrule
        & \xmark      & $78.2\pm0.5$    & $89.8\pm0.2$    & $91.7\pm0.4$    & $84.4\pm0.2$    & $93.7\pm0.3$            \\
        ResNet-50  & \cmark      & $80.3\pm0.4$    & $90.4\pm0.2$    & $91.7\pm0.2$    & $84.5\pm0.2$    & $93.7\pm0.3$            \\
                   &             & \grntxt{$+2.1$} & \grntxt{$+0.6$} & \gtxt{$\pm0.0$} & \gtxt{$+0.1$}   & \gtxt{$\pm0.0$}         \\
        \midrule
        & \xmark      & $78.4\pm0.6$    & $90.3\pm0.1$    & $91.2\pm0.5$    & $86.0\pm0.2$    & $94.3\pm0.2$            \\
        ResNet-101 & \cmark      & $81.4\pm0.5$    & $91.3\pm0.1$    & $92.9\pm0.2$    & $86.3\pm0.1$    & $94.0\pm0.3$            \\
                   &             & \grntxt{$+3.0$} & \grntxt{$+1.3$} & \grntxt{$+1.7$} & \grntxt{$+0.3$} & \textcolor{red}{$-0.3$} \\
        \bottomrule
      \end{tabular}}
\end{table}

\textbf{Downstream tasks.}
To assess the transferability of \schemename-trained models, we finetune models pretrained on ImageNet on five fine-grained datasets:
FGVC-Aircraft \cite{Maji2013}, Stanford Cars~\cite{Dehghan2017}, Oxford Flowers \cite{Nilsback2008}, Food-101 \cite{Kaur2017}, and Oxford-IIIT Pets \cite{Parkhi2012}.
In \Cref{tab:downstream-results}, we see transformer accuracies improve on all datasets by up to $7.3$\,p.p.
ResNet does see an improvement on average, but not on all individual tasks.
Notably, pretraining with \schemename boosts the downstream performance of DeiT-S and DeiT-B, despite \emph{not} increasing their ImageNet accuracy (see \Cref{tab:imagenet-results}).
Thus, the improved representations from training with \schemename translate to gains beyond what's explained by pretraining accuracy alone.

\begin{table}[t]
  \caption{\textbf{Standard robustness benchmarks.} \schemename generally increases models' robustness to different image distribution shifts. Note that ViT-S \emph{with} \schemename outperforms DeiT-S, the only model where \schemename does not increase robustness.}
  \label{tab:robustness-datasets}
  \setlength{\tabcolsep}{3pt}
  \centering
    \resizebox{.95\columnwidth}{!}{
      \begin{tabular}{lccccccc}
        \toprule
        Model   & \schemename & IN-Hard         & IN-A             & IN-C             & IN-R             & IN-V2           \\
        \midrule
        & \xmark         & $18.1 \pm 0.6$  & $18.8 \pm 0.2$   & $44.7 \pm 0.8$   & $41.6 \pm 0.6$   & $67.3 \pm 0.4$  \\
        ViT-S   & \cmark         & $21.0 \pm 0.4$  & $26.5 \pm 0.4$   & $52.6 \pm 0.6$   & $49.8 \pm 0.3$   & $70.6 \pm 0.1$  \\
                &                & \grntxt{$+2.9$} & \grntxt{$+7.7$}  & \grntxt{$+7.9$}  & \grntxt{$+8.1$}  & \grntxt{$+3.3$} \\
        \midrule
        & \xmark         & $17.0 \pm 0.4$  & $15.8 \pm 0.7$   & $40.4 \pm 0.8$   & $38.4 \pm 0.7$   & $65.1 \pm 0.6$  \\
        ViT-B   & \cmark         & $22.0 \pm 0.9$  & $31.9 \pm 1.5$   & $51.6 \pm 1.8$   & $48.7 \pm 1.7$   & $70.3 \pm 0.9$  \\
                &                & \grntxt{$+5.0$} & \grntxt{$+16.0$} & \grntxt{$+11.2$} & \grntxt{$+10.3$} & \grntxt{$+5.2$} \\
        \midrule
        & \xmark         & $15.6 \pm 0.4$  & $11.3 \pm 0.9$   & $38.4 \pm 1.0$   & $36.8 \pm 0.8$   & $61.6 \pm 0.8$  \\
        ViT-L   & \cmark         & $20.6 \pm 0.1$  & $30.4 \pm 0.5$   & $48.2 \pm 0.7$   & $46.0 \pm 0.4$   & $68.7 \pm 0.3$  \\
                &                & \grntxt{$+5.0$} & \grntxt{$+19.0$} & \grntxt{$+9.8$}  & \grntxt{$+9.3$}  & \grntxt{$+7.1$} \\
        \midrule
        & \xmark         & $19.5 \pm 0.2$   & $18.4 \pm 0.3$  & $58.8 \pm 0.7$  & $43.0 \pm 0.1$  & $68.8 \pm 0.2$  \\
        DeiT-S    & \cmark         & $18.5 \pm 0.5$   & $17.3 \pm 1.0$  & $57.0 \pm 0.9$  & $43.8 \pm 0.2$  & $68.7 \pm 0.6$  \\
                  &                & \rdtxt{$-1.0$}   & \rdtxt{$-1.1$}  & \rdtxt{$-1.8$}  & \grntxt{$+0.8$} & \gtxt{$-0.1$}   \\
        \midrule
        & \xmark         & $22.6 \pm 0.2$   & $26.0 \pm 0.2$  & $62.1 \pm 1.0$  & $45.6 \pm 1.9$  & $70.6 \pm 0.9$  \\
        DeiT-B    & \cmark         & $22.6 \pm 0.2$   & $25.0 \pm 0.3$  & $62.8 \pm 0.6$  & $47.7 \pm 0.8$  & $70.8 \pm 0.5$  \\
                  &                & \gtxt{$\pm 0.0$} & \rdtxt{$-1.0$}  & \grntxt{$+0.8$} & \grntxt{$+2.0$} & \gtxt{$+0.2$}   \\
        \midrule
        & \xmark         & $21.2 \pm 2.0$   & $20.2 \pm 3.4$  & $59.3 \pm 4.3$  & $41.3 \pm 2.7$  & $66.9 \pm 2.8$  \\
        DeiT-L    & \cmark         & $23.4 \pm 0.3$   & $28.8 \pm 2.0$  & $63.4 \pm 0.7$  & $47.8 \pm 0.6$  & $71.6 \pm 0.5$  \\
                  &                & \grntxt{$+2.2$}  & \grntxt{$+8.7$} & \grntxt{$+4.1$} & \grntxt{$+6.5$} & \grntxt{$+4.7$} \\
        \midrule
        & \xmark         & $16.2 \pm 0.4$  & $15.0 \pm 0.3$   & $36.0 \pm 0.8$   & $36.6 \pm 0.2$   & $65.5 \pm 0.4$  \\
        Swin-Ti & \cmark         & $18.3 \pm 0.3$  & $20.3 \pm 0.4$   & $41.4 \pm 0.8$   & $41.4 \pm 0.2$   & $68.2 \pm 0.4$  \\
                &                & \grntxt{$+2.2$} & \grntxt{$+5.4$}  & \grntxt{$+5.4$}  & \grntxt{$+4.8$}  & \grntxt{$+2.7$} \\
        \midrule
        & \xmark         & $18.2 \pm 0.3$  & $19.4 \pm 0.3$   & $39.0 \pm 0.7$   & $39.1 \pm 0.2$   & $67.5 \pm 0.1$  \\
        Swin-S  & \cmark         & $20.5 \pm 0.1$  & $27.7 \pm 0.4$   & $45.6 \pm 0.8$   & $44.1 \pm 0.3$   & $69.6 \pm 0.1$  \\
                &                & \grntxt{$+2.2$} & \grntxt{$+8.4$}  & \grntxt{$+6.6$}  & \grntxt{$+5.0$}  & \grntxt{$+2.2$} \\
        \midrule
        & \xmark         & $16.1 \pm 0.2$   & $9.7 \pm 0.1$   & $38.0 \pm 1.0$  & $40.5 \pm 0.6$  & $66.8 \pm 0.4$  \\
        ResNet50  & \cmark         & $17.2 \pm 0.1$   & $10.8 \pm 0.4$  & $41.0 \pm 0.7$  & $43.7 \pm 0.3$  & $67.5 \pm 0.1$  \\
                  &                & \grntxt{$+1.1$}  & \grntxt{$+1.1$} & \grntxt{$+3.0$} & \grntxt{$+3.2$} & \grntxt{$+0.7$} \\
        \midrule
        & \xmark         & $18.2 \pm 0.4$   & $14.3 \pm 0.1$  & $41.7 \pm 0.7$  & $42.3 \pm 0.1$  & $67.7 \pm 0.5$  \\
        ResNet101 & \cmark         & $19.9 \pm 0.2$   & $17.6 \pm 0.5$  & $46.3 \pm 0.6$  & $46.3 \pm 0.3$  & $69.5 \pm 0.3$  \\
                  &                & \grntxt{$+1.7$}  & \grntxt{$+3.2$} & \grntxt{$+4.6$} & \grntxt{$+4.0$} & \grntxt{$+1.8$} \\
        \bottomrule
      \end{tabular}
    }
\end{table}

\subsection{Bias and Robustness Evaluation}
Beyond its use for training, \schemename's unique properties and controlled data generation capabilities make it a powerful tool for analyzing behavior and biases of black-box models.
We exploit this in two complementary ways.
First, we evaluate the effect of \schemename during training on standard ImageNet robustness benchmarks.
Second, we use \schemename{}’s control over image composition to evaluate specific model biases, such as background reliance and center/size bias.

\textbf{Robustness on Standard Benchmarks.}
\Cref{tab:robustness-datasets} summarizes accuracy on five widely used ImageNet robustness benchmarks: ImageNet-Hard~\cite{Taesiri2023}, ImageNet-A~\cite{Hendrycks2021}, ImageNet-C~\cite{Hendrycks2019}, ImageNet-R~\cite{Hendrycks2021a}, and ImageNetV2~\cite{Recht2019}.
Across ViTs, Swin Transformers, and ResNets, incorporating \schemename during training generally improves robustness to all considered distribution shifts.
For ViTs, the gains are substantial: 
ViT-B improves $+16.0$\,p.p.\ accuracy on ImageNet-A and $+11.2$\,p.p.\ on ImageNet-C, with similar improvements for ViT-S and ViT-L.
Swin also benefits consistently, with increases of roughly $2$--$8$\,p.p.\ on most benchmarks, and ResNet sees smaller but steady gains. %

For DeiT, the picture is more nuanced: DeiT-B and DeiT-L still enjoy robustness improvements, whereas DeiT-S exhibits small decreases on several benchmarks.
Interestingly, however, ViT-S trained with \schemename outperforms the DeiT-S baseline.
This suggests that controlled composition can partially close the robustness gap between lightly and heavily regularized models.
Overall, the consistent improvements on corruption-based, natural and hard examples indicate that the compositional invariances induced by \schemename extend beyond the specific foreground/background manipulations used in its construction.

\begin{figure*}[t]
  \centering
  \includegraphics[width=.85\textwidth]{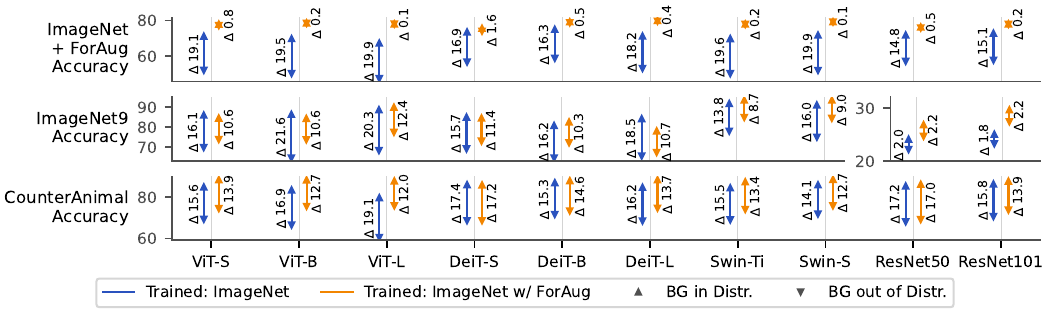}
  \caption[Background Robustness]{\textbf{Background robustness.} We plot the in-distribution (top of arrow, $\blacktriangle$) and the out-of-distribution (bottom of arrow, $\blacktriangledown$) accuracy when training with and without \schemename. Each arrow is annotated with its length $\Delta$. Shorter arrows indicate higher background robustness. Training with \schemename improves the background robustness of all transformers by mostly boosting the out-of-distribution accuracy.
  }
  \label{fig:background-robustness}
\end{figure*}

\textbf{Background Robustness.}
We assess the robustness of models to shifts in the background distribution from a class-related background to any background.
\Cref{fig:background-robustness} presents the background robustness results for three datasets: ImageNet with \schemename (all backgrounds\,$\blacktriangledown$ vs. same class\,$\blacktriangle$), ImageNet9~\cite{Xiao2021} (random\,$\blacktriangledown$ vs. original backgrounds\,$\blacktriangle$), and CounterAnimal~\cite{Wang2024f} (counter\,$\blacktriangledown$ vs. common background\,$\blacktriangle$).
The top triangle of each arrow represents the in-distribution backgrounds and the bottom triangle represents the out-of-distribution ones.
We follow ImageNet9 and CounterAnimal and assess the background robustness in terms of the accuracy gap when evaluating a model on images of normal background distribution compared to out-of-distribution backgrounds (length of each arrow; $\Delta$).
Crucially, \schemename improves the background robustness of all models and across datasets, reducing the background-gap by boosting the performance on the out-of-background-distribution samples more than the in-distribution ones.
We find a similar trend for the Corner-Cases~\cite{Fatima2025} dataset (see \Cref{apdx:corner-cases}), highlighting the generalization benefits of \schemename to unusual image compositions.

\begin{table}[t]
  \caption{\textbf{Center bias.}
    Accuracy relative to the center when the foreground object is in different cells of a $3 \times 3$ grid.
    Three independent runs for each setup.
    We calculate center bias according to \Cref{eq:center-bias}.
    Using \schemename significantly reduces center bias.}
  \label{tab:center-bias}
  \setlength{\tabcolsep}{4pt}
  \centering
    \resizebox{.8\columnwidth}{!}{
      \begin{tabular}{lccc}
        \toprule
        \multirow{2.5}{*}{Model} & \multicolumn{2}{c}{\makecell{Center Bias [\%]}}                                                                                                                                                                                   & \multirow{2.5}{*}{Delta}                                                                                                                                                                                                                                                         \\
        \cmidrule(lr){2-3}
                                 & Baseline                                                                                                                                                                                                                                & +\schemename                                                                                                                                                                                                                                                                   \\
        \midrule
        \multirow{2}{*}{ViT-S}                    & \includegraphics[width=.08\columnwidth, valign=c]{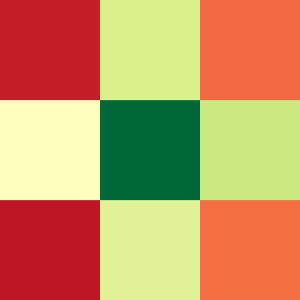} \includegraphics[width=.08\columnwidth, valign=c]{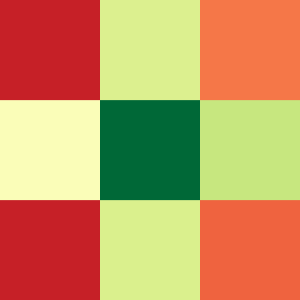} \includegraphics[width=.08\columnwidth, valign=c]{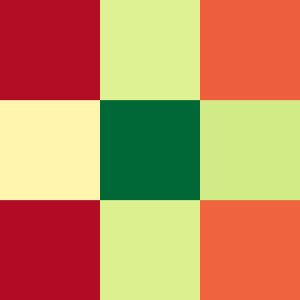}         & \includegraphics[width=.08\columnwidth, valign=c]{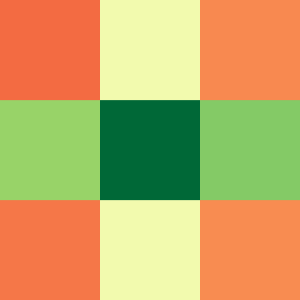} \includegraphics[width=.08\columnwidth, valign=c]{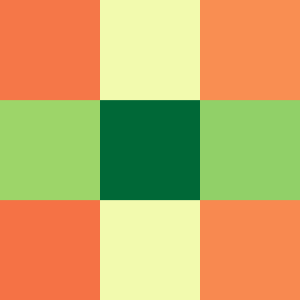} \includegraphics[width=.08\columnwidth, valign=c]{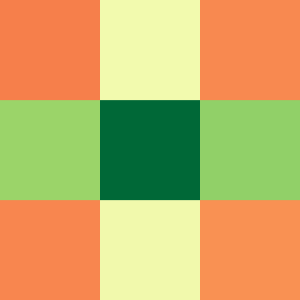}                            \\
                                 & $22.8\pm0.4$                                                                                                                                                                                                                                   & $19.9\pm0.1$                                                                                                                                                                                                                                                  & \grntxt{$-2.9$}  \\
        \multirow{2}{*}{ViT-B}                    & {\includegraphics[width=.08\columnwidth, valign=c]{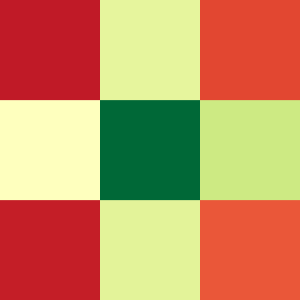} \includegraphics[width=.08\columnwidth, valign=c]{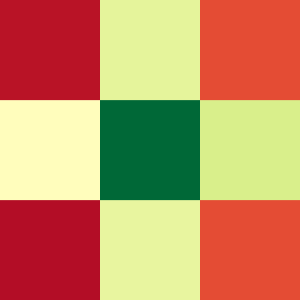} \includegraphics[width=.08\columnwidth, valign=c]{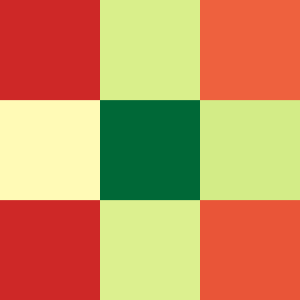}}       & \includegraphics[width=.08\columnwidth, valign=c]{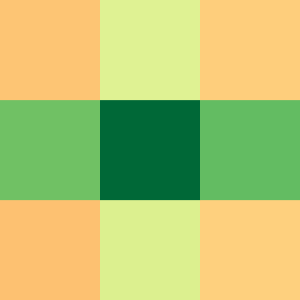} \includegraphics[width=.08\columnwidth, valign=c]{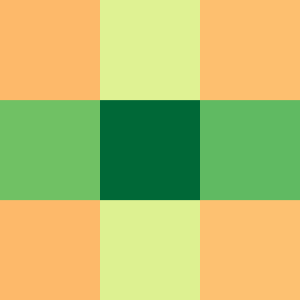} \includegraphics[width=.08\columnwidth, valign=c]{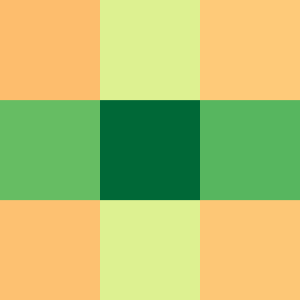}                            \\
                                 & $23.2\pm0.5$                                                                                                                                                                                                                                   & $16.9\pm0.2$                                                                                                                                                                                                                                                  & \grntxt{$-6.3$}  \\
        \multirow{2}{*}{ViT-L}                    & \includegraphics[width=.08\columnwidth, valign=c]{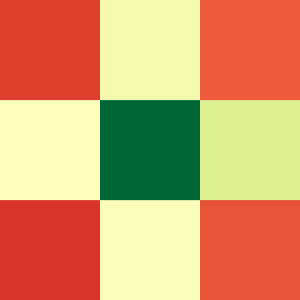} \includegraphics[width=.08\columnwidth, valign=c]{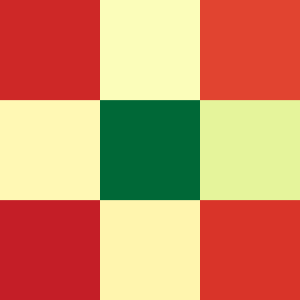} \includegraphics[width=.08\columnwidth, valign=c]{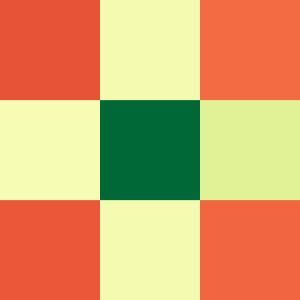}         & \includegraphics[width=.08\columnwidth, valign=c]{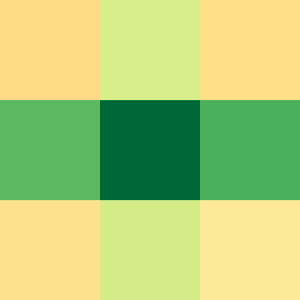} \includegraphics[width=.08\columnwidth, valign=c]{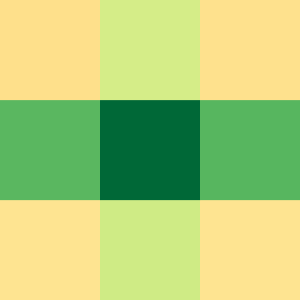} \includegraphics[width=.08\columnwidth, valign=c]{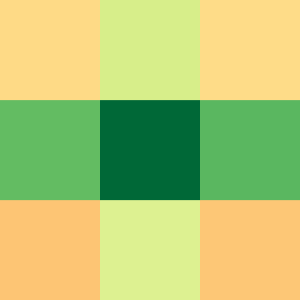}                            \\
                                 & $23.3\pm0.8$                                                                                                                                                                                                                                   & $15.6\pm0.5$                                                                                                                                                                                                                                                  & \grntxt{$-7.7$} \\
        \midrule
        \multirow{2}{*}{Swin-Ti}                  & {\includegraphics[width=.08\columnwidth, valign=c]{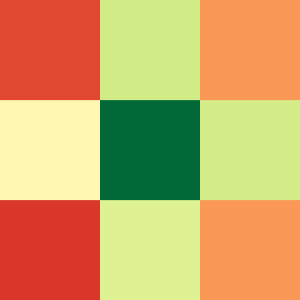} \includegraphics[width=.08\columnwidth, valign=c]{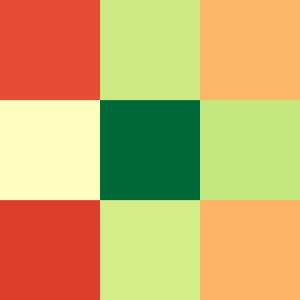} \includegraphics[width=.08\columnwidth, valign=c]{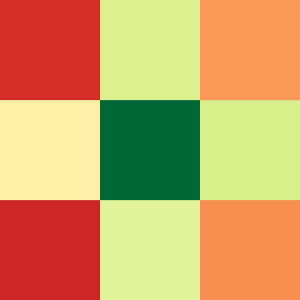}} & {\includegraphics[width=.08\columnwidth, valign=c]{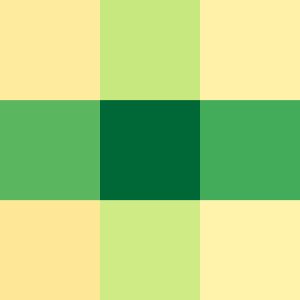} \includegraphics[width=.08\columnwidth, valign=c]{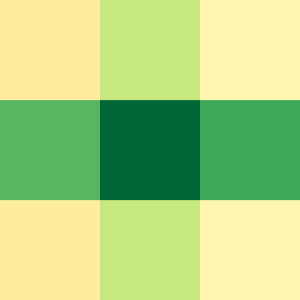} \includegraphics[width=.08\columnwidth, valign=c]{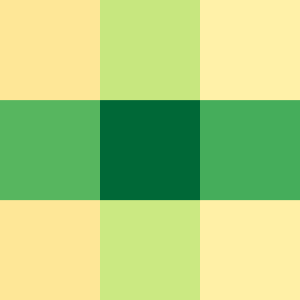}}                    \\
                                 & $21.6\pm0.7$                                                                                                                                                                                                                                   & $14.4\pm0.1$                                                                                                                                                                                                                                                  & \grntxt{$-7.2$}  \\
        \multirow{2}{*}{Swin-S}                   & {\includegraphics[width=.08\columnwidth, valign=c]{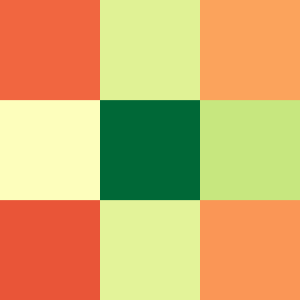} \includegraphics[width=.08\columnwidth, valign=c]{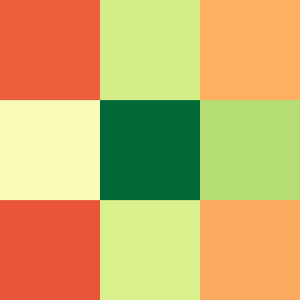} \includegraphics[width=.08\columnwidth, valign=c]{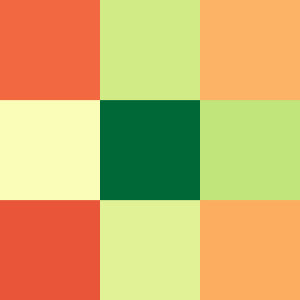}}    & {\includegraphics[width=.08\columnwidth, valign=c]{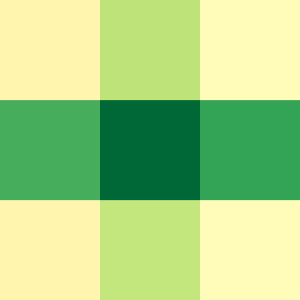} \includegraphics[width=.08\columnwidth, valign=c]{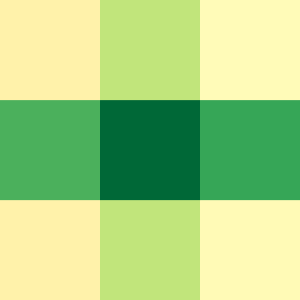} \includegraphics[width=.08\columnwidth, valign=c]{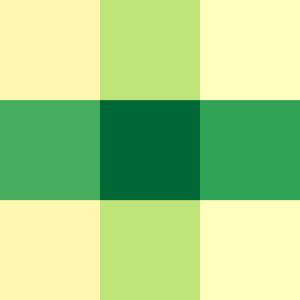}}                       \\
                                 & $20.6\pm0.3$                                                                                                                                                                                                                                   & $13.5\pm0.1$                                                                                                                                                                                                                                                  & \grntxt{$-7.1$}  \\
        \midrule
        \multirow{2}{*}{DeiT-S}                   & {\includegraphics[width=.08\columnwidth, valign=c]{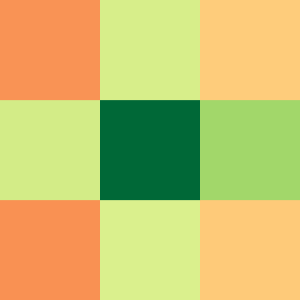} \includegraphics[width=.08\columnwidth, valign=c]{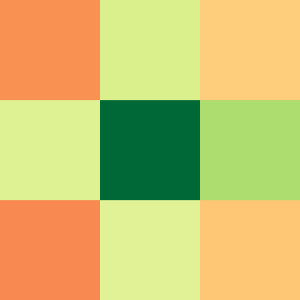} \includegraphics[width=.08\columnwidth, valign=c]{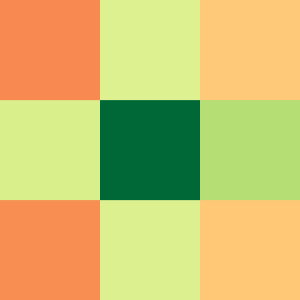}  }     & {\includegraphics[width=.08\columnwidth, valign=c]{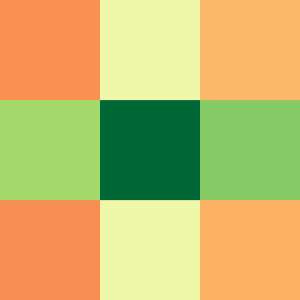} \includegraphics[width=.08\columnwidth, valign=c]{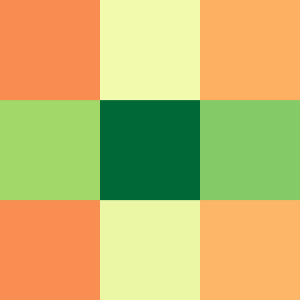} \includegraphics[width=.08\columnwidth, valign=c]{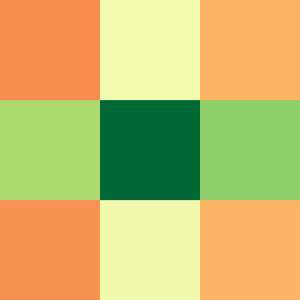}}                   \\
                                 & $18.9\pm0.2$ & $19.5\pm0.8$                                                                                                                                                                                                                                                         & \gtxt{$+0.6$}   \\
        \multirow{2}{*}{DeiT-B}                   & {\includegraphics[width=.08\columnwidth, valign=c]{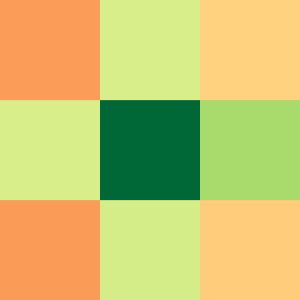} \includegraphics[width=.08\columnwidth, valign=c]{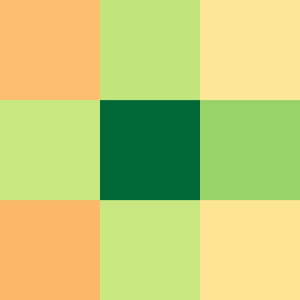} \includegraphics[width=.08\columnwidth, valign=c]{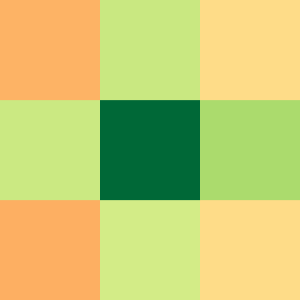}     }  & {\includegraphics[width=.08\columnwidth, valign=c]{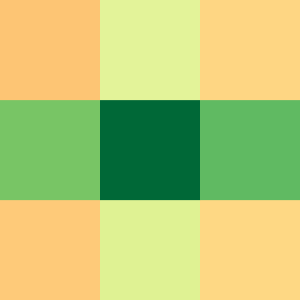} \includegraphics[width=.08\columnwidth, valign=c]{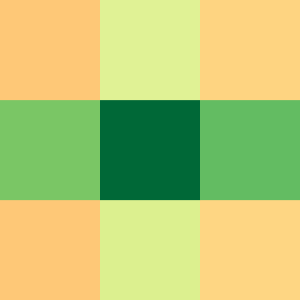} \includegraphics[width=.08\columnwidth, valign=c]{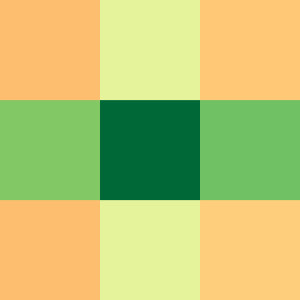}}                            \\
                                 & $17.7\pm0.8$                                                                                                                                                                                                                                       & $16.8\pm0.3$                                                                                                                                                                                                                                                         & \gtxt{$-0.9$} \\
        \multirow{2}{*}{DeiT-L}                   & { \includegraphics[width=.08\columnwidth, valign=c]{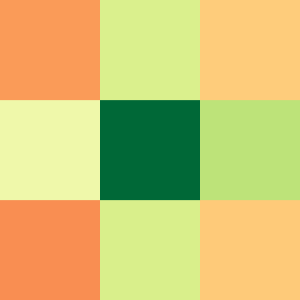} \includegraphics[width=.08\columnwidth, valign=c]{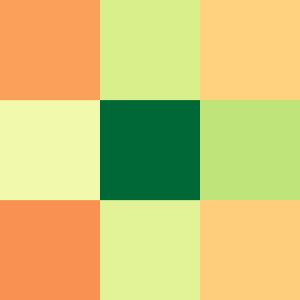} \includegraphics[width=.08\columnwidth, valign=c]{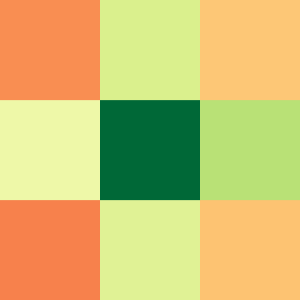}   }      & { \includegraphics[width=.08\columnwidth, valign=c]{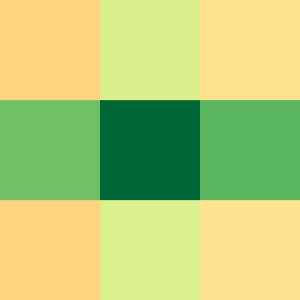} \includegraphics[width=.08\columnwidth, valign=c]{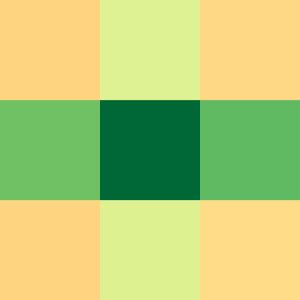} \includegraphics[width=.08\columnwidth, valign=c]{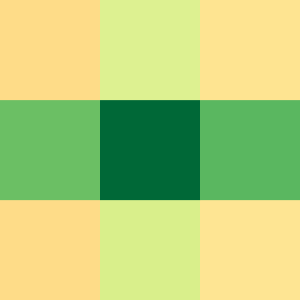} }                          \\
                                 & $19.2\pm0.2$                                                                                                                                                                                                                                       & $16.0\pm0.3$                                                                                                                                                                                                                                                         & \grntxt{$-3.2$} \\
        \midrule
        \multirow{2}{*}{ResNet50}                 & {\includegraphics[width=.08\columnwidth, valign=c]{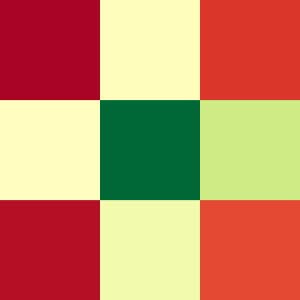} \includegraphics[width=.08\columnwidth, valign=c]{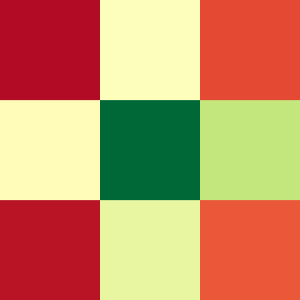} \includegraphics[width=.08\columnwidth, valign=c]{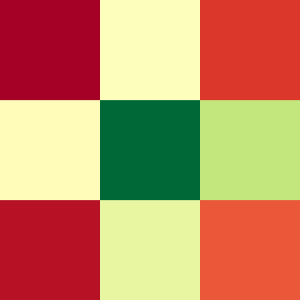}}    & {\includegraphics[width=.08\columnwidth, valign=c]{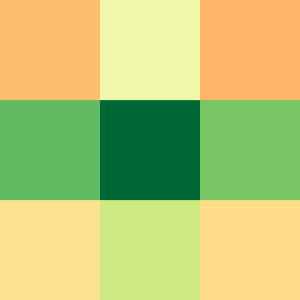} \includegraphics[width=.08\columnwidth, valign=c]{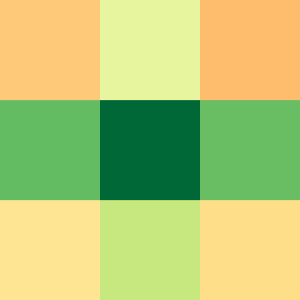} \includegraphics[width=.08\columnwidth, valign=c]{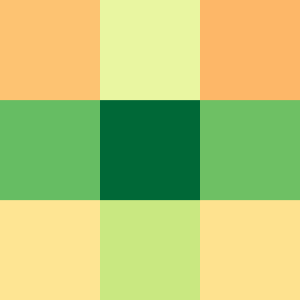}}                         \\
                                 & $24.0\pm0.3$                                                                                                                                                                                                                                         & $16.4\pm0.2$                                                                                                                                                                                                                                                           & \grntxt{$-7.6$} \\
        \multirow{2}{*}{ResNet101}                & {\includegraphics[width=.08\columnwidth, valign=c]{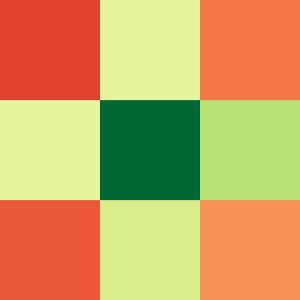} \includegraphics[width=.08\columnwidth, valign=c]{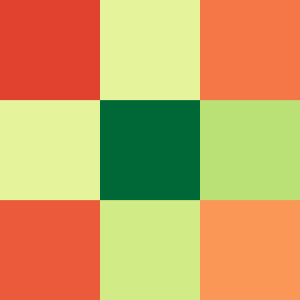} \includegraphics[width=.08\columnwidth, valign=c]{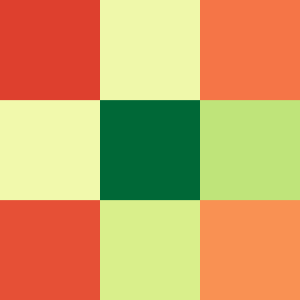}} & {\includegraphics[width=.08\columnwidth, valign=c]{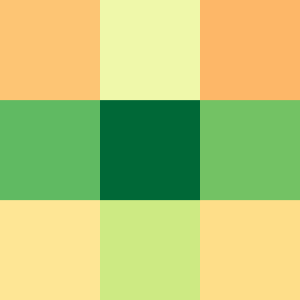} \includegraphics[width=.08\columnwidth, valign=c]{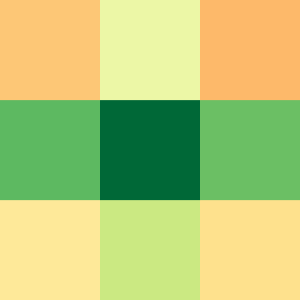} \includegraphics[width=.08\columnwidth, valign=c]{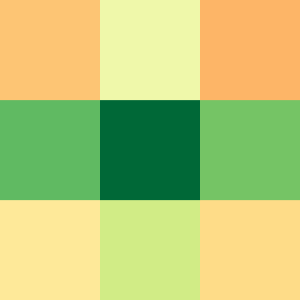}}                      \\
                                 & $21.2\pm0.3$                                                                                                                                                                                                                                         & $16.4\pm0.2$                                                                                                                                                                                                                                                           & \grntxt{$-4.8$} \\
        \bottomrule
      \end{tabular} }
  \includegraphics[width=.8\columnwidth]{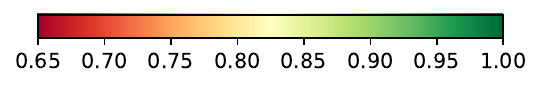}
\end{table}

\textbf{Center Bias.}
With \schemename we have unique control over the position of the foreground object in the image.
This lets us quantify the center bias of models trained with and without \schemename.
We divide the image into a $3 \times 3$ grid and evaluate model accuracy when the (scaled-down) foreground object is in each of the $9$ grid cells.
Each cell's accuracy is divided by the accuracy in the center cell for normalization, which gives us the relative performance drop when the foreground is in each part of the image.
The center bias is the relative performance difference of the center to other cells:
\begin{align} \label{eq:center-bias}
  \text{Center Bias} = 1 - \frac{\mean_{c \in \text{cells} \setminus \{ c_\text{center} \}}(\text{Acc}(c))}{\text{Acc}(c_\text{center})}
\end{align}
\Cref{tab:center-bias} visualizes the center bias of three instantiations of each model.
Performance is generally highest in the center and lowest in the four corners.
Interestingly, ImageNet-trained models perform slightly better when the foreground object is on the right side of the image, compared to the left side, despite our use of random flipping with a probability of $0.5$ during training.
Using \schemename significantly reduces center bias across models, with a more uniform performance especially across the middle row.
Thus, \schemename makes the model recognize objects across a wider spatial distribution, counteracting the center-bias of datasets like ImageNet.

\begin{figure}[t!]
  \centering
  \includegraphics[width=.95\columnwidth]{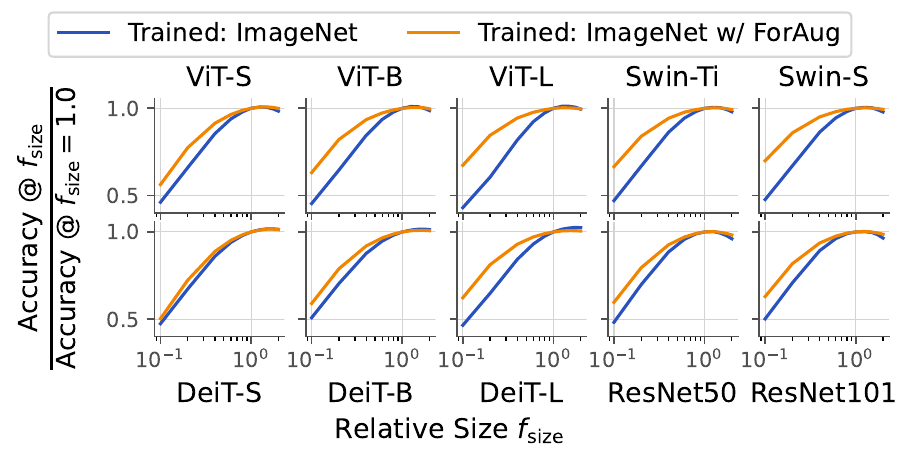}
  \caption{\textbf{Size bias.} We plot accuracy when scaling object size with the size factor $f_\text{size}$ relative to using the default size ($f_\text{size} = 1.0$). \schemename improves performance especially for small objects.}
  \label{fig:size-bias}
\end{figure}

\textbf{Size Bias.}
Finally, we evaluate the impact of different sized objects on the accuracy.
For this evaluation, we use the \emph{mean} foreground size strategy (see \Cref{sec:recombination}).
We introduce a size factor $f_\text{size}$ by which we additionally scale the foreground object before pasting it onto the background.
Results are normalized by the accuracy when using $f_\text{size} = 1.0$.
Size bias curves in \Cref{fig:size-bias} show that models using \schemename perform better, especially on small foreground objects (left in each plot).
\schemename-training improves robustness to variations in object scale, on top of better $f_\text{size} = 1.0$ accuracy.

\subsection{Design Choices of \schemename}
We next analyze key components of \schemename, focusing on three questions: 
How does it compare to copy-paste?
How does background choice affect performance?
And how reliably are labels preserved after recomposition?
Ablations of \schemename{}'s hyperparameters are in \Cref{apdx:ablation}.

\begin{table}[t]
  \caption{\textbf{Comparison of \schemename and Copy-Paste.} We train ViT-S on ImageNet using the same 3-augment data augmentation on top of copy-paste. Each line builds upon the previous one.}
  \label{tab:copy-paste-comparison}
  \centering
  \resizebox{.92\columnwidth}{!}{
    \begin{tabular}{lcc S[table-format=+2.1,retain-explicit-plus,detect-weight=true]}
      \toprule
      Augmentation (cumulative)                   & labels    & \makecell{     Accuracy [\%]} & {\makecell{Delta \\to Prev.}} \\
      \midrule
      3-Augment + \textbf{Simple Copy-Paste}       & bg        & $31.3 \pm 0.6$                &                  \\
      + mixed labels                               & fg + bg   & $32.0 \pm 0.8$                & +0.7             \\
      + fg labels                                  & fg        & $31.6 \pm 0.9$                & -0.4             \\
      + \emph{range} foreground size variation     & \gtxt{fg} & $43.0 \pm 1.2$                & \bfseries +11.4  \\
      + infilled backgrounds                       & \gtxt{fg} & $68.7 \pm 0.2$                & \bfseries +25.7  \\
      + \emph{cos} mixing strategy                 & \gtxt{fg} & $81.2 \pm 0.1$                & \bfseries +12.5  \\
      + edge smoothing                             & \gtxt{fg} & $81.3 \pm 0.1$                & +0.1             \\
      + background pruning$=$ \textbf{\schemename} & \gtxt{fg} & $81.4 \pm 0.1$                & +0.1             \\
      \bottomrule
    \end{tabular}}
\end{table}

\textbf{Comparison to Copy-Paste.}
We ablate \schemename{}'s individual components via comparison to a direct adaption of the Copy-Paste augmentation inspired by \cite{Ge2023,Ghiasi2021,Shermaine2025} in \Cref{tab:copy-paste-comparison}.
Contrary to semantic segmentation we do not have ground-truth masks available.
Thus, we paste the extracted objects from \textbf{\schemename's segmentation stage} onto original dataset images.
We observe 3 large jumps in accuracy: (\textbf{1}) From our \emph{range} foreground size variation (+11.4\,p.p.), (\textbf{2}) from using our infilled backgrounds instead of images from the dataset (+25.7\,p.p.), and (\textbf{3}) from our \emph{cos} mixing strategy with non-augmented images (+12.5\,p.p.).
\schemename's changes to the copy-paste augmentation are thus imperative for good performance in the classification task.

\begin{figure}[t]
    \centering
    \includegraphics[width=.75\columnwidth]{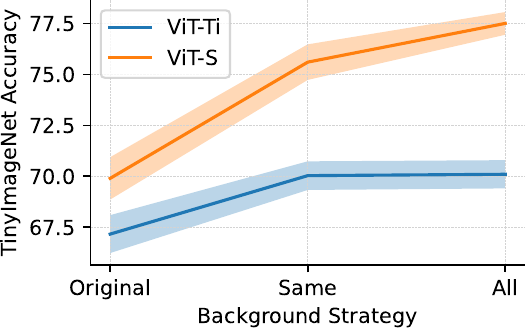}
    \caption{\textbf{Background sampling strategy.} We compare Original, Same-class, and All-classes background selection using ViT-Ti and ViT-S on TinyImageNet.
      Increasing background diversity consistently improves classification accuracy.
    }
    \label{fig:background-strategy}
  \end{figure}

  \begin{figure}[t]
    \centering
    \includegraphics[width=.75\columnwidth]{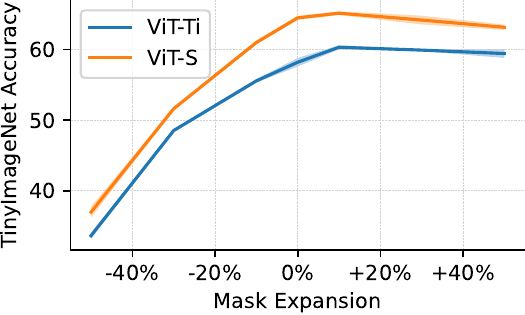}
    \captionof{figure}{\textbf{Mask corruption robustness.}
      We perturb masks for TinyImageNet by shrinking or expanding and report accuracy when training only on augmented samples.
      Performance is stable for expanded masks and degrades rapidly after shrinking masks.
    }
    \label{fig:mask-expansion}
\end{figure}

\textbf{Background Choice Strategy.}
\Cref{fig:background-strategy} shows the effect of background selection on TinyImageNet accuracy, where we trade off diversity against context plausibility.
Best performance is achieved by sampling backgrounds from all classes, which introduces substantial context shifts, but leads to the highest accuracy for both ViT-Ti and ViT-S.
Thus, aggressive background diversification is more important than context plausibility and acts as an effective form of context-based regularization rather than introducing harmful noise.

\textbf{Label Integrity.}
To quantify whether recombined images still depict the intended class, we evaluate the segmentation stage of \schemename using the ImageNet validation ground-truth bounding boxes.
Our predicted masks achieve a mean box precision (fraction of mask area inside ground-truth bounding boxes) of $91.0\%$ and a high box-to-box IoU of $76.1\%$, indicating our masks they tightly capture the target object.
Qualitative examples of the predicted masks and bounding boxes are provided in \Cref{apdx:segmentation-examples}.
We further probe robustness to mask imprecision by eroding or dilating masks such that the foreground area changes by a fixed percentage before composition.
Accuracy remains stable for expansions but drops under erosion (\Cref{fig:mask-expansion}), consistent with removing semantically important object parts.
Together, these results indicate that \schemename reliably isolates the target class and thus also fully neutralizes backgrounds by infilling, making recombined images faithful to the assigned labels.
Additionally, \schemename is robust to artifacts from an object's original background and degrades mainly when the foreground no longer contains the full object.

%% file: sec/conclusion.tex
\section{Conclusion}
\label{sec:conclusion}

We introduce \schemename, a controlled composition augmentation that factorizes images into objects and backgrounds and recombines them with explicit control over background identity, object position, and object scale.
Across diverse architectures, training with \schemename on top of standard strong augmentations yields substantial gains on ImageNet (up to $+6$\,p.p.) and fine-grained downstream tasks (up to $+7.3$\,p.p.), and consistently improves robustness on well-recognized benchmarks (up to $+19$\,p.p.).
\schemename's compositional controls additionally provide a framework for analyzing model behavior and quantifying biases, including background robustness, center bias, and size bias.
This dual role of \schemename as both a training mechanism and an evaluation tool highlights the value of explicit compositional factorization in understanding and improving image classifiers.
In future work, we aim to extend controlled composition beyond classification to dense prediction settings, including detection, segmentation, and video recognition.
More generally, we believe that designing augmentations around explicitly controllable and interpretable generative setups is a promising direction for building robust and reliable vision systems.

%% file: sec/acks.tex
\subsection*{Acknowledgements}
\label{sec:acknowledgements}
This work was funded by the Carl-Zeiss Foundation under the Sustainable Embedded AI project (P2021-02-009) and by the EU project SustainML (Horizon Europe grant agreement No 101070408).
All compute was done thanks to the Pegasus cluster at DFKI Kaiserslautern.

%% file: sec/appendix.tex
\section{Comparison of FG-BG Augmentation Methods}
\label{apdx:augment-method-comparison}
We compare \schemename to various foreground-background data augmentation methods.

\subsection{Conceptual Comparison}
\begin{table}
	\caption{\textbf{Comparison of FG-BG augmentations.} \schemename is the first controllable foreground-background augmentation that works for large-scale image classification.}\label{tab:novelty}
	\centering
	\resizebox{.95\columnwidth}{!}{
		\begin{tabular}[c]{lcccccccc}
			\toprule
			Method                         & Purpose        & \makecell{Controls                                                                    \\ pos+scale} & \makecell{Natural\\ FGs} & \makecell{Natural\\ BGs} & \makecell{Label-preserving\\ for Classification} & \makecell{Training\\ scale}      & \makecell{Stacks w/ \\ MixUp/CutMix} & \makecell{Online\\ Recombination} \\
			\midrule
			ImageNet-D \cite{Zhang2024f}   & eval           & \xmark             & \xmark & \cmark & \cmark & -                   & -      & -      \\
			ImageNet-9 \cite{Xiao2021}     & eval           & \xmark             & \cmark & \cmark & \cmark & -                   & -      & -      \\
			ObjectCompose \cite{Malik2024} & eval           & \xmark             & \cmark & \xmark & \cmark & -                   & -      & -      \\
			BG-Random \cite{Ryali2021}     & self-sup train & \xmark             & \cmark & \cmark & -      & ImageNet (1.3M)     & -      & \cmark \\
			Copy-Paste \cite{Ghiasi2021}   & seg train      & \cmark             & \cmark & \cmark & -      & COCO (118K)         & -      & \cmark \\
			StAug \cite{Kang2022}          & train          & \xmark             & \cmark & \cmark & \xmark & Plant Disease (48K) & \xmark & \cmark \\
			AGA \cite{Rahat2025}           & train          & \xmark (off. code) & \cmark & \xmark & \cmark & ImageNet10 (13K)    & \xmark & \xmark \\
			\textbf{\schemename} (ours)    & train + eval   & \cmark             & \cmark & \cmark & \cmark & ImageNet-1k (1.3M)  & \cmark & \cmark \\
			\bottomrule
		\end{tabular}}
\end{table}

\schemename's contribution is not the individual components, but the first controllable FG-BG augmentation viable for large-scale image-classification. %
No prior method offers this (\Cref{tab:novelty}).
AGA~\cite{Rahat2025}, the closest prior work, is demonstrated only on ImageNet10, where we outperform it at $40 \times$ lower cost (\Cref{tab:vs-aga}).
Concretely, \schemename is the only training augment to achieve \emph{any} of the following, let alone all four:
\begin{enumerate}
	\item explicity controls BG, position, and scale.
	\item produces unambiguous labels with natural FGs on natural (inpainted) BGs, avoiding synthetic-image distribution shift~\cite{Adamkiewicz2026}.
	\item is demonstrated on ImageNet-1k while stacking with MixUp/CutMix.
	\item doubles as a controlled diagnostic tool.
\end{enumerate}

\begin{table}[t!]
	\caption{\textbf{\schemename vs AGA.} Comparison of \schemename to AGA~\cite{Rahat2025} on ImageNet10.}\label{tab:vs-aga}
	\centering
	\begin{tabular}[c]{lccccc}
		\toprule
		\multirow{2.5}{*}{Method} & \multicolumn{4}{c}{Accuracy [\%] with} & \multirow{2.5}{*}{\makecell{preprocessing                                                                    \\                cost [H100-hours]}} \\
		\cmidrule(lr){2-5}
		                          & ResNet18                               & ResNet50                                  & ViT-Ti                  & ViT-S                                  \\
		\midrule
		CutMix                    & $88.5 \pm 0.1$                         & $90.2 \pm 0.3$                            & $73.6 \pm 1.1$          & $78.9 \pm 0.1$          & -            \\
		AGA$\times 2$             & $88.8 \pm 0.2$                         & $89.8 \pm 0.7$                            & $75.3 \pm 1.1$          & $80.5 \pm 1.6$          & 99.8         \\
		\textbf{\schemename}      & $\mathbf{89.9} \pm 0.3$                & $\mathbf{91.7} \pm 0.8$                   & $\mathbf{76.4} \pm 0.6$ & $\mathbf{80.9} \pm 0.3$ & \textbf{2.5} \\
		\bottomrule
	\end{tabular}
\end{table}

\subsection{Comparison of \schemename to AGA}
We compare to AGA on ImageNet10 (\Cref{tab:vs-aga}), the only scale at which AGA is demonstrated.
AGA does not specify training hyperparameters, and the released codebase contains only generation code for ImageNet10, no training code.
We thus train all methods under the same recipe~\cite{Touvron2022} for 300 effective epochs.
Both \schemename and AGA$\times 2$ replace CutMix in this setup, following AGA's original protocol.
AGA$\times 2$ receives 100 epochs on the $3 \times$ larger dataset, matching the number of gradient updates across methods.
\schemename outperforms AGA on ResNet and ViT while requiring $\mathbf{40 \times}$ \textbf{less preprocessing}.

\subsection{Compute Cost}
Per-sample synthesis methods~\cite{Rahat2025, Zhang2024f} scale linearly with dataset size: each sample needs its own generation step.
\schemename instead inpaints once per \emph{input}; recombination is then free, making it $\mathbf{40\times}$ \textbf{cheaper} than AGA (\Cref{tab:vs-aga}) and the only FG-BG augmentation tractable on ImageNet-1k.
Preprocessing costs roughly two ViT-B training runs (300 epochs), with filtering (Equation 1) handled by off-the-shelf pretrained models rather than custom classifiers.
This upfront cost enables \schemename's novel BG, position, and scale manipulation; exactly the axes where it delivers gains, including up to +19 p.p. on IN-A, and for fine-grained tasks.
Diminishing returns are a small-model artifact:
ViT-S with 3-augment+\schemename beats RandAugment and
\schemename adds +3.1 p.p. for ViT-L with strong RandAugment.

\subsection{On Background Distributions}
There are two distributions at play: the marginal $\mathbb P(\mathrm{bg})$ and the joint $\mathbb P(\mathrm{fg}, \mathrm{bg})$.
\schemename{}'s \emph{all-backgrounds} strategy intentionally breaks the joint distribution $\mathbb P(\mathrm{fg}, \mathrm{bg})$ by sampling backgrounds independently of foregrounds, but the marginals $\mathbb P(\mathrm{fg})$ and $\mathbb P(\mathrm{bg})$ are preserved.
This is fundamentally different from synthetic background generation (for example in AGA), which alters $\mathbb P(\mathrm{bg})$.
This creates "implausible" pairings, which is precisely the mechanism by which \schemename{} breaks spurious foreground-background correlations.
Empirically, this shift outperforms our more "plausible" same-class strategy.

\section{Resource Usage of \schemename}
\label{apdx:foraug-compute}
To utilize the proposed \schemename, specific computational resources are necessary, particularly for computing and storing for the output of the segmentation stage and for on-the-fly processing of the recombination stage.

\subsection{Segmentation.}
\schemename involves a computationally expensive segmentation and infill stage, which is a one-time calculation per dataset.
Once computed, the segmentation and infill results can be perpetually reused, amortizing the initial cost over all subsequent experiments and applications.
On NVIDIA H100 GPUs, the segmentation stage will compute at a rate of $374.3 \frac{\text{img}}{\text{GPU} \times \text{h}}$ when using Attentive Eraser or $5 338.6 \frac{\text{img}}{\text{GPU} \times \text{h}}$ for LaMa.
For ImageNet this comes down to just under 9 days (Attentive Eraser) or 16 hours (LaMa) on two 8 GPU nodes.
To facilitate immediate use and reproduction of results, we publicly provide the precalculated segmentation stage output for the ImageNet dataset for download\footnote{Link will go here.}.
The output of \schemename's segmentation step on ImageNet dataset requires 73 GB of additional disk space for the segmentation output, which is separate from the base 147 GB ImageNet size.

\subsection{Recombination.}
The recombination step of \schemename is implemented as a based data loader operation.
It's thus offloaded to the CPU, where it can be heavily parallelized and thus only results in a very minor increase in the training step-time.
For example, using a ViT-B model on an NVIDIA A100 GPU, the average update step-time increased by $1\%$, from $528 \pm 2$ ms to $534 \pm 1$ ms.

\section{\schemename on Other Datasets}
\label{apdx:other-datasets}

\begin{table}[t!]
	\caption{\textbf{\schemename on other datasets.} \schemename{} works when training on different datasets, especially for finegrained classification.}\label{tab:other-datasets}
	\centering
	\begin{tabular}[c]{llccr}
		\toprule
		Dataset                           & Model    & DeiT III [54]  & DeiT III + ForAug & Delta          \\
		\midrule
		\multirow{4}{*}{CIFAR100}         & ViT-Ti   & $77.6 \pm 0.2$ & $78.5 \pm 0.5$    & \grntxt{+0.9}  \\
		                                  & ViT-S    & $78.4 \pm 0.2$ & $80.4 \pm 0.3$    & \grntxt{+2.0}  \\
		                                  & ResNet18 & $75.4 \pm 0.3$ & $79.8 \pm 0.2$    & \grntxt{+4.4}  \\
		                                  & ResNet50 & $76.3 \pm 0.3$ & $79.8 \pm 0.3$    & \grntxt{+3.5}  \\
		\midrule
		\multirow{4}{*}{TinyImageNet}     & ViT-Ti   & $66.1 \pm 0.5$ & $70.1 \pm 0.7$    & \grntxt{+4.0}  \\
		                                  & ViT-S    & $68.3 \pm 0.7$ & $77.5 \pm 0.6$    & \grntxt{+9.2}  \\
		                                  & ResNet18 & $74.3 \pm 0.1$ & $77.0 \pm 0.3$    & \grntxt{+2.7}  \\
		                                  & ResNet50 & $77.8 \pm 0.2$ & $81.0 \pm 0.2$    & \grntxt{+3.2}  \\
		\midrule
		\multirow{3}{*}{iNaturalist 2011} & ViT-S    & $59.3 \pm 0.9$ & $68.2 \pm 0.1$    & \grntxt{+8.9}  \\
		                                  & ViT-B    & $56.3 \pm 4.2$ & $67.9 \pm 1.4$    & \grntxt{+11.6} \\
		                                  & ResNet50 & $55.3 \pm 0.2$ & $63.6 \pm 0.1$    & \grntxt{+8.3}  \\
		\bottomrule
	\end{tabular}
\end{table}

Additional to our ImageNet-1k results, we show in \Cref{tab:other-datasets} that ForAug consistently increases performance when training from scratch on different datasets of different scale: CIFAR100, TinyImageNet, and iNaturalist2011.

\section{Training Setup}
\label{apdx:training_setup}

\begin{table*}[h!]
	\centering
	\caption{\textbf{Training hyperparameters.} Training setup and hyperparameters for our ImageNet training.}
	\label{tab:in-setup}
	\resizebox{.8\textwidth}{!}{
		\begin{tabular}{lccc}
			\toprule
			Augmentation Pipeline: & Basic                                               & 3-Augment~\cite{Touvron2022} & RandAugment~\cite{Touvron2021b} \\
			\midrule
			Image Resolution       & \multicolumn{3}{c}{$224 \times 224$}                                                                                 \\
			Epochs                 & \multicolumn{3}{c}{300}                                                                                              \\
			Learning Rate          & S/B: 1e-3, L: 5e-4                                  & 3e-3                         & S/B: 1e-3, L: 5e-4              \\
			Learning Rate Schedule & \multicolumn{3}{c}{cosine decay}                                                                                     \\
			Batch Size             & 1024                                                & 2048                         & 1024                            \\
			GPUs                   & \multicolumn{3}{c}{$4\times$ NVIDIA A100/H100/H200}                                                                  \\
			Warmup Schedule        & \multicolumn{3}{c}{linear}                                                                                           \\
			Warmup Epochs          & \multicolumn{3}{c}{3}                                                                                                \\
			Weight Decay           & 0.05                                                & 0.02                         & 0.05                            \\
			Label Smoothing        & \multicolumn{3}{c}{0.1}                                                                                              \\
			Optimizer              & AdamW                                               & Lamb \cite{You2020}          & AdamW                           \\
			\midrule
			Augmentations          & \makecell{RandomResizedCrop                                                                                          \\ Horizontal Flip \\ ColorJitter}                                         & \makecell{Resize                                               \\ RandomCrop                                    \\ Horizontal Flip \\ Grayscale \\ Solarize \\ Gaussian-Blur \\ Color Jitter} &              \makecell{RandomResizedCrop \\ Horizontal Flip \\ RandomErase \cite{Zhong2020} \\ RandAugment \cite{Cubuk2020} \\ Color Jitter}                   \\
			\bottomrule
		\end{tabular}
	}
\end{table*}

\begin{table}[h!]
	\centering
	\caption{\textbf{Downstream training setup.} Training setup for finetuning on different downstream datasets. Other settings are the same as in \Cref{tab:in-setup}. For finetuning, we always utilize 3-Augment and the related parameters from the \emph{ViT, Swin, ResNet} column of \Cref{tab:in-setup}}
	\label{tab:downstream-setup}
	\begin{tabular}{lcccc}
		\toprule
		Dataset  & Batch Size & Epochs & Learning Rate & Num. GPUs \\
		\midrule
		Aircraft & 512        & 500    & 3e-4          & 2         \\
		Cars     & 1024       & 500    & 3e-4          & 4         \\
		Flowers  & 256        & 500    & 3e-4          & 1         \\
		Food     & 2048       & 100    & 3e-4          & 4         \\
		Pets     & 512        & 500    & 3e-4          & 2         \\
		\bottomrule
	\end{tabular}
\end{table}
On ImageNet, we test three different data augmentation pipelines and hyperparameter settings as shown in \Cref{tab:in-setup}: A basic pipeline, a pipeline using RandAugment based on the DeiT~\cite{Touvron2021b} setup and 3-Augment, as used in \cite{Touvron2022,Nauen2025}.
When comparing different architectures, ViT, Swin, and ResNet are trained with the 3-Augment pipeline and DeiT is trained with the RandAugment pipeline.
As our focus is on evaluating the changes in accuracy due to \schemename, like \cite{Nauen2025}, we stick to one set of hyperparameters for all models.
We list the settings used for training on ImageNet in \Cref{tab:in-setup} and the ones used for finetuning those weights on the downstream datasets in \Cref{tab:downstream-setup}.
Our implementation is using PyTorch \cite{Paszke2019} and the \emph{timm} library \cite{Wightman2019} for model architectures and basic functions.

\begin{table*}[ht!]
	\centering
	\caption{\textbf{Software versions.} Hardware and Software specifics used for both training and evaluation.}
	\label{tab:hw-sw-versions}
	\begin{tabular}{ll}
		\toprule
		Parameter        & Value                                                \\
		\midrule
		GPU              & $4 \times$ NVIDIA A100/H100/H200                     \\
		CPU              & 24 CPU cores (Intel Xenon) per GPU                   \\
		Memory           & up to 120 GB per GPU                                 \\
		Operating System & Enroot container for SLURM based on Ubuntu 24.04 LTS \\
		Python           & 3.12.3                                               \\
		PyTorch          & 2.7.0                                                \\
		TorchVision      & 0.22.0                                               \\
		Timm             & 1.0.15                                               \\
		\bottomrule
	\end{tabular}
\end{table*}
\Cref{tab:hw-sw-versions} lists the specific hardware we use, as well as versions of the relevant software packages.

\begin{figure*}[h!]
	\centering
	\includegraphics[width=.95\textwidth]{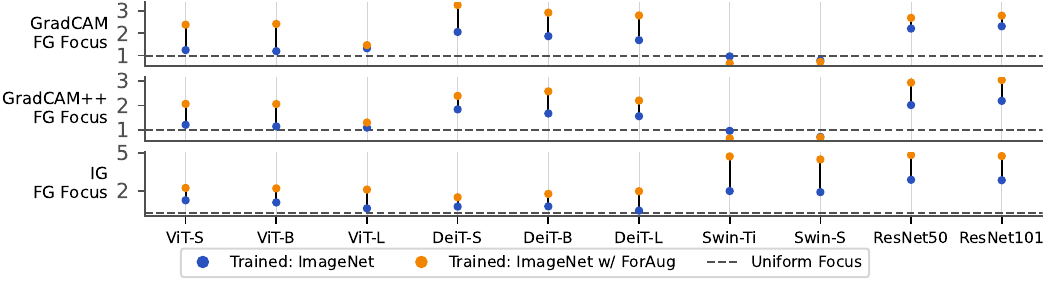}
	\caption{\textbf{Foreground focus.} Evaluation of the foreground focus (\Cref{eq:fg-focus}) using GradCam, GradCam++ and IntegratedGradients (IG) of models trained on ImageNet. Higher focus means more weight is placed on the object, relative to the background. Training with \schemename improves the foreground focus of almost all models.}
	\label{fig:foreground-focus}
\end{figure*}

\section{Foreground Focus}
\label{apdx:fg-focus}
Leveraging our inherent knowledge of the foreground masks when using \schemename, as well as common XAI techniques~\cite{Selvaraju2016,Chattopadhay2018,Sundararajan2017}, we can evaluate a model's focus on the foreground object.
We can directly evaluate ImageNet-trained models, but this technique can also be extended to other datasets without relying on manually annotated foreground masks.
To evaluate the foreground focus, we employ Grad-CAM \cite{Selvaraju2016}, Grad-CAM++ \cite{Chattopadhay2018} and IntegratedGradients (IG) \cite{Sundararajan2017} to compute the per-pixel importance of an image for the model's prediction.
The foreground focus is defined to be the ratio of the foreground's relative importance to its relative size in the image:
\begin{align} \label{eq:fg-focus}
	\text{FG Focus}(\text{img}) = \frac{\text{Importance}(\text{fg})}{\text{Importance}(\text{img})} \times \left(\frac{\text{Area}(\text{fg})}{\text{Area}(\text{img})}\right)^{-1}
\end{align}
If all pixels uniformly receive the same importance value, the foreground focus is one.
The foreground focus of a model is its average focus over all test images.
\Cref{fig:foreground-focus} shows that, using \schemename significantly increases the foreground focus of ViT, DeiT and ResNet across all XAI metrics.
We hypothesize Swin's below-uniform foreground focus with GradCam is due to its specific implementation.

\section{Extended Bates Distribution}
\label{apdx:bates-distribution}
\begin{figure}[h!]
	\centering
	\includegraphics[width=.5\columnwidth]{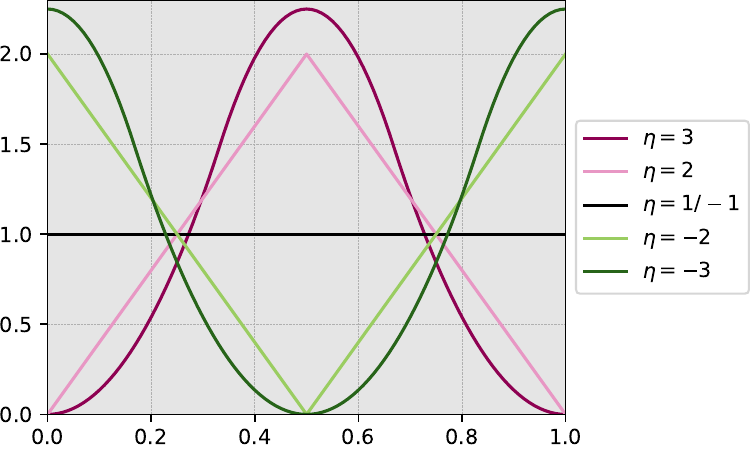}
	\caption{\textbf{Extended Bates PDF.} Plot of the probability distribution function (PDF) of the extended Bates distribution for different parameters $\eta$. Higher values of $\eta$ concentrate the distribution around the center.}
	\label{fig:bates-pdf}
\end{figure}

We introduce an extension of the Bates distribution~\cite{Bates1955} to include negative parameters, enabling sampling of foreground object positions away from the image center.
The standard Bates distribution, for $\eta \in \N$, is defined as the mean of $\eta$ independent random variables drawn from a uniform distribution \cite{Jonhson1995}.
A larger $\eta$ value increases the concentration of samples around the distribution's mean, which in this case is the image center.

To achieve an opposite effect--concentrating samples at the image borders--we extend the distribution to $\eta \leq 1$.
\begin{align*}
	X \sim \text{Bates}(\eta) :\Leftrightarrow s(X) \sim \text{Bates}(-\eta)
\end{align*}
This is accomplished by sampling from a standard Bates distribution with parameter $-\eta \geq 1$ and then applying a sawtooth function.
The sawtooth function on the interval $[0,1]$ is defined as
\begin{align}
	s(x) = \begin{cases}
		       x + 0.5 & \text{if } 0 < x < 0.5       \\
		       x - 0.5 & \text{if } 0.5 \leq x \leq 1
	       \end{cases}
\end{align}
This function effectively maps the central portion of the interval to the edges and the edge portions to the center.
For example, a value of 0.3 (central-left) is mapped to 0.8 (edge-right), while 0.8 (edge-right) is mapped to 0.3 (central-left).
This transformation inverts the distribution's concentration, shifting the probability mass from the center to the borders.
We visualize the distribution function of the extended Bates distribution in \Cref{fig:bates-pdf}.
Both $\eta = 1$ and $\eta = -1$ result in a uniform distribution across the image.

\section{Design Choices of \schemename}
\label{apdx:ablation}

We start by ablating the design choices of \schemename on TinyImageNet~\cite{Le2015}, a subset of ImageNet containing 200 categories with 500 images each. %
\Cref{tab:ablation-segment} presents ablations for segmentation and \Cref{tab:ablation-recombine} for recombination.

\begin{table}
	\caption{\textbf{\schemename segmentation ablation.} Ablation of the design decisions in the segmentation phase of \schemename on TinyImageNet.
		The first line is our baseline, while the other lines are using \schemename.
		We use basic settings with the \emph{same} background strategy during recombination for this experiment.
	}
	\label{tab:ablation-segment}
	\centering
	\small
	\begin{tabular}{llcc}
		\toprule
		\multirow{2.5}{*}{\makecell{Detect.                                                                          \\Prompt}} & \multirow{2.5}{*}{\makecell{Infill \\ Model}} & \multicolumn{2}{c}{TinyImageNet Accuracy [\%]}         \\
		\cmidrule{3-4}
		                                          &                                & ViT-Ti         & ViT-S          \\
		\midrule
		\multicolumn{2}{l}{\textbf{TinyImageNet}} & $66.1 \pm 0.5$                 & $68.3 \pm 0.7$                  \\
		specific                                  & LaMa \cite{Suvorov2022}        & $65.5 \pm 0.4$ & $71.2 \pm 0.5$ \\
		general                                   & \gtxt{LaMa \cite{Suvorov2022}} & $66.4 \pm 0.6$ & $72.9 \pm 0.6$ \\
		\gtxt{general}                            & Att. Eraser \cite{Sun2025}     & $67.5 \pm 1.2$ & $72.4 \pm 0.5$ \\
		\bottomrule
	\end{tabular}
\end{table}

\begin{table}[t]
	\caption{\textbf{\schemename recombination ablation.} Ablation of the recombination phase of \schemename on TinyImageNet (top) and ImageNet (bottom). The first experiments use the initial segmentation settings with LaMa \cite{Suvorov2022}.}
	\label{tab:ablation-recombine}
	\centering
	\begin{tabular}{ccccccccccc}
		\toprule
		\multirow{2.5}{*}{\makecell{FG.                                                                                                                                                      \\size}} & \multirow{2.5}{*}{\makecell{Augment.\\Order}} & \multirow{2.5}{*}{\makecell{BG\\Strat.}} & \multirow{2.5}{*}{\makecell{BG.\\Prune}} & \multirow{2.5}{*}{\makecell{Original\\Mixing}} & \multirow{2.5}{*}{\makecell{Edge\\Smooth.}} & \multicolumn{2}{c}{Accuracy [\%]} \\
		\cmidrule{7-8}
		                                          &                       &                     &            &              &                                  & ViT-Ti       & ViT-S        \\
		\midrule
		\multicolumn{6}{l}{\textbf{TinyImageNet}} & \gtxt{$66.1\pm0.5$}   & \gtxt{$68.3\pm0.7$}                                                                                              \\
		mean                                      & crop$\to$paste        & same                & -          & -            & \gtxt{-}                         & $64.6\pm0.5$ & $70.0\pm0.6$ \\
		range                                     & \gtxt{crop$\to$paste} & \gtxt{same}         & \gtxt{-}   & \gtxt{-}     & \gtxt{-}                         & $65.5\pm0.4$ & $71.2\pm0.5$ \\
		\midrule
		{range}                                   & {crop$\to$paste}      & {same}              & {-}        & {-}          & {-}                              & $67.5\pm1.2$ & $72.4\pm0.5$ \\
		\gtxt{range}                              & paste$\to$crop        & \gtxt{same}         & \gtxt{-}   & \gtxt{-}     & \gtxt{-}                         & $67.1\pm1.2$ & $72.9\pm0.5$ \\
		\gtxt{range}                              & \gtxt{paste$\to$crop} & \gtxt{same}         & 1.0        & \gtxt{-}     & \gtxt{-}                         & $67.0\pm1.2$ & $73.0\pm0.3$ \\
		\gtxt{range}                              & \gtxt{paste$\to$crop} & \gtxt{same}         & 0.8        & \gtxt{-}     & \gtxt{-}                         & $67.2\pm1.2$ & $72.9\pm0.8$ \\
		\gtxt{range}                              & \gtxt{paste$\to$crop} & \gtxt{same}         & 0.6        & \gtxt{-}     & \gtxt{-}                         & $67.5\pm1.0$ & $72.8\pm0.7$ \\
		\gtxt{range}                              & \gtxt{paste$\to$crop} & \gtxt{same}         & \gtxt{0.8} & $p=0.2$      & \gtxt{-}                         & $69.8\pm0.5$ & $75.0\pm0.3$ \\
		\gtxt{range}                              & \gtxt{paste$\to$crop} & \gtxt{same}         & \gtxt{0.8} & $p=0.33$     & \gtxt{-}                         & $69.5\pm0.4$ & $75.2\pm1.0$ \\
		\gtxt{range}                              & \gtxt{paste$\to$crop} & \gtxt{same}         & \gtxt{0.8} & $p=0.5$      & \gtxt{-}                         & $70.3\pm1.0$ & $74.2\pm0.2$ \\
		\gtxt{range}                              & \gtxt{paste$\to$crop} & \gtxt{same}         & \gtxt{0.8} & linear       & \gtxt{-}                         & $70.1\pm0.7$ & $74.9\pm0.8$ \\
		\gtxt{range}                              & \gtxt{paste$\to$crop} & \gtxt{same}         & \gtxt{0.8} & reverse lin. & \gtxt{-}                         & $67.6\pm0.2$ & $73.2\pm0.3$ \\
		\gtxt{range}                              & \gtxt{paste$\to$crop} & \gtxt{same}         & \gtxt{0.8} & cos          & \gtxt{-}                         & $71.3\pm1.0$ & $75.7\pm0.8$ \\
		\gtxt{range}                              & \gtxt{paste$\to$crop} & \gtxt{same}         & \gtxt{0.8} & \gtxt{cos}   & $\sigma_\text{max} = 4.0$        & $70.0\pm0.8$ & $75.5\pm0.7$ \\
		\gtxt{range}                              & \gtxt{paste$\to$crop} & orig.               & \gtxt{0.8} & \gtxt{cos}   & \gtxt{$\sigma_\text{max} = 4.0$} & $67.2\pm0.9$ & $69.9\pm1.0$ \\
		\gtxt{range}                              & \gtxt{paste$\to$crop} & all                 & \gtxt{0.8} & \gtxt{cos}   & \gtxt{$\sigma_\text{max} = 4.0$} & $70.1\pm0.7$ & $77.5\pm0.6$ \\
		\midrule
		\multicolumn{6}{l}{\textbf{ImageNet}}     & \gtxt{-}              & \gtxt{$79.1\pm0.1$}                                                                                              \\
		\gtxt{range}                              & \gtxt{paste$\to$crop} & \gtxt{same}         & \gtxt{0.8} & \gtxt{cos}   & \gtxt{-}                         & -            & $80.5\pm0.1$ \\
		\gtxt{range}                              & \gtxt{paste$\to$crop} & \gtxt{same}         & \gtxt{0.8} & \gtxt{cos}   & $\sigma_\text{max} = 4.0$        & -            & $80.7\pm0.1$ \\
		\gtxt{range}                              & \gtxt{paste$\to$crop} & all                 & \gtxt{0.8} & \gtxt{cos}   & \gtxt{$\sigma_\text{max} = 4.0$} & -            & $81.4\pm0.1$ \\
		\bottomrule
	\end{tabular}
\end{table}

\textbf{Prompt.}
First, we evaluate the type of prompt used to detect the foreground object.
Here, the \emph{general} prompt, which contains the class and the more general object category, outperforms only having the class name (\emph{specific}).

\textbf{Inpainting.} Among inpainting models, Attentive Eraser~\cite{Sun2025} produces slightly better results compared to LaMa~\cite{Suvorov2022} ($+0.5$ p.p. on average).
For inpainting examples, see \Cref{apdx:foraug-samples}.

\textbf{Foreground size}
significantly impacts performance.
Employing a \emph{range} of sizes during recombination, rather than a fixed \emph{mean} size, boosts accuracy by approximately 1 p.p.
This suggests that the added variability is beneficial.

\textbf{Order of data augmentation.}
Applying all augmentations after foreground-background recombination (\emph{paste$\to$crop$\to$color}) improves ViT-S's performance compared to applying crop-related augmentations before pasting (\emph{crop$\to$paste$\to$color}).
ViT-Ti results are ambiguous.

\textbf{Background pruning.}
When it comes to the backgrounds to use, we test different pruning thresholds ($t_\text{prune}$) to exclude backgrounds with large inpainting.
A threshold of $t_\text{prune}=1.0$ means that we use all backgrounds that are not fully infilled.
Varying $t_\text{prune}$ has minimal impact.
We choose $t_\text{prune} = 0.8$ to exclude predominantly artificial backgrounds.

\textbf{Mixing} \schemename-augmented samples with the original ImageNet data proves crucial.
While constant and linear mixing schedules improve performance over no mixing by $2-3$ p.p. compared to only augmented samples, the cosine annealing schedule proves optimal, boosting accuracy by $3-4$ p.p.

\textbf{Edge smoothing.}
We evaluate the impact of using Gaussian blurring to smooth the edges of the foreground masks.
For larger models, this gives us a slight performance boost on the full ImageNet (second to last line in \Cref{tab:ablation-recombine}).

\textbf{Background strategy.}
Another point is the allowed choice of background image for each foreground object.
We compare using the original background, a background from the same class, and any background.
These strategies go from low diversity and high shared information content between the foreground and background to high diversity and low shared information content.
For \emph{ViT-Ti}, the latter two strategies perform comparably, while \emph{ViT-S} benefits from the added diversity of using any background.
The same is true when training on the full ImageNet.

\begin{table}
	\caption{\textbf{Position distribution ablation.} Accuracy of ViT-S on TinyImageNet (TIN) in percent using \schemename with different foreground position distributions by varying the Bates parameter $\eta$.
		The best performance is achieved when using the uniform distribution ($\eta=1$) for training.}
	\label{tbl:foreground-eta}
	\centering
	\small
	\begin{tabular}{ccccccc}
		\toprule
		\multirow{2.5}{*}{\makecell{Bates Parameter                  \\during training}} & \multirow{2.5}{*}{\makecell{TIN                                           \\w/o \schemename}} & \multicolumn{5}{c}{TIN w/ \schemename}                               \\
		\cmidrule(l){3-7}
		            &      & $\eta=-3$ & $-2$ & $1/-1$ & $2$  & $3$  \\
		\midrule
		Baseline    & 68.9 & 60.5      & 60.2 & 60.8   & 62.6 & 63.1 \\
		$\eta=-3$   & 71.3 & 79.3      & 79.5 & 79.1   & 79.3 & 79.1 \\
		$\eta=-2$   & 71.5 & 80.0      & 78.7 & 79.3   & 79.1 & 78.8 \\
		$\eta=1/-1$ & 72.3 & 79.5      & 78.9 & 80.2   & 79.7 & 80.4 \\
		$\eta=2$    & 71.3 & 78.2      & 77.8 & 79.1   & 79.6 & 79.9 \\
		$\eta=3$    & 71.4 & 77.2      & 76.9 & 78.6   & 79.6 & 79.7 \\
		\bottomrule
	\end{tabular}
\end{table}

\textbf{Foreground position.}
Finally, we analyze the foreground object's positioning in the image, using a
generalization of the Bates distribution~\cite{Bates1955} with parameter $\eta \in \Z$ (see \Cref{apdx:bates-distribution}).
The Bates distribution presents an easy way to sample from a bounded domain with just one hyperparameter that controls its concentration.
$\eta = 1/-1$ corresponds to the uniform distribution; $\eta > 1$ concentrates the distribution around the center; and for $\eta < -1$, the distribution is concentrated at the borders (see \Cref{apdx:bates-distribution} for details).
When sampling more towards the center of the image, the difficulty of the task is reduced, which reduces performance on TinyImageNet (\Cref{tbl:foreground-eta}).
This is reflected in the performance when evaluating using \schemename with $\eta=2$ and $\eta=3$ compared to $\eta=-1/1$.
We observe a similar reduction for $\eta < -1$.

\begin{table}[t]
	\caption{\textbf{\schemename dataset statistics.} Dataset statistics for TinyImageNet and ImageNet with and without \schemename. For \schemename we report the number of foreground/background pairs.}
	\label{tab:dataset-stats}
	\centering
	\begin{tabular}{l S[table-format=4.0] S[table-format=7.0] S[table-format=5.0]}
		\toprule
		Dataset                    & {Classes} & {\makecell{Training         \\ Images}} & {\makecell{Validation \\ Images}} \\
		\midrule
		TinyImageNet               & 200       & 100000              & 10000 \\
		TinyImageNet + \schemename & 200       & 99404               & 9915  \\
		ImageNet                   & 1000      & 1281167             & 50000 \\
		ImageNet + \schemename     & 1000      & 1274557             & 49751 \\
		\bottomrule
	\end{tabular}
\end{table}
After fixing the optimal design parameters in \Cref{tab:ablation-segment,tab:ablation-recombine} (last rows), we run \schemename's segmentation step on the entire ImageNet dataset.
\Cref{tab:dataset-stats} shows the resulting dataset statistics.
The slightly reduced image count for \schemename is due to instances where Grounded SAM fails to produce valid segmentation masks.

\section{Robustness Evaluation on Corner-Cases}
\label{apdx:corner-cases}
\begin{table}[t]
	\centering
	\caption{\textbf{Evaluation on Corner-Cases.} Objects cut from ImageNet evaluation bounding boxes are pasted onto infilled backgrounds. Objects have three sizes: $56$px, $84$px, and $112$px. Objects are places in the center (CeX) or corner (CoX) of an image its original background (XxO) or a random background (XxR).}
	\label{tab:corner-cases}
	\resizebox{\textwidth}{!}{
		\begin{tabular}{lcccccccccccccc}
			\toprule
			\multirow{4}{*}{Model} & \multirow{4}{*}{w/ \schemename} & \multicolumn{12}{c}{Corner Cases Accuracy [\%]}                                                                                                                                                                                                                               \\
			\cmidrule(l){3-14}
			                       &                                 & \multicolumn{4}{c}{56}                          & \multicolumn{4}{c}{84} & \multicolumn{4}{c}{112}                                                                                                                                                                            \\
			\cmidrule(lr){3-6} \cmidrule(lr){7-10} \cmidrule(l){11-14}
			                       &                                 & CeO                                             & CoO                    & CeR                     & CoR              & CeO              & CoO              & CeR              & CoR              & CeO              & CoO              & CeR              & CoR              \\
			\midrule
			ViT-S                  & \xmark                          & $40.5 \pm 2.0$                                  & $28.6 \pm 0.8$         & $10.3 \pm 0.9$          & $6.4 \pm 0.2$    & $56.8 \pm 1.2$   & $47.6 \pm 1.0$   & $31.3 \pm 0.7$   & $25.5 \pm 0.5$   & $70.9 \pm 0.1$   & $66.9 \pm 1.6$   & $55.2 \pm 0.2$   & $51.1 \pm 0.8$   \\
			ViT-S                  & \cmark                          & $49.4 \pm 0.6$                                  & $39.9 \pm 0.5$         & $22.7 \pm 0.4$          & $17.6 \pm 0.3$   & $66.3 \pm 0.3$   & $60.0 \pm 0.3$   & $47.7 \pm 0.7$   & $43.2 \pm 0.2$   & $76.5 \pm 0.2$   & $74.9 \pm 0.4$   & $66.8 \pm 0.6$   & $64.9 \pm 0.1$   \\
			                       &                                 & \grntxt{$+8.9$}                                 & \grntxt{$+11.3$}       & \grntxt{$+12.4$}        & \grntxt{$+11.2$} & \grntxt{$+9.4$}  & \grntxt{$+12.4$} & \grntxt{$+16.4$} & \grntxt{$+17.7$} & \grntxt{$+5.6$}  & \grntxt{$+8.0$}  & \grntxt{$+11.6$} & \grntxt{$+13.7$} \\
			\cmidrule(r){1-2}
			ViT-B                  & \xmark                          & $37.9 \pm 1.4$                                  & $29.3 \pm 0.7$         & $14.0 \pm 1.7$          & $11.9 \pm 1.1$   & $51.5 \pm 0.7$   & $45.0 \pm 0.8$   & $27.3 \pm 0.8$   & $26.3 \pm 0.8$   & $64.7 \pm 0.3$   & $61.8 \pm 0.6$   & $46.3 \pm 0.3$   & $45.5 \pm 0.5$   \\
			ViT-B                  & \cmark                          & $50.4 \pm 0.8$                                  & $42.4 \pm 0.6$         & $26.5 \pm 0.6$          & $22.8 \pm 0.8$   & $65.3 \pm 0.9$   & $60.9 \pm 0.6$   & $47.6 \pm 0.3$   & $45.6 \pm 0.1$   & $75.7 \pm 0.6$   & $74.0 \pm 0.6$   & $65.7 \pm 0.7$   & $64.3 \pm 0.5$   \\
			                       &                                 & \grntxt{$+12.5$}                                & \grntxt{$+13.1$}       & \grntxt{$+12.4$}        & \grntxt{$+10.9$} & \grntxt{$+13.8$} & \grntxt{$+15.9$} & \grntxt{$+20.2$} & \grntxt{$+19.3$} & \grntxt{$+11.0$} & \grntxt{$+12.2$} & \grntxt{$+19.3$} & \grntxt{$+18.8$} \\
			\cmidrule(r){1-2}
			ViT-L                  & \xmark                          & $32.8 \pm 1.6$                                  & $24.8 \pm 1.1$         & $14.8 \pm 2.2$          & $9.7 \pm 1.2$    & $42.7 \pm 0.9$   & $33.8 \pm 0.7$   & $21.3 \pm 1.5$   & $16.3 \pm 1.0$   & $55.7 \pm 0.7$   & $49.7 \pm 0.7$   & $36.0 \pm 1.3$   & $32.5 \pm 0.9$   \\
			ViT-L                  & \cmark                          & $45.7 \pm 0.6$                                  & $39.0 \pm 0.5$         & $25.6 \pm 0.6$          & $24.1 \pm 0.8$   & $59.1 \pm 0.3$   & $55.2 \pm 0.4$   & $41.9 \pm 1.0$   & $42.7 \pm 0.6$   & $71.4 \pm 0.3$   & $69.0 \pm 0.4$   & $60.7 \pm 1.0$   & $60.3 \pm 0.8$   \\
			                       &                                 & \grntxt{$+12.9$}                                & \grntxt{$+14.2$}       & \grntxt{$+10.8$}        & \grntxt{$+14.4$} & \grntxt{$+16.3$} & \grntxt{$+21.5$} & \grntxt{$+20.5$} & \grntxt{$+26.4$} & \grntxt{$+15.7$} & \grntxt{$+19.3$} & \grntxt{$+24.7$} & \grntxt{$+27.8$} \\
			\cmidrule(r){1-2}
			DeiT-S                 & \xmark                          & $46.3 \pm 0.7$                                  & $38.1 \pm 0.3$         & $13.1 \pm 0.5$          & $9.9 \pm 0.1$    & $62.8 \pm 0.4$   & $58.2 \pm 0.2$   & $37.1 \pm 0.7$   & $34.3 \pm 0.5$   & $73.3 \pm 0.2$   & $73.9 \pm 0.4$   & $58.8 \pm 0.4$   & $59.4 \pm 0.6$   \\
			DeiT-S                 & \cmark                          & $44.7 \pm 1.4$                                  & $37.1 \pm 1.4$         & $15.6 \pm 1.3$          & $12.1 \pm 0.9$   & $62.1 \pm 1.2$   & $57.8 \pm 1.1$   & $41.6 \pm 1.1$   & $37.9 \pm 1.2$   & $73.2 \pm 0.7$   & $73.3 \pm 0.4$   & $62.3 \pm 0.7$   & $61.4 \pm 0.9$   \\
			                       &                                 & \rdtxt{$-1.6$}                                  & \rdtxt{$-1.1$}         & \grntxt{$+2.4$}         & \grntxt{$+2.2$}  & \rdtxt{$-0.7$}   & \rdtxt{$-0.4$}   & \grntxt{$+4.4$}  & \grntxt{$+3.5$}  & \gtxt{$-0.1$}    & \rdtxt{$-0.6$}   & \grntxt{$+3.5$}  & \grntxt{$+2.0$}  \\
			\cmidrule(r){1-2}
			DeiT-B                 & \xmark                          & $48.1 \pm 0.9$                                  & $40.4 \pm 2.0$         & $15.8 \pm 0.2$          & $12.9 \pm 0.6$   & $64.0 \pm 0.9$   & $59.5 \pm 1.3$   & $39.0 \pm 0.9$   & $37.2 \pm 0.8$   & $74.1 \pm 0.7$   & $74.8 \pm 0.7$   & $59.1 \pm 0.8$   & $60.0 \pm 0.6$   \\
			DeiT-B                 & \cmark                          & $50.7 \pm 0.1$                                  & $44.0 \pm 0.4$         & $19.3 \pm 0.2$          & $16.3 \pm 0.2$   & $66.0 \pm 0.2$   & $62.0 \pm 0.3$   & $43.4 \pm 0.3$   & $40.9 \pm 0.4$   & $75.4 \pm 0.1$   & $76.4 \pm 0.3$   & $62.8 \pm 0.2$   & $63.9 \pm 0.2$   \\
			                       &                                 & \grntxt{$+2.6$}                                 & \grntxt{$+3.6$}        & \grntxt{$+3.5$}         & \grntxt{$+3.5$}  & \grntxt{$+2.0$}  & \grntxt{$+2.5$}  & \grntxt{$+4.4$}  & \grntxt{$+3.8$}  & \grntxt{$+1.3$}  & \grntxt{$+1.6$}  & \grntxt{$+3.8$}  & \grntxt{$+3.9$}  \\
			\cmidrule(r){1-2}
			DeiT-L                 & \xmark                          & $39.2 \pm 2.6$                                  & $32.6 \pm 1.5$         & $10.5 \pm 2.8$          & $9.1 \pm 2.3$    & $55.7 \pm 2.5$   & $51.0 \pm 2.7$   & $30.3 \pm 4.0$   & $29.5 \pm 3.9$   & $68.5 \pm 2.1$   & $68.1 \pm 1.7$   & $51.7 \pm 3.1$   & $52.1 \pm 2.7$   \\
			DeiT-L                 & \cmark                          & $51.9 \pm 0.7$                                  & $46.6 \pm 0.5$         & $21.5 \pm 1.3$          & $19.0 \pm 1.2$   & $66.6 \pm 0.6$   & $64.1 \pm 0.7$   & $45.3 \pm 1.3$   & $43.6 \pm 1.1$   & $75.6 \pm 0.4$   & $77.3 \pm 0.4$   & $63.8 \pm 0.8$   & $65.4 \pm 0.6$   \\
			                       &                                 & \grntxt{$+12.8$}                                & \grntxt{$+14.0$}       & \grntxt{$+11.0$}        & \grntxt{$+9.9$}  & \grntxt{$+11.0$} & \grntxt{$+13.1$} & \grntxt{$+15.0$} & \grntxt{$+14.1$} & \grntxt{$+7.1$}  & \grntxt{$+9.2$}  & \grntxt{$+12.1$} & \grntxt{$+13.4$} \\
			\cmidrule(r){1-2}
			Swin-Ti                & \xmark                          & $41.2 \pm 1.8$                                  & $32.5 \pm 0.3$         & $17.4 \pm 2.6$          & $12.2 \pm 0.2$   & $60.0 \pm 1.6$   & $51.4 \pm 0.2$   & $39.6 \pm 2.6$   & $34.8 \pm 0.9$   & $71.7 \pm 0.8$   & $66.1 \pm 0.7$   & $58.2 \pm 1.1$   & $53.6 \pm 1.2$   \\
			Swin-Ti                & \cmark                          & $49.8 \pm 0.6$                                  & $42.8 \pm 0.7$         & $24.2 \pm 0.7$          & $21.4 \pm 0.9$   & $66.4 \pm 0.6$   & $60.5 \pm 0.2$   & $47.8 \pm 0.5$   & $44.6 \pm 0.5$   & $76.0 \pm 0.3$   & $72.7 \pm 0.2$   & $65.7 \pm 0.5$   & $62.1 \pm 0.3$   \\
			                       &                                 & \grntxt{$+8.5$}                                 & \grntxt{$+10.3$}       & \grntxt{$+6.8$}         & \grntxt{$+9.2$}  & \grntxt{$+6.4$}  & \grntxt{$+9.2$}  & \grntxt{$+8.2$}  & \grntxt{$+9.8$}  & \grntxt{$+4.3$}  & \grntxt{$+6.5$}  & \grntxt{$+7.5$}  & \grntxt{$+8.5$}  \\
			\cmidrule(r){1-2}
			Swin-S                 & \xmark                          & $41.3 \pm 0.6$                                  & $33.0 \pm 0.1$         & $18.4 \pm 0.7$          & $13.3 \pm 0.5$   & $59.2 \pm 0.1$   & $51.2 \pm 0.5$   & $39.1 \pm 0.2$   & $35.9 \pm 0.3$   & $71.5 \pm 0.2$   & $65.6 \pm 0.1$   & $56.8 \pm 0.5$   & $53.2 \pm 0.2$   \\
			Swin-S                 & \cmark                          & $48.6 \pm 0.7$                                  & $39.9 \pm 1.6$         & $22.2 \pm 0.9$          & $16.8 \pm 1.1$   & $64.4 \pm 0.9$   & $57.9 \pm 1.5$   & $43.8 \pm 1.1$   & $42.3 \pm 1.0$   & $75.7 \pm 0.2$   & $71.8 \pm 0.8$   & $63.2 \pm 0.4$   & $60.6 \pm 0.6$   \\
			                       &                                 & \grntxt{$+7.3$}                                 & \grntxt{$+7.0$}        & \grntxt{$+3.8$}         & \grntxt{$+3.6$}  & \grntxt{$+5.1$}  & \grntxt{$+6.7$}  & \grntxt{$+4.7$}  & \grntxt{$+6.4$}  & \grntxt{$+4.2$}  & \grntxt{$+6.2$}  & \grntxt{$+6.4$}  & \grntxt{$+7.4$}  \\
			\cmidrule(r){1-2}
			ResNet50               & \xmark                          & $48.6 \pm 0.6$                                  & $35.1 \pm 0.4$         & $23.0 \pm 0.7$          & $13.0 \pm 0.3$   & $65.8 \pm 0.4$   & $58.2 \pm 0.3$   & $44.4 \pm 0.6$   & $38.1 \pm 0.5$   & $73.2 \pm 0.2$   & $69.9 \pm 0.2$   & $56.9 \pm 0.1$   & $56.9 \pm 0.1$   \\
			ResNet50               & \cmark                          & $52.3 \pm 0.6$                                  & $39.5 \pm 0.1$         & $27.4 \pm 0.6$          & $17.6 \pm 0.1$   & $68.5 \pm 0.3$   & $61.9 \pm 0.1$   & $48.5 \pm 0.4$   & $43.7 \pm 0.3$   & $75.2 \pm 0.1$   & $72.4 \pm 0.1$   & $61.7 \pm 0.3$   & $61.7 \pm 0.3$   \\
			                       &                                 & \grntxt{$+3.7$}                                 & \grntxt{$+4.4$}        & \grntxt{$+4.4$}         & \grntxt{$+4.6$}  & \grntxt{$+2.8$}  & \grntxt{$+3.8$}  & \grntxt{$+4.2$}  & \grntxt{$+5.5$}  & \grntxt{$+2.0$}  & \grntxt{$+2.5$}  & \grntxt{$+4.8$}  & \grntxt{$+4.8$}  \\
			\cmidrule(r){1-2}
			ResNet101              & \xmark                          & $47.8 \pm 0.7$                                  & $37.2 \pm 0.5$         & $20.4 \pm 1.2$          & $14.2 \pm 0.3$   & $64.9 \pm 0.2$   & $58.6 \pm 0.5$   & $41.1 \pm 0.5$   & $38.3 \pm 0.7$   & $73.6 \pm 0.3$   & $70.5 \pm 0.3$   & $56.2 \pm 0.4$   & $57.0 \pm 0.5$   \\
			ResNet101              & \cmark                          & $52.3 \pm 0.1$                                  & $42.2 \pm 0.1$         & $24.7 \pm 0.1$          & $19.2 \pm 0.4$   & $68.8 \pm 0.6$   & $62.9 \pm 0.3$   & $46.4 \pm 1.5$   & $44.3 \pm 0.9$   & $76.0 \pm 0.4$   & $73.7 \pm 0.3$   & $61.0 \pm 1.2$   & $62.6 \pm 0.5$   \\
			                       &                                 & \grntxt{$+4.4$}                                 & \grntxt{$+5.0$}        & \grntxt{$+4.3$}         & \grntxt{$+5.0$}  & \grntxt{$+3.9$}  & \grntxt{$+4.3$}  & \grntxt{$+5.3$}  & \grntxt{$+6.0$}  & \grntxt{$+2.4$}  & \grntxt{$+3.2$}  & \grntxt{$+4.7$}  & \grntxt{$+5.7$}  \\
			\bottomrule
		\end{tabular}
	}
\end{table}

\Cref{tab:corner-cases} reports accuracy on the corner-cases dataset~\cite{Fatima2025} for models trained with and without \schemename.
The dataset is constructed by pasting objects cropped by their full bounding boxes (which are available for the ImageNet validation set) onto 224$\times$224 infilled backgrounds.
The dataset has three factors: foreground size (56, 84, 112 pixels), spatial position (center, CeX, vs.\ corner, CoX), and background type (original image background, XxO, vs.\ a random background, XxR), yielding $3 \times 2 \times 2$ controlled configurations per model.

Across all architectures, training with \schemename consistently improves robustness to these composition shifts.
For ViT-S/B/L, gains range from roughly $+8$ to over $+27$ percentage points, with the largest improvements occurring in the most challenging settings with foregrounds placed in corners on random backgrounds (e.g., CoR and CeR).
Swin and ResNet models also benefit across all configurations, with increases typically between $+3$ and $+10$ points.
DeiT-S shows small drops on some same-background center cases (CeO/CoO), but still improves notably on random-background conditions (XxR), while DeiT-B/L gain across nearly all settings.

Three trends are apparent.
First, all baselines perform substantially worse when moving from original to random backgrounds and from centered to corner placements, indicating strong background and center biases.
Second, \schemename reduces this sensitivity: the absolute gap between center and corner, and between original and random backgrounds, shrinks for almost all models and sizes.
Third, the relative improvements are especially pronounced for smaller objects and off-center placements, suggesting that \schemename makes models more foreground-focused and less reliant on canonical object scale and position.

\section{\schemename Segmentation Samples}
\label{apdx:segmentation-examples}
\begin{figure}[t!]
	\centering
	\begin{subfigure}{.49\textwidth}

		\includegraphics[width=\textwidth]{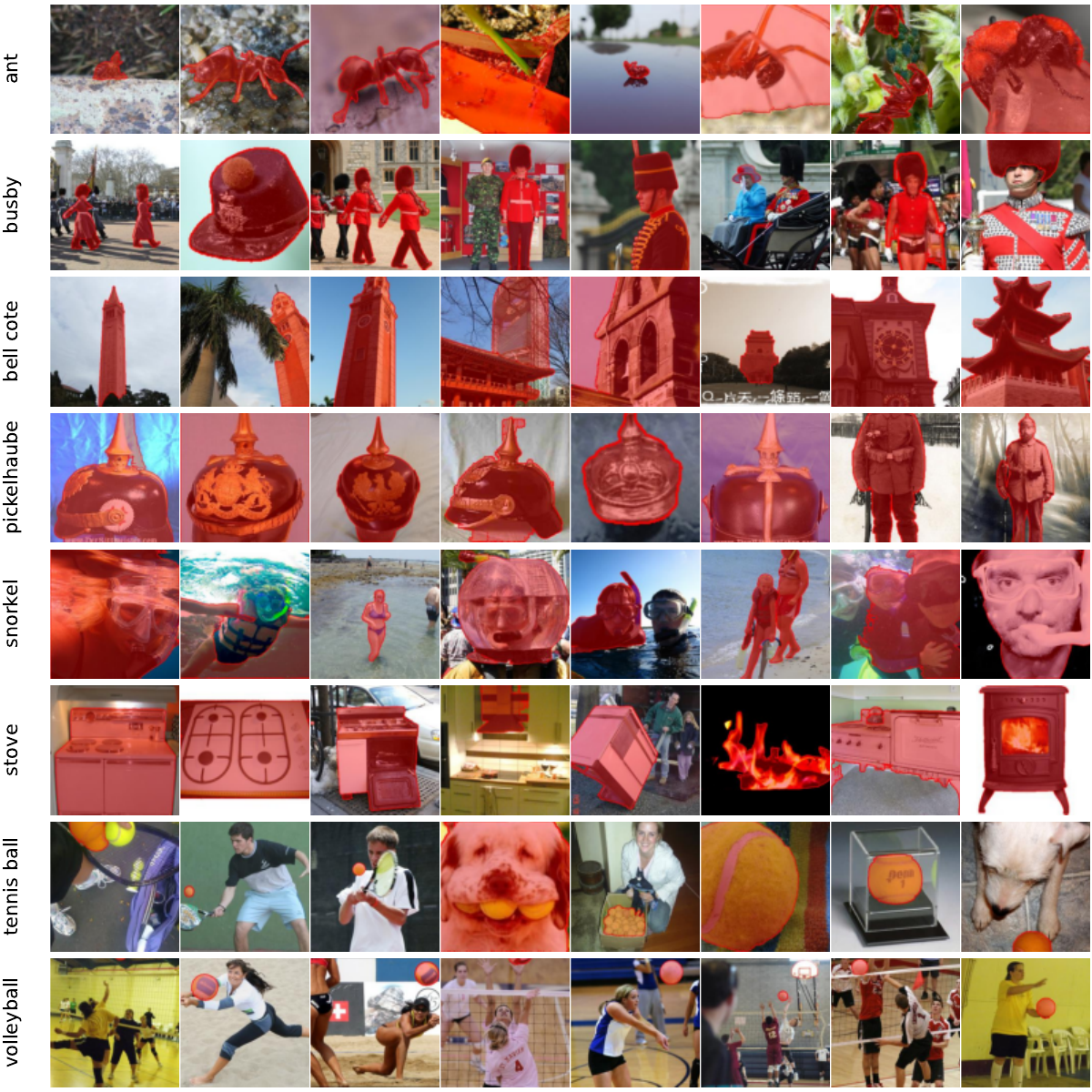}
	\end{subfigure}
	\hfill
	\begin{subfigure}{.49\textwidth}

		\includegraphics[width=\textwidth]{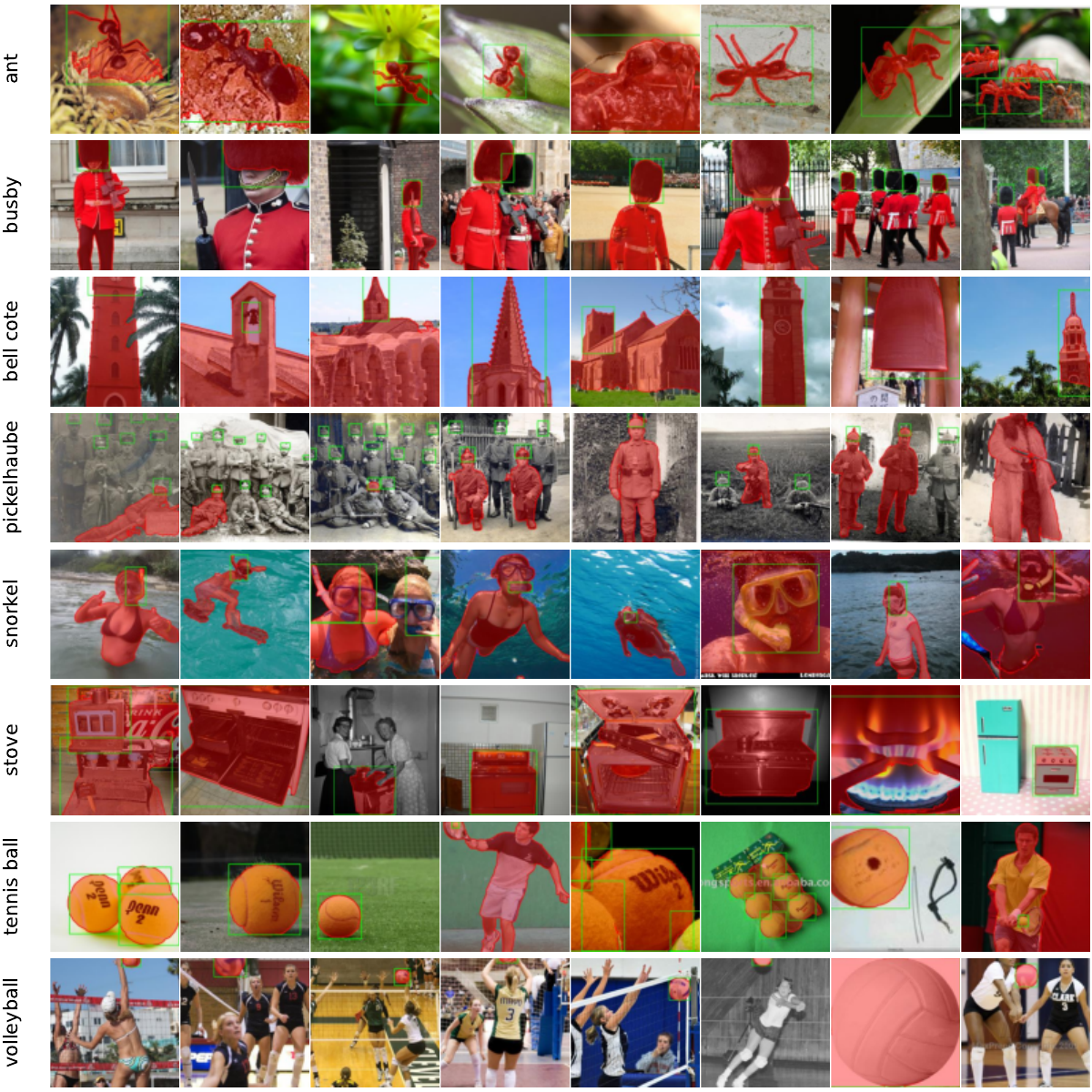}
	\end{subfigure}
	\caption{\textbf{Segmentation mask samples.} ImageNet training samples (left) and validation samples (right) of our segmentation masks with annotated bounding boxes.}
	\label{fig:mask-examples}
\end{figure}
We show examples of the automatically generated segmentation masks for a diverse subset of object categories (``ant,'' ``busby,'' ``bell cote,'' ``pickelhaube,'' ``snorkel,'' ``stove,'' ``tennis ``ball,'' and ``volleyball'').
Note that ``busby,'' ``bell cote,'' ``pickelhaube,'' and ``snorkel'' are the four classes with the \textbf{worst} mean box precision and box-to-box IoU on the validation set.
\Cref{fig:mask-examples} (right) illustrates masks from the evaluation split, while \Cref{fig:mask-examples} (left) shows examples from the training split.
Across both sets, the masks accurately isolate foreground objects with clean boundaries, despite large variations in object scale, shape, and appearance, supporting their use for background removal and resampling in our training pipeline.
We find that the main failure cases are:
(\textit{i}) When the ground-truth annotation corresponds to only a part of an object, the predicted mask often expands to cover the entire object rather than the annotated region.
See for example ``busby'' or ``bell cote''.
(\textit{ii}) In images containing multiple instances, some objects may be missed, resulting in incomplete foreground coverage.
This is especially visible for ``busby'' and ``pickelhaube''.
However, note that especially for ``pickelhaube'' the training distribution is noticeably different from the validation distribution, showing many images with just the head instead of groups of people wearing it.
(\textit{iii}) In rare cases, the predicted mask degenerates and covers nearly the entire image, effectively eliminating the background.
This happens in $<10\%$ of all training images, and we do not use the resulting backgrounds for recombination (see \Cref{apdx:infill-ratio}).

\section{\schemename Sample Images}
\label{apdx:foraug-samples}
\begin{table*}[t!]
	\centering
	\caption{\textbf{Recombination samples.} Sample Images from using \schemename on ImageNet.}
	\label{tbl:example-images}
	\resizebox{.93\textwidth}{!}{
		\begin{tabular}{ccccc}
			\toprule
			Class & \makecell{Original \\Image} & \makecell{Extracted \\Foreground} & \makecell{Infilled \\Background} & \schemename's Recombinations \\
			\midrule
			\makecell{n01531178        \\Goldfinch} & \includegraphics[max width=.1\columnwidth, max height=2cm, valign=c]{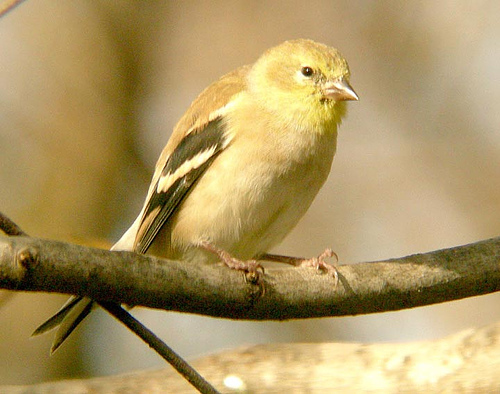} & \includegraphics[max width=.1\columnwidth, max height=2cm, valign=c]{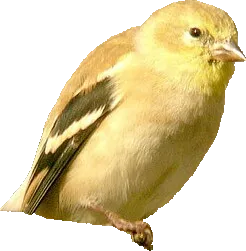} & \includegraphics[max width=.1\columnwidth, max height=2cm, valign=c]{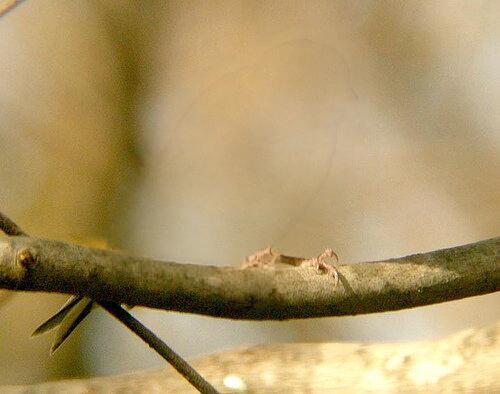} & \includegraphics[max width=.1\columnwidth, max height=2cm, valign=c]{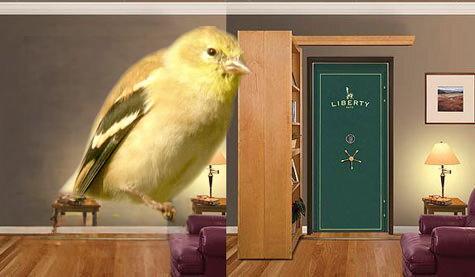}  \includegraphics[max width=.1\columnwidth, max height=2cm, valign=c]{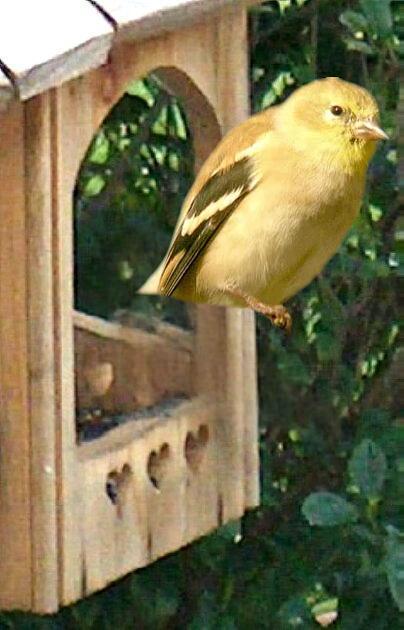}  \includegraphics[max width=.1\columnwidth, max height=2cm, valign=c]{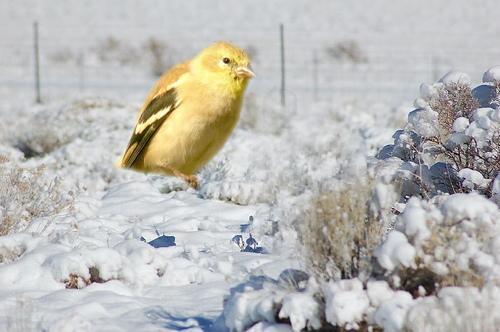}  \includegraphics[max width=.1\columnwidth, max height=2cm, valign=c]{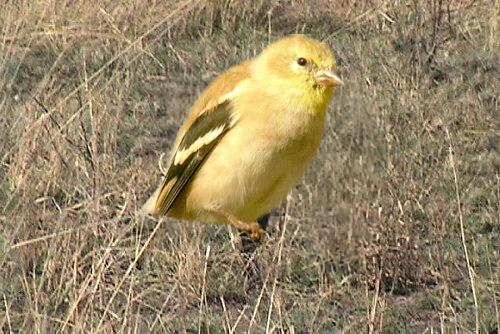}  \includegraphics[max width=.1\columnwidth, max height=2cm, valign=c]{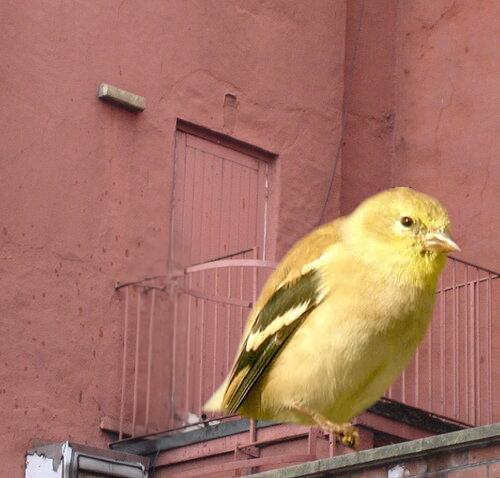}  \includegraphics[max width=.1\columnwidth, max height=2cm, valign=c]{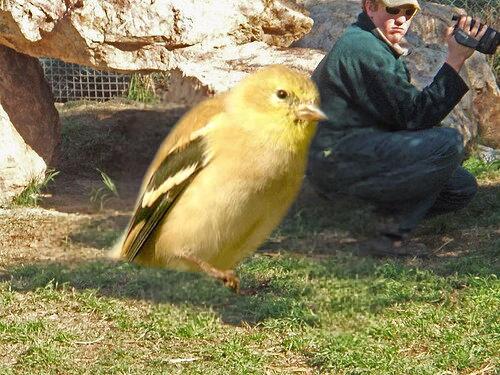} \\
			\makecell{n01818515        \\Macaw}  &  \includegraphics[max width=.1\columnwidth, max height=2cm, valign=c]{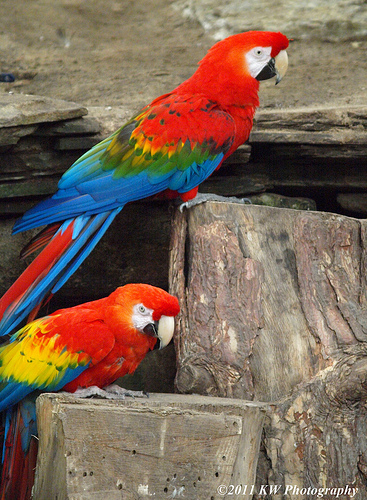} & \includegraphics[max width=.1\columnwidth, max height=2cm, valign=c]{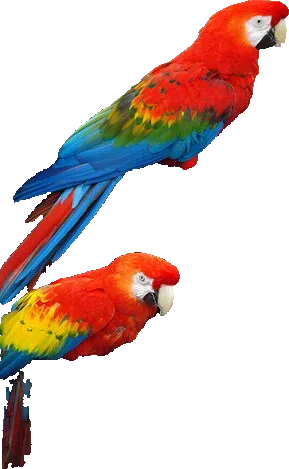} & \includegraphics[max width=.1\columnwidth, max height=2cm, valign=c]{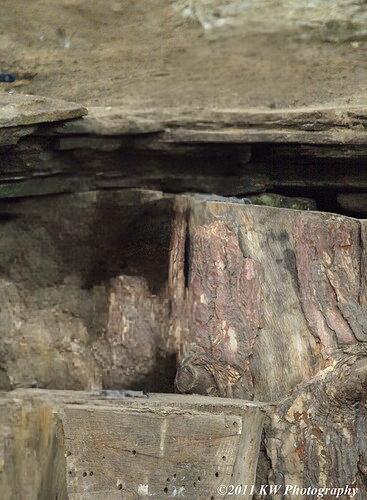} & \includegraphics[max width=.1\columnwidth, max height=2cm, valign=c]{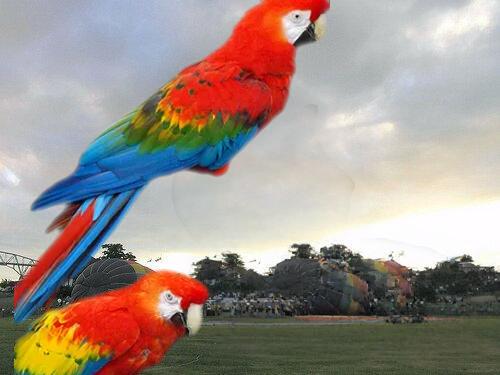} \includegraphics[max width=.1\columnwidth, max height=2cm, valign=c]{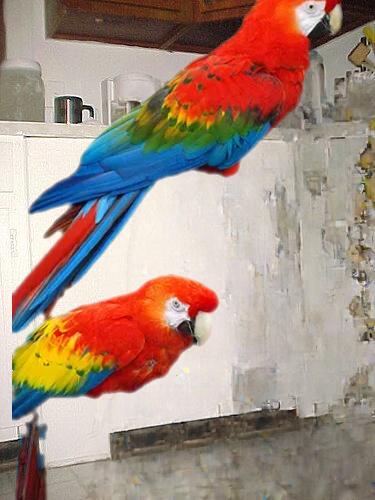} \includegraphics[max width=.1\columnwidth, max height=2cm, valign=c]{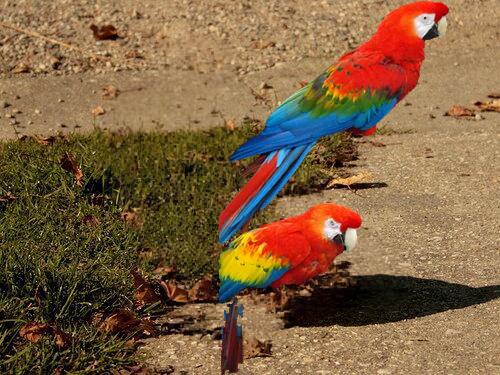} \includegraphics[max width=.1\columnwidth, max height=2cm, valign=c]{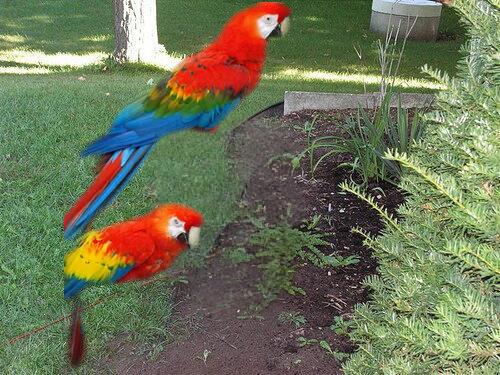} \includegraphics[max width=.1\columnwidth, max height=2cm, valign=c]{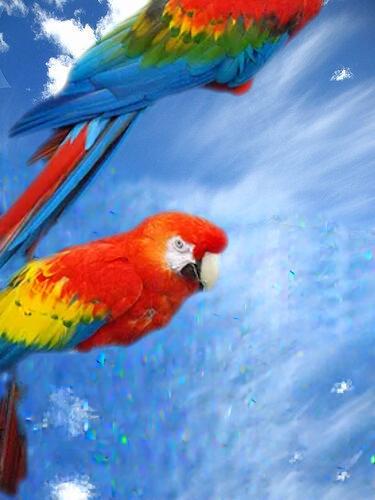} \includegraphics[max width=.1\columnwidth, max height=2cm, valign=c]{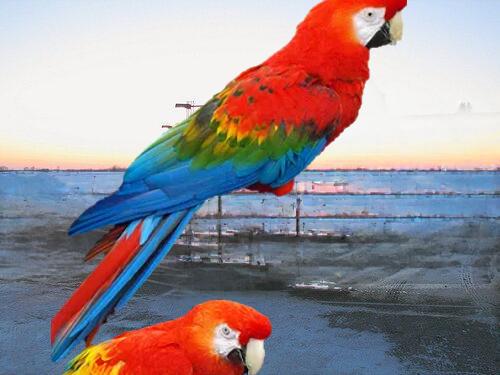} \\
			\makecell{n01943899        \\Conch} & \includegraphics[max width=.1\columnwidth, max height=2cm, valign=c]{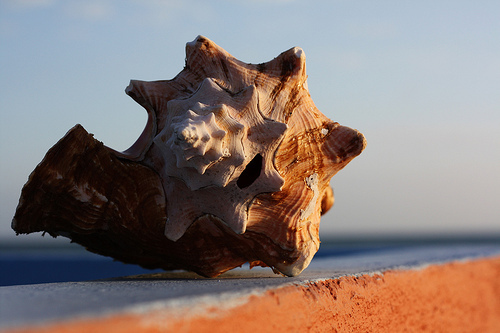} & \includegraphics[max width=.1\columnwidth, max height=2cm, valign=c]{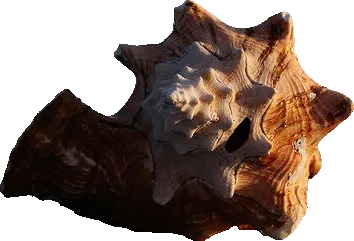} & \includegraphics[max width=.1\columnwidth, max height=2cm, valign=c]{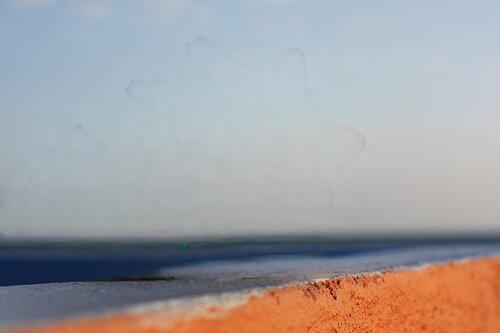} & \includegraphics[max width=.1\columnwidth, max height=2cm, valign=c]{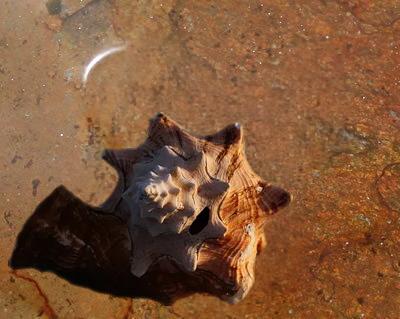} \includegraphics[max width=.1\columnwidth, max height=2cm, valign=c]{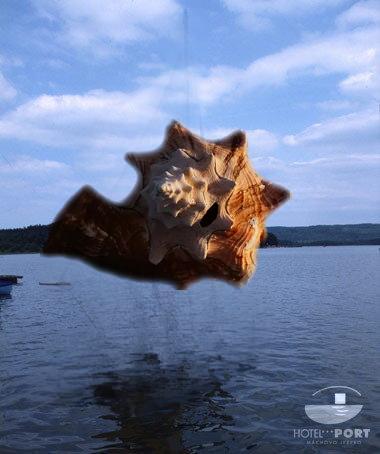} \includegraphics[max width=.1\columnwidth, max height=2cm, valign=c]{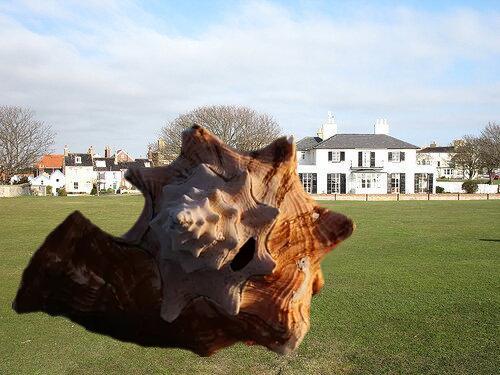} \includegraphics[max width=.1\columnwidth, max height=2cm, valign=c]{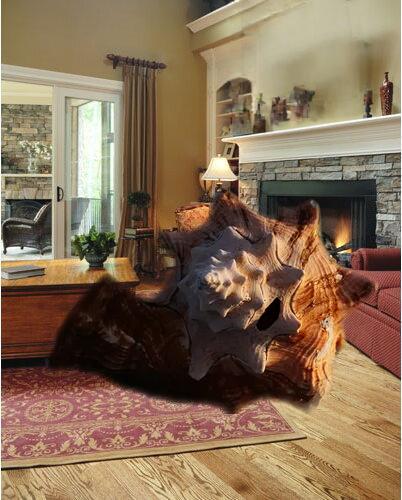} \includegraphics[max width=.1\columnwidth, max height=2cm, valign=c]{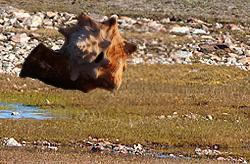} \includegraphics[max width=.1\columnwidth, max height=2cm, valign=c]{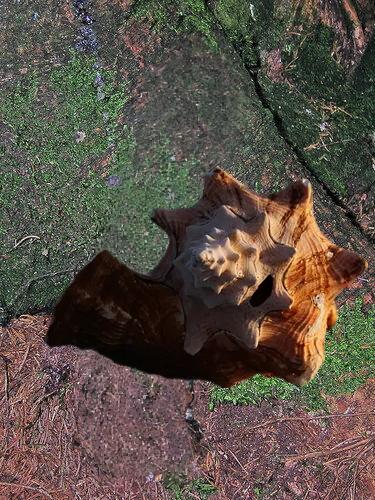} \\
			\makecell{n01986214        \\ Hermit Crab} & \includegraphics[max width=.1\columnwidth, max height=2cm, valign=c]{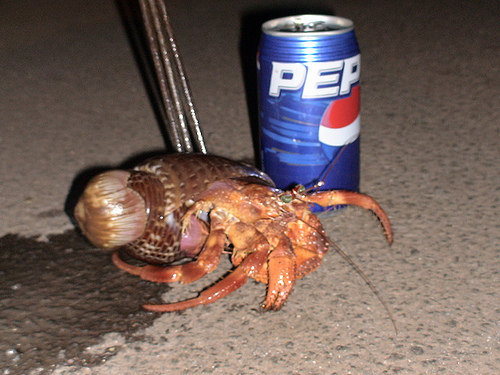} & \includegraphics[max width=.1\columnwidth, max height=2cm, valign=c]{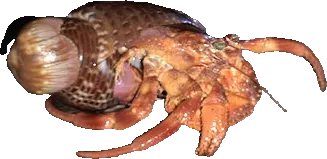} & \includegraphics[max width=.1\columnwidth, max height=2cm, valign=c]{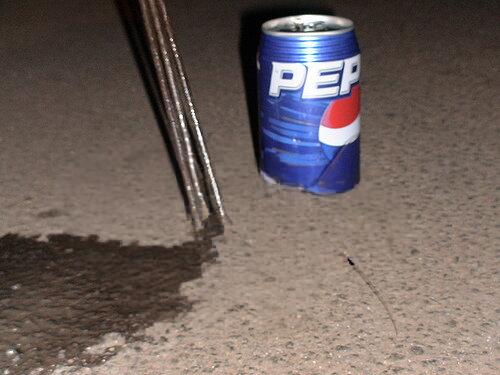} & \includegraphics[max width=.1\columnwidth, max height=2cm, valign=c]{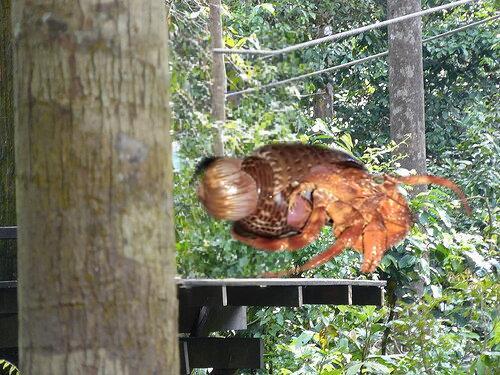} \includegraphics[max width=.1\columnwidth, max height=2cm, valign=c]{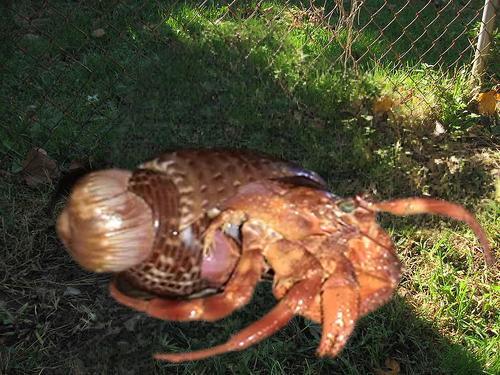} \includegraphics[max width=.1\columnwidth, max height=2cm, valign=c]{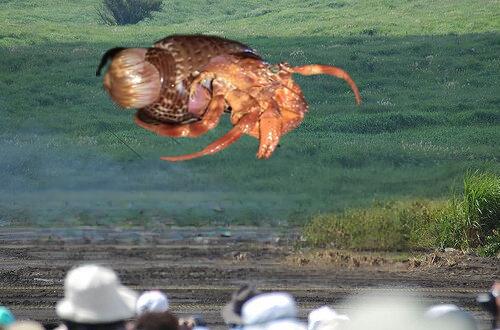} \includegraphics[max width=.1\columnwidth, max height=2cm, valign=c]{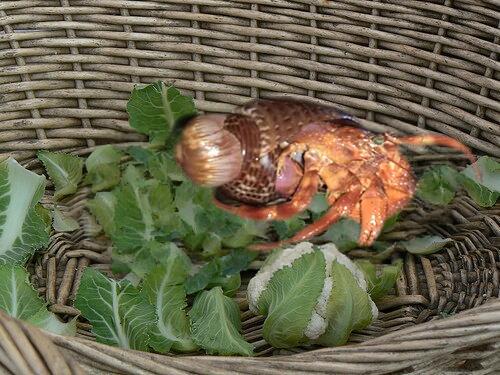} \includegraphics[max width=.1\columnwidth, max height=2cm, valign=c]{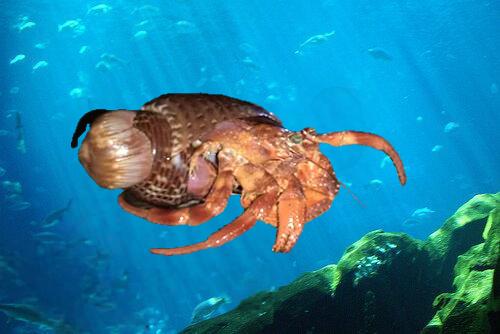} \includegraphics[max width=.1\columnwidth, max height=2cm, valign=c]{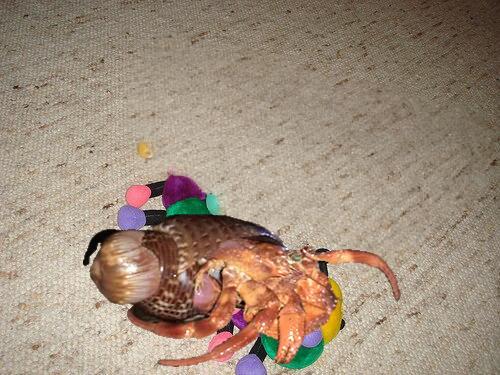} \\
			\makecell{n02190166        \\Fly} & \includegraphics[max width=.1\columnwidth, max height=2cm, valign=c]{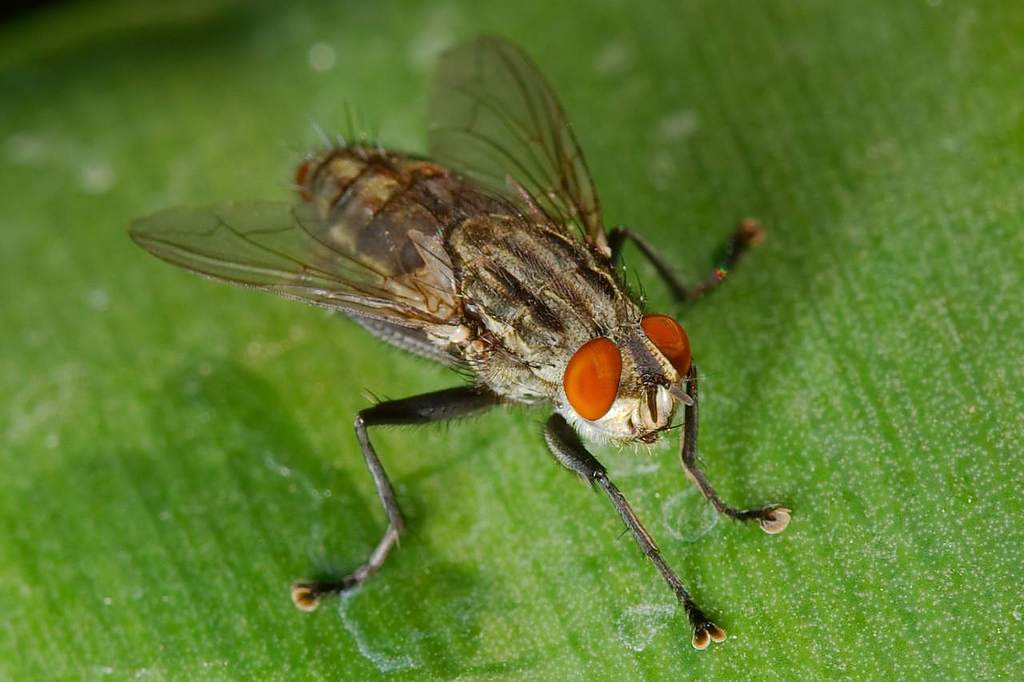} & \includegraphics[max width=.1\columnwidth, max height=2cm, valign=c]{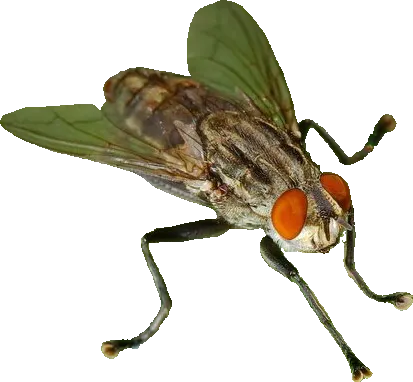} & \includegraphics[max width=.1\columnwidth, max height=2cm, valign=c]{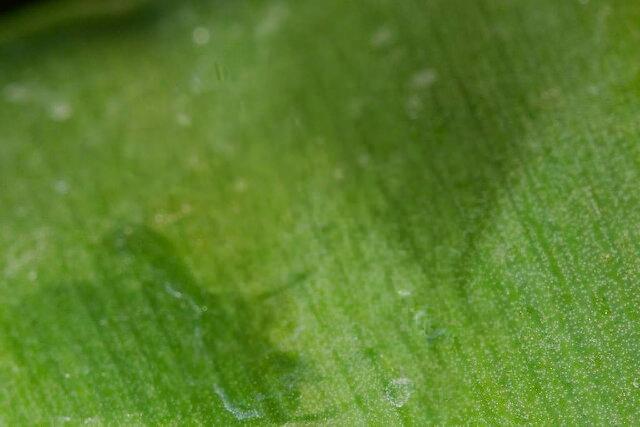} & \includegraphics[max width=.1\columnwidth, max height=2cm, valign=c]{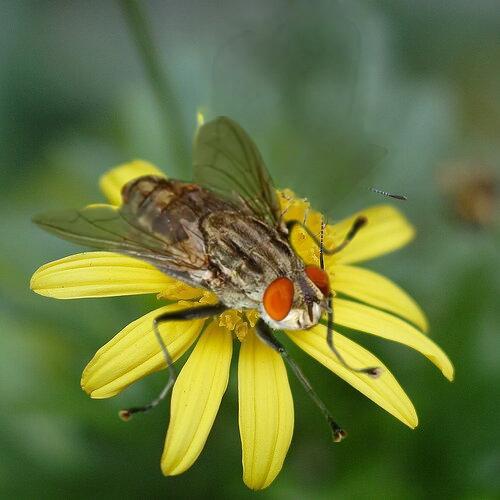} \includegraphics[max width=.1\columnwidth, max height=2cm, valign=c]{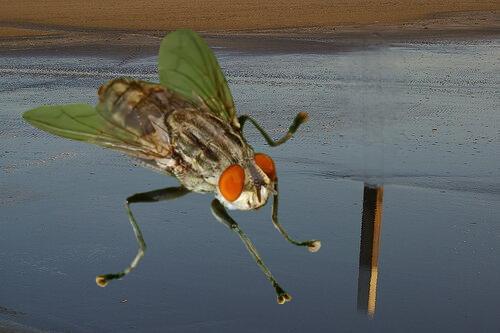} \includegraphics[max width=.1\columnwidth, max height=2cm, valign=c]{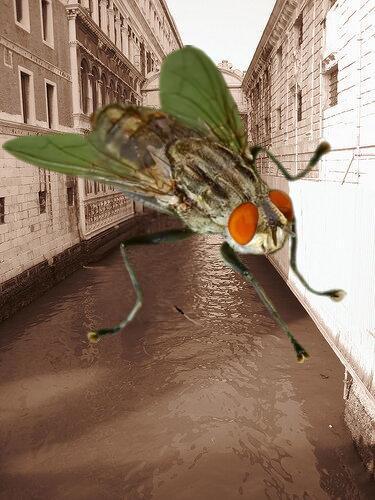} \includegraphics[max width=.1\columnwidth, max height=2cm, valign=c]{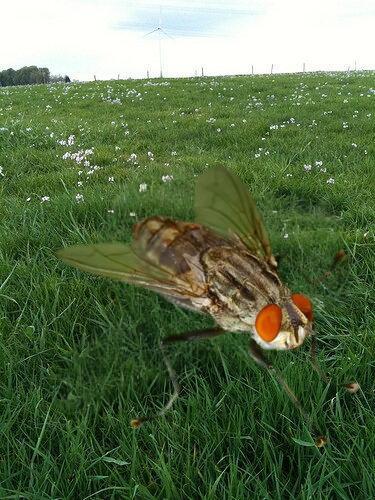} \includegraphics[max width=.1\columnwidth, max height=2cm, valign=c]{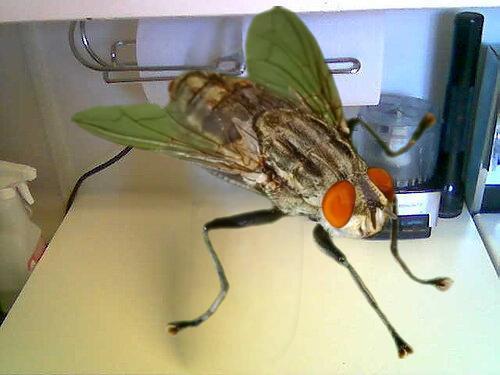} \includegraphics[max width=.1\columnwidth, max height=2cm, valign=c]{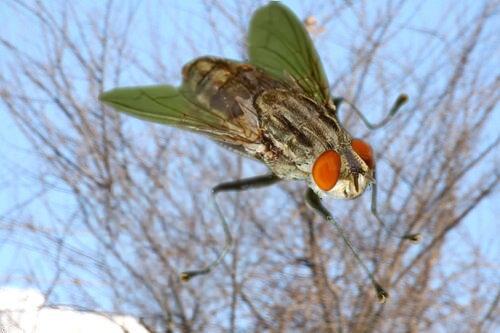} \\
			\makecell{n02229544        \\Cricket} & \includegraphics[max width=.1\columnwidth, max height=2cm, valign=c]{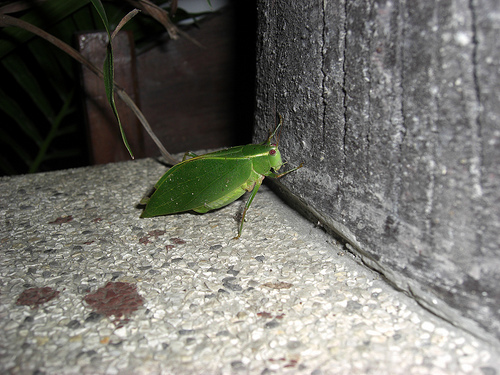} & \includegraphics[max width=.1\columnwidth, max height=2cm, valign=c]{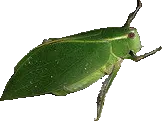} & \includegraphics[max width=.1\columnwidth, max height=2cm, valign=c]{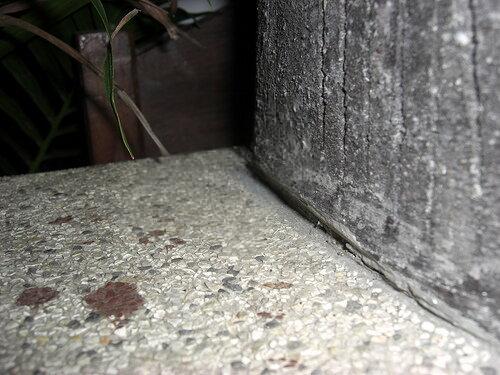} & \includegraphics[max width=.1\columnwidth, max height=2cm, valign=c]{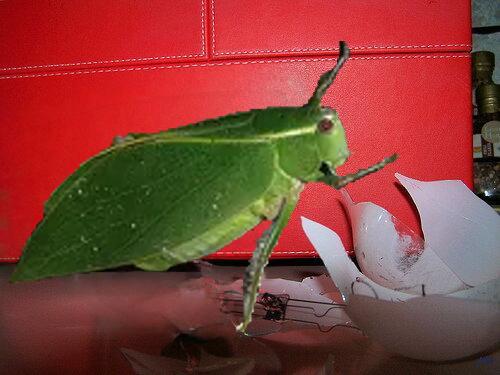} \includegraphics[max width=.1\columnwidth, max height=2cm, valign=c]{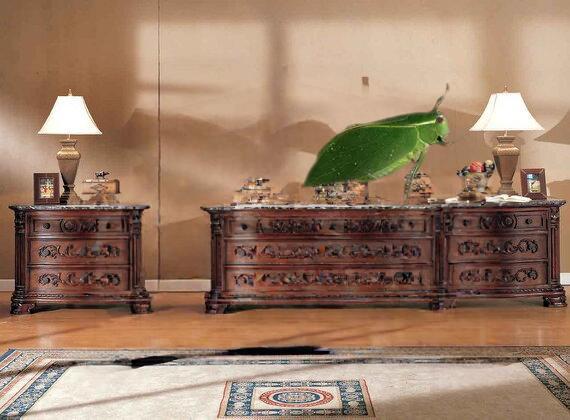} \includegraphics[max width=.1\columnwidth, max height=2cm, valign=c]{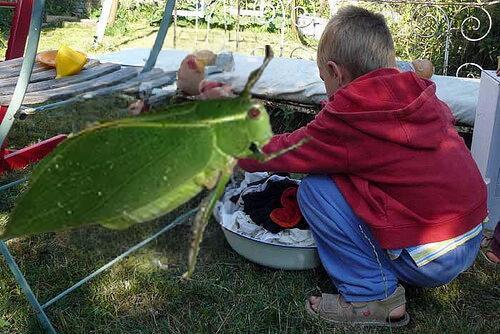} \includegraphics[max width=.1\columnwidth, max height=2cm, valign=c]{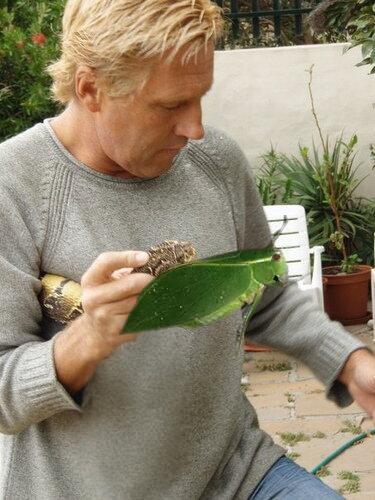} \includegraphics[max width=.1\columnwidth, max height=2cm, valign=c]{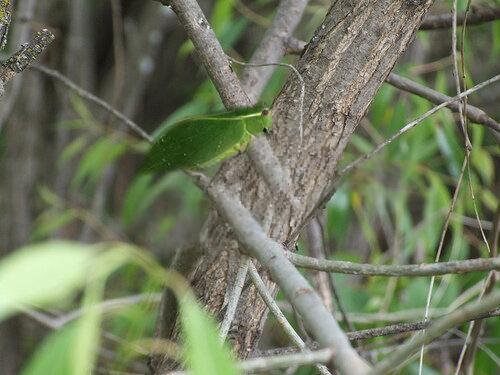} \includegraphics[max width=.1\columnwidth, max height=2cm, valign=c]{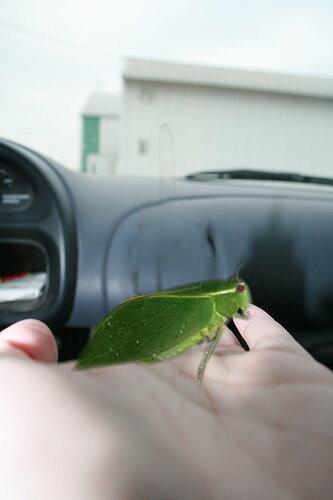} \\
			\makecell{n02443484        \\Black-Footed \\Ferret} & \includegraphics[max width=.1\columnwidth, max height=2cm, valign=c]{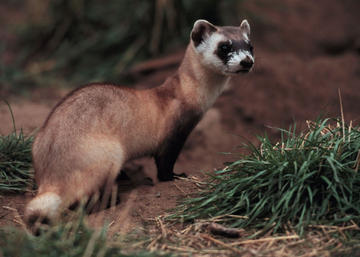} & \includegraphics[max width=.1\columnwidth, max height=2cm, valign=c]{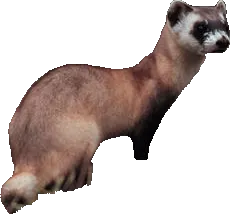} & \includegraphics[max width=.1\columnwidth, max height=2cm, valign=c]{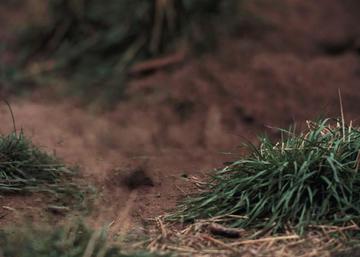} & \includegraphics[max width=.1\columnwidth, max height=2cm, valign=c]{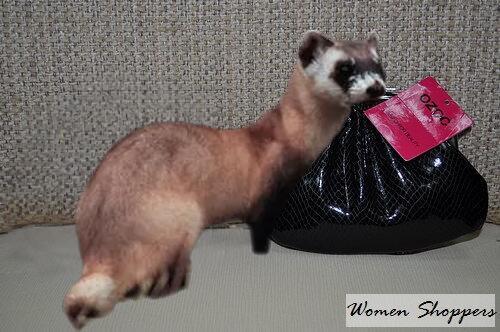} \includegraphics[max width=.1\columnwidth, max height=2cm, valign=c]{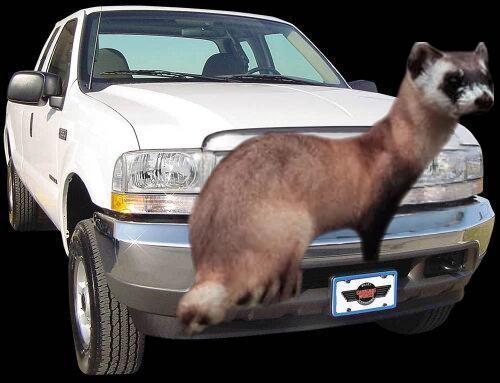} \includegraphics[max width=.1\columnwidth, max height=2cm, valign=c]{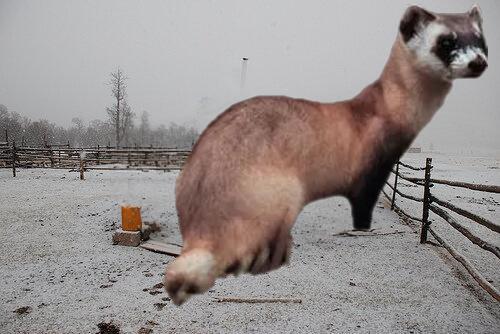} \includegraphics[max width=.1\columnwidth, max height=2cm, valign=c]{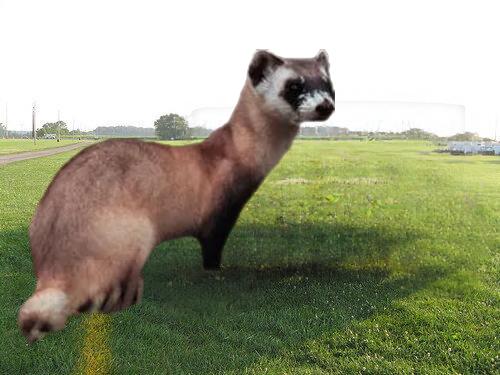} \includegraphics[max width=.1\columnwidth, max height=2cm, valign=c]{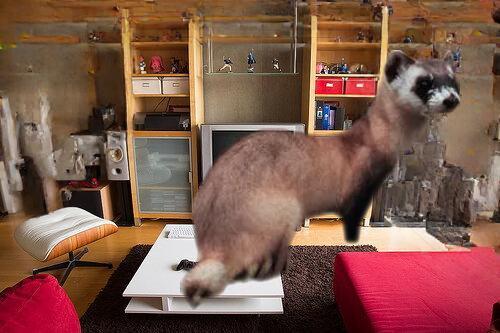} \includegraphics[max width=.1\columnwidth, max height=2cm, valign=c]{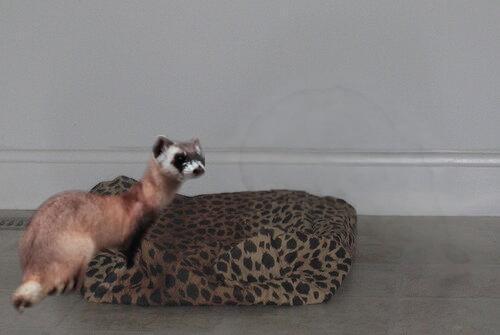} \\
			\makecell{n03201208        \\Dining Table} & \includegraphics[max width=.1\columnwidth, max height=2cm, valign=c]{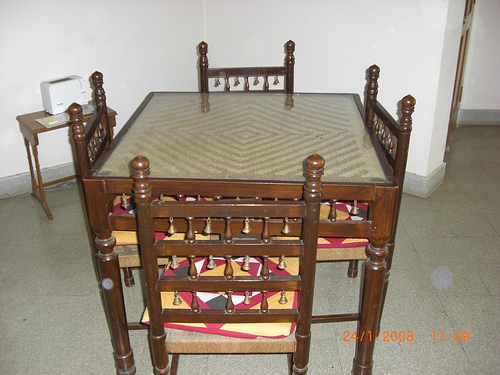} & \includegraphics[max width=.1\columnwidth, max height=2cm, valign=c]{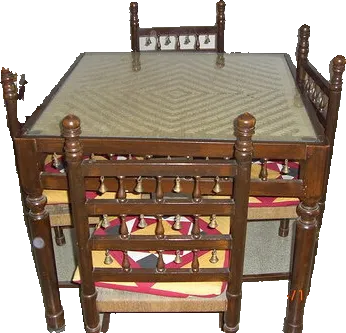} & \includegraphics[max width=.1\columnwidth, max height=2cm, valign=c]{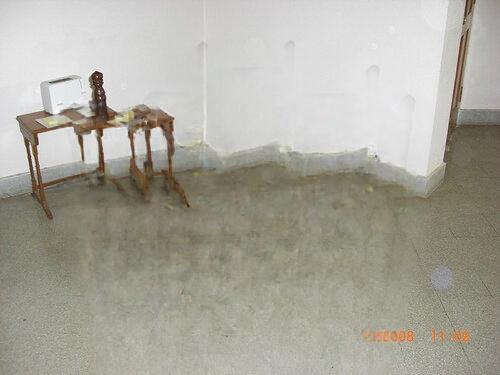} & \includegraphics[max width=.1\columnwidth, max height=2cm, valign=c]{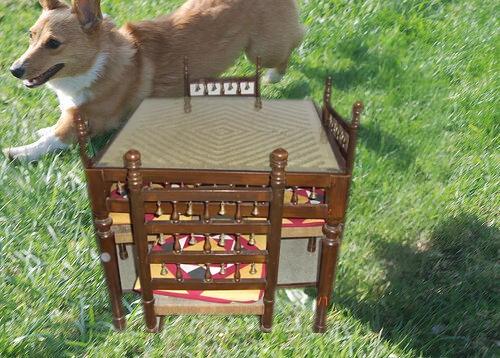} \includegraphics[max width=.1\columnwidth, max height=2cm, valign=c]{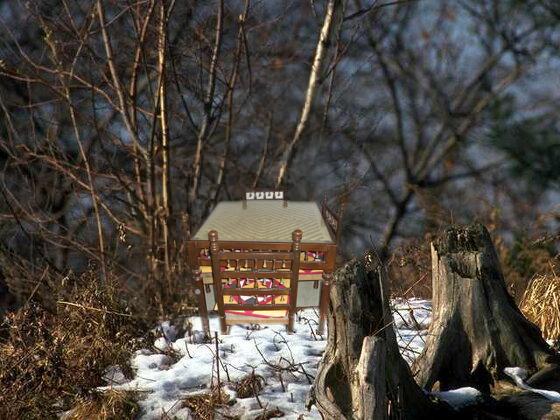} \includegraphics[max width=.1\columnwidth, max height=2cm, valign=c]{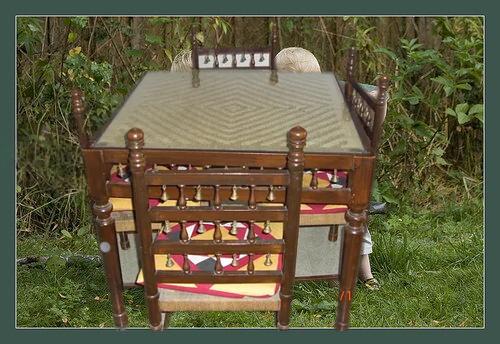} \includegraphics[max width=.1\columnwidth, max height=2cm, valign=c]{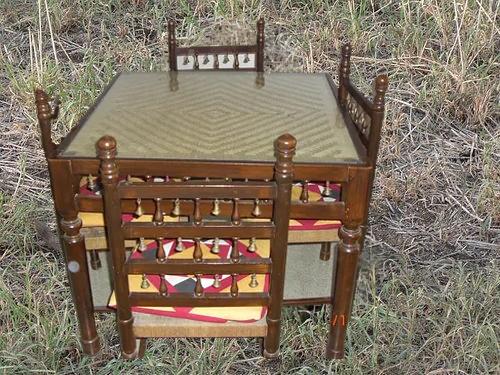} \includegraphics[max width=.1\columnwidth, max height=2cm, valign=c]{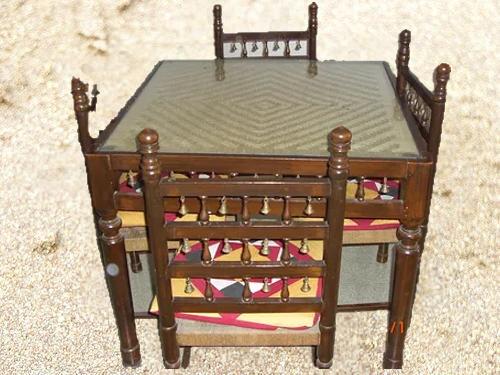} \includegraphics[max width=.1\columnwidth, max height=2cm, valign=c]{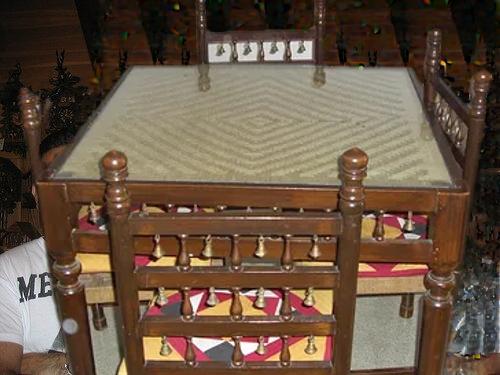} \\
			\makecell{n03424325        \\Gasmask} & \includegraphics[max width=.1\columnwidth, max height=2cm, valign=c]{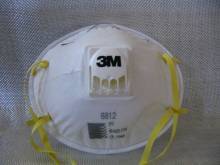} & \includegraphics[max width=.1\columnwidth, max height=2cm, valign=c]{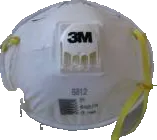} & \includegraphics[max width=.1\columnwidth, max height=2cm, valign=c]{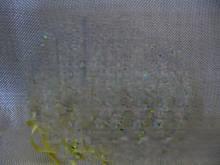} & \includegraphics[max width=.1\columnwidth, max height=2cm, valign=c]{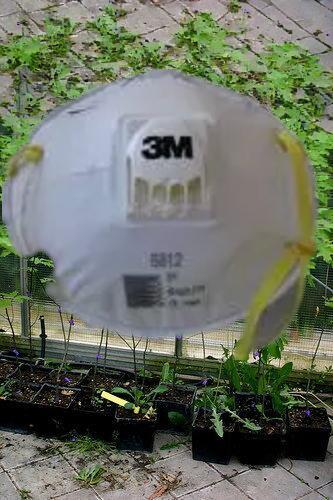} \includegraphics[max width=.1\columnwidth, max height=2cm, valign=c]{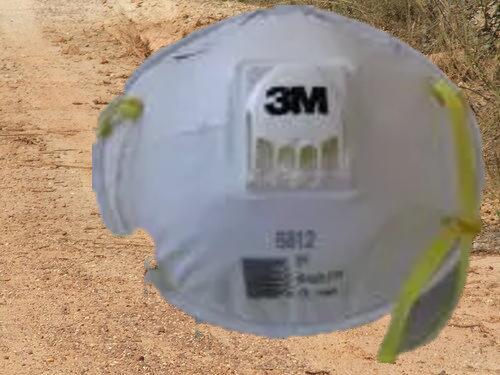} \includegraphics[max width=.1\columnwidth, max height=2cm, valign=c]{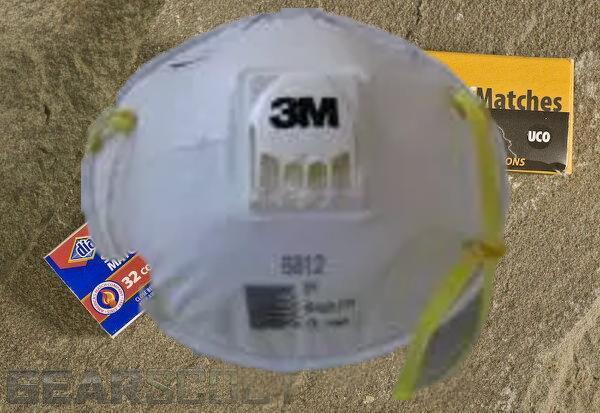} \includegraphics[max width=.1\columnwidth, max height=2cm, valign=c]{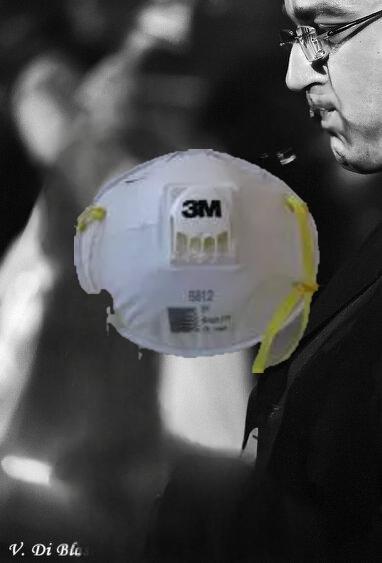} \includegraphics[max width=.1\columnwidth, max height=2cm, valign=c]{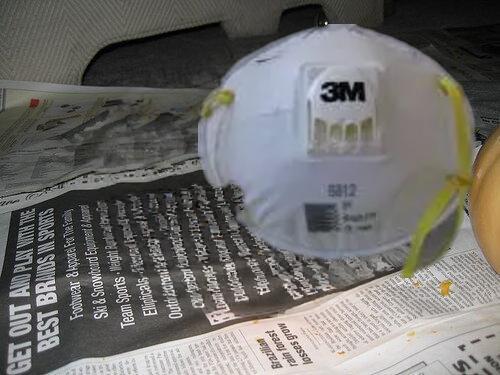} \includegraphics[max width=.1\columnwidth, max height=2cm, valign=c]{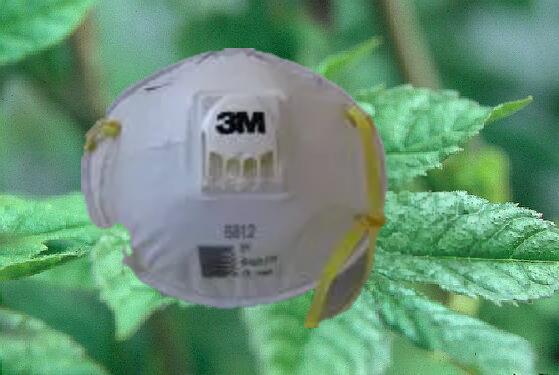} \\
			\makecell{n03642806        \\Laptop} & \includegraphics[max width=.1\columnwidth, max height=2cm, valign=c]{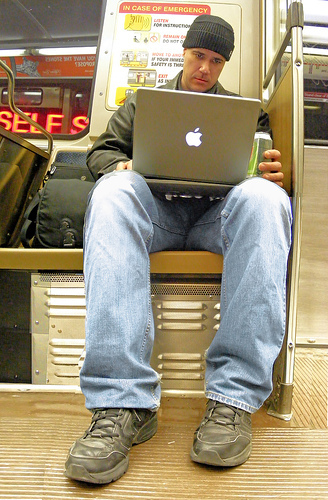} & \includegraphics[max width=.1\columnwidth, max height=2cm, valign=c]{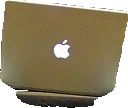} & \includegraphics[max width=.1\columnwidth, max height=2cm, valign=c]{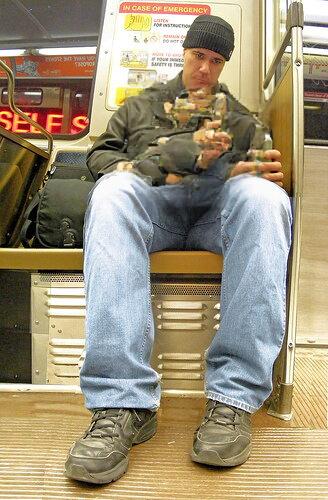} & \includegraphics[max width=.1\columnwidth, max height=2cm, valign=c]{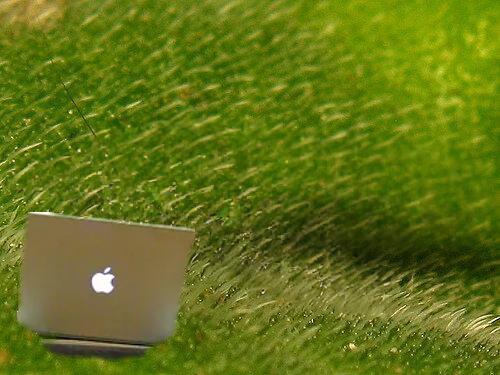} \includegraphics[max width=.1\columnwidth, max height=2cm, valign=c]{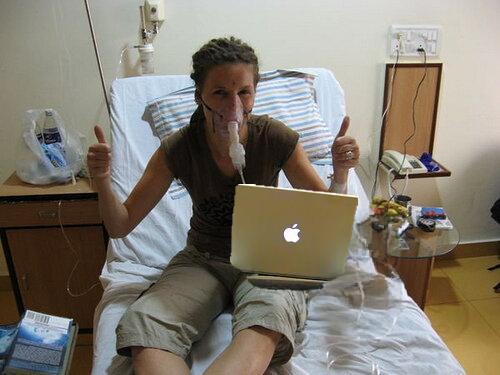} \includegraphics[max width=.1\columnwidth, max height=2cm, valign=c]{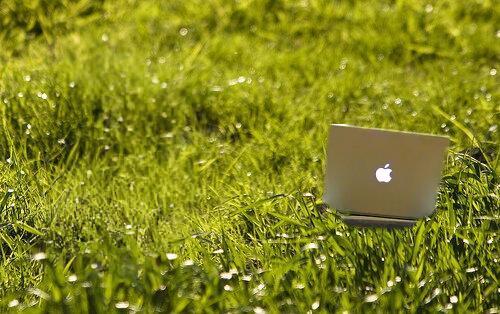} \includegraphics[max width=.1\columnwidth, max height=2cm, valign=c]{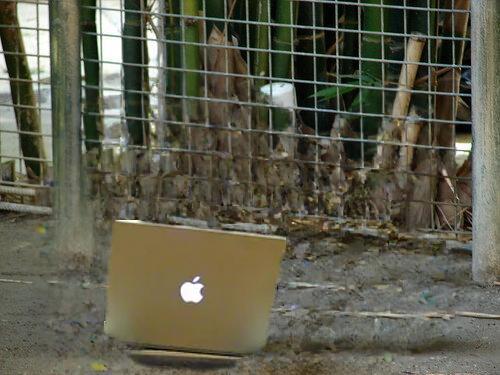} \includegraphics[max width=.1\columnwidth, max height=2cm, valign=c]{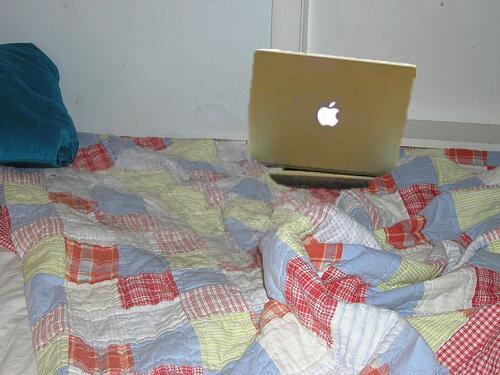} \includegraphics[max width=.1\columnwidth, max height=2cm, valign=c]{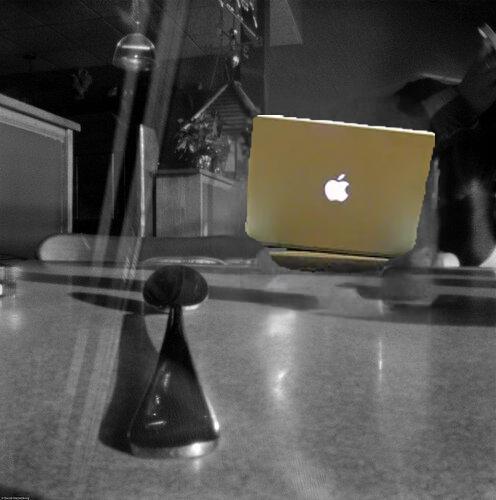} \\
			\makecell{n04141975        \\Scale} & \includegraphics[max width=.1\columnwidth, max height=2cm, valign=c]{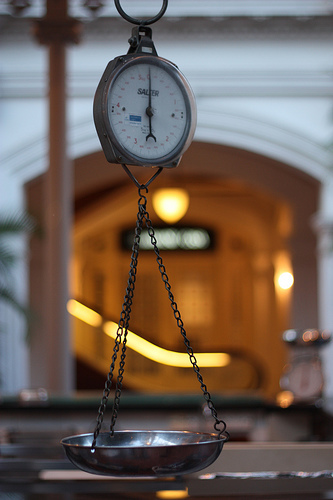} & \includegraphics[max width=.1\columnwidth, max height=2cm, valign=c]{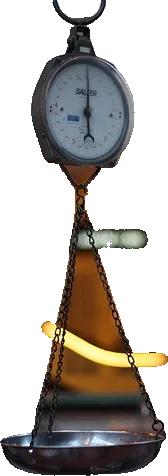} & \includegraphics[max width=.1\columnwidth, max height=2cm, valign=c]{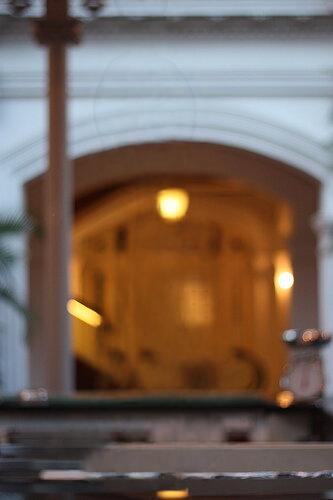} & \includegraphics[max width=.1\columnwidth, max height=2cm, valign=c]{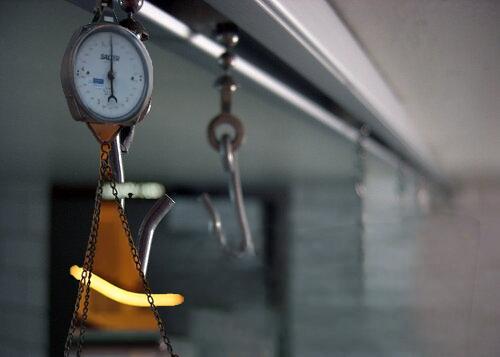}  \includegraphics[max width=.1\columnwidth, max height=2cm, valign=c]{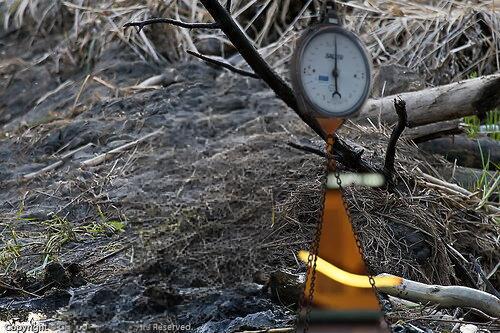} \includegraphics[max width=.1\columnwidth, max height=2cm, valign=c]{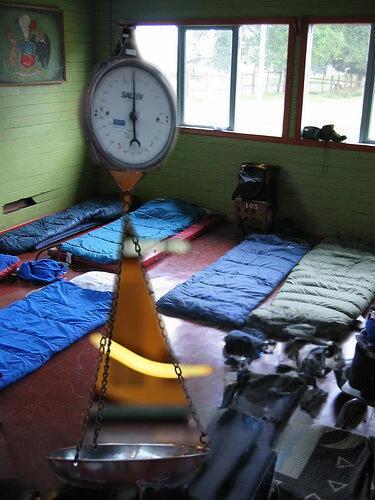} \includegraphics[max width=.1\columnwidth, max height=2cm, valign=c]{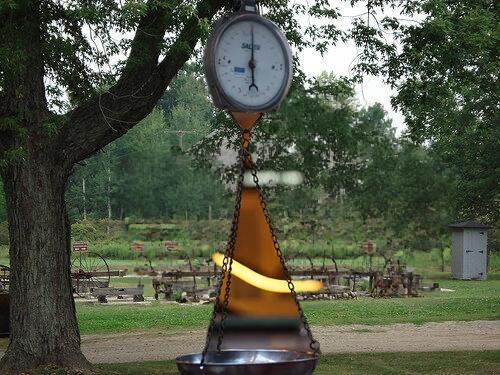} \includegraphics[max width=.1\columnwidth, max height=2cm, valign=c]{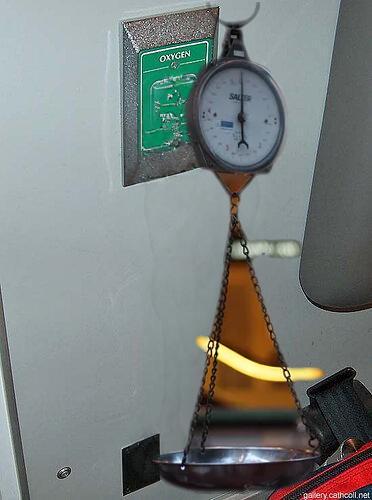}\includegraphics[max width=.1\columnwidth, max height=2cm, valign=c]{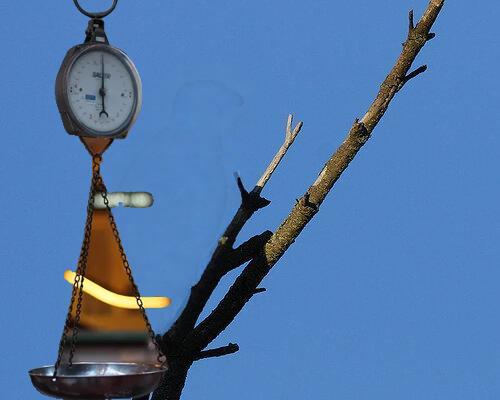} \\
			\makecell{n07714990        \\Broccoli} & \includegraphics[max width=.1\columnwidth, max height=2cm, valign=c]{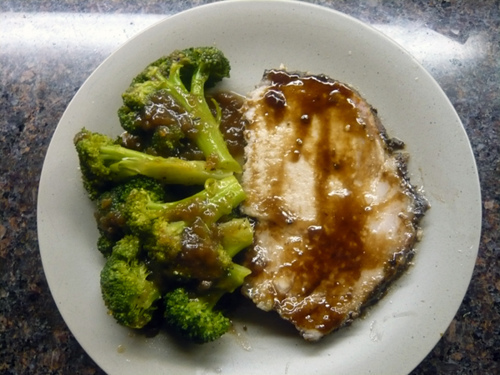} & \includegraphics[max width=.1\columnwidth, max height=2cm, valign=c]{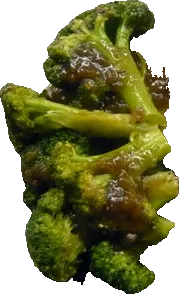} & \includegraphics[max width=.1\columnwidth, max height=2cm, valign=c]{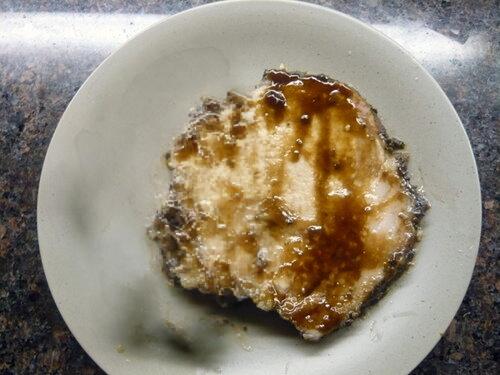} & \includegraphics[max width=.1\columnwidth, max height=2cm, valign=c]{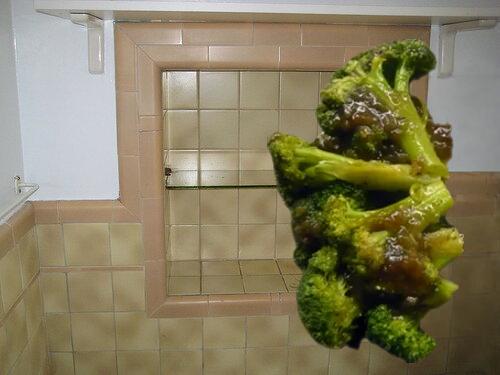} \includegraphics[max width=.1\columnwidth, max height=2cm, valign=c]{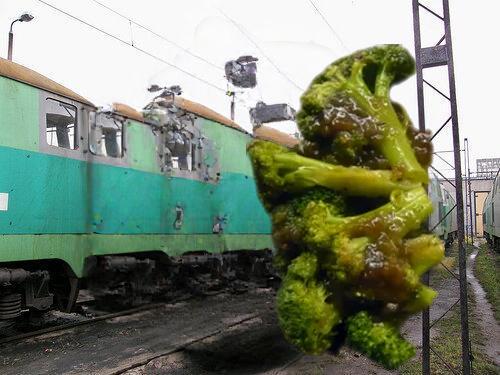} \includegraphics[max width=.1\columnwidth, max height=2cm, valign=c]{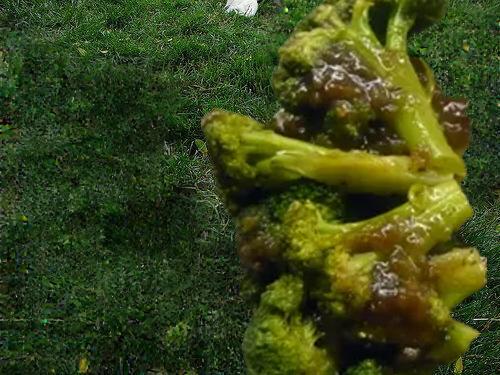} \includegraphics[max width=.1\columnwidth, max height=2cm, valign=c]{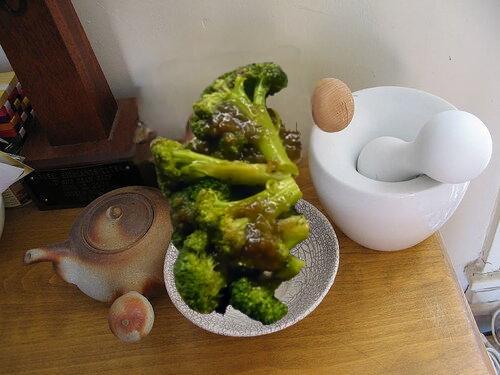} \includegraphics[max width=.1\columnwidth, max height=2cm, valign=c]{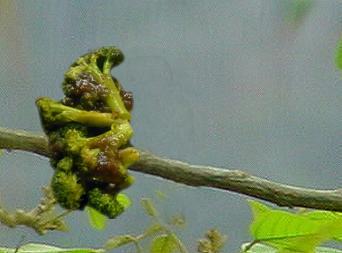} \includegraphics[max width=.1\columnwidth, max height=2cm, valign=c]{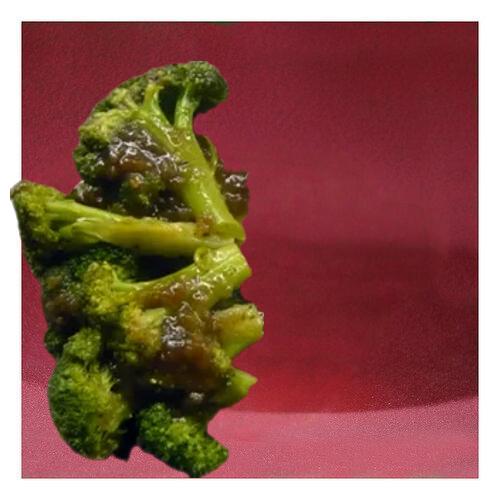} \\
			\makecell{n07749582        \\Lemon} & \includegraphics[max width=.1\columnwidth, max height=2cm, valign=c]{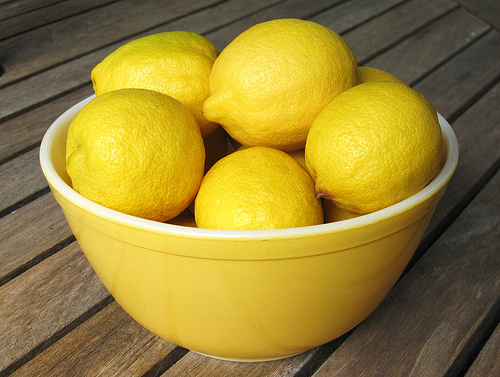} & \includegraphics[max width=.1\columnwidth, max height=2cm, valign=c]{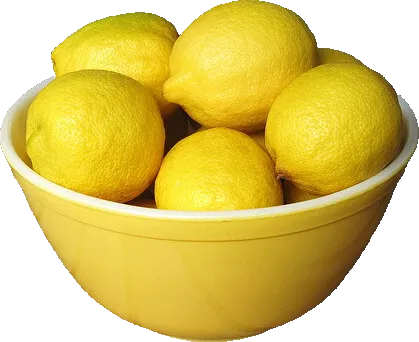} & \includegraphics[max width=.1\columnwidth, max height=2cm, valign=c]{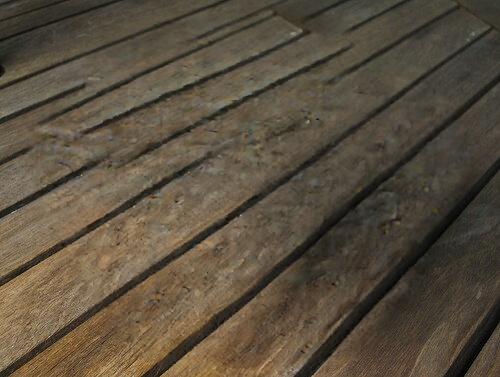} & \includegraphics[max width=.1\columnwidth, max height=2cm, valign=c]{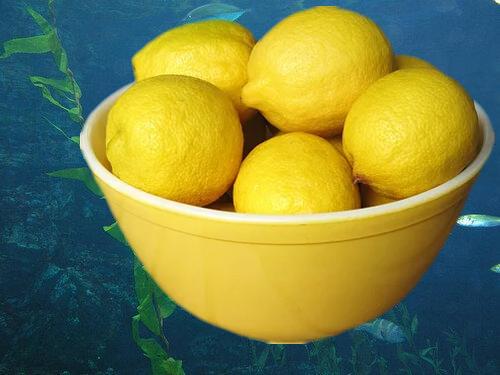} \includegraphics[max width=.1\columnwidth, max height=2cm, valign=c]{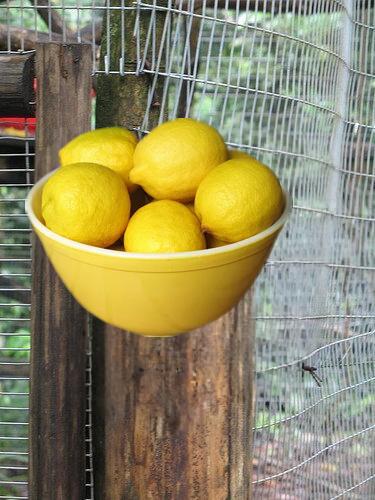} \includegraphics[max width=.1\columnwidth, max height=2cm, valign=c]{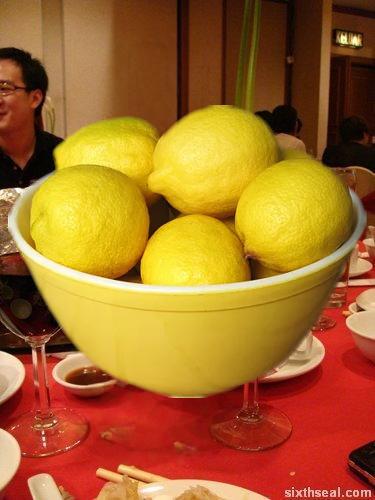} \includegraphics[max width=.1\columnwidth, max height=2cm, valign=c]{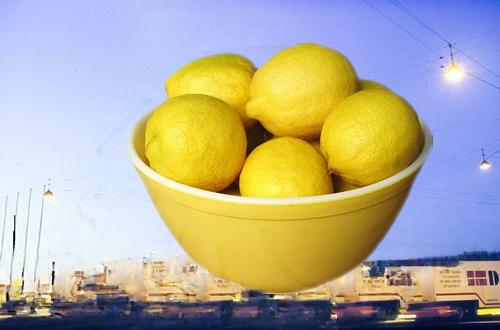} \includegraphics[max width=.1\columnwidth, max height=2cm, valign=c]{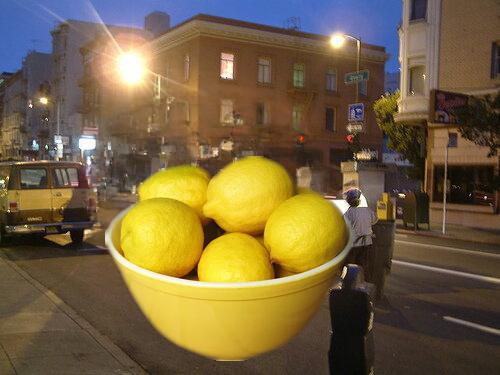} \includegraphics[max width=.1\columnwidth, max height=2cm, valign=c]{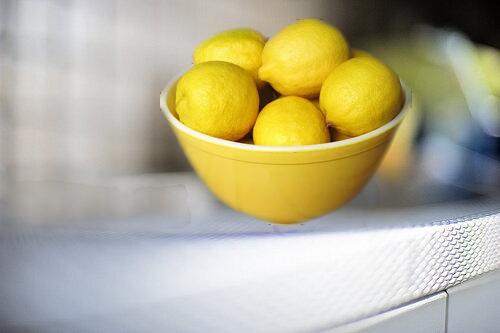} \\
			\makecell{n09332890        \\Lakeside} & \includegraphics[max width=.1\columnwidth, max height=2cm, valign=c]{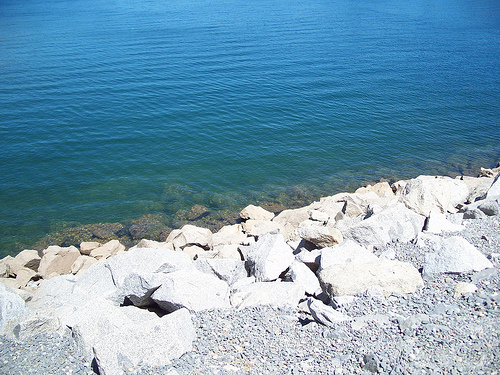} & \includegraphics[max width=.1\columnwidth, max height=2cm, valign=c]{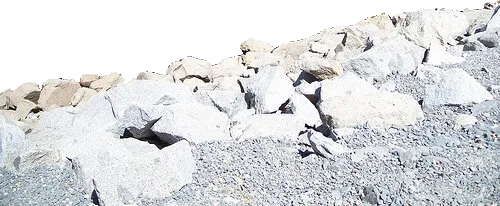} & \includegraphics[max width=.1\columnwidth, max height=2cm, valign=c]{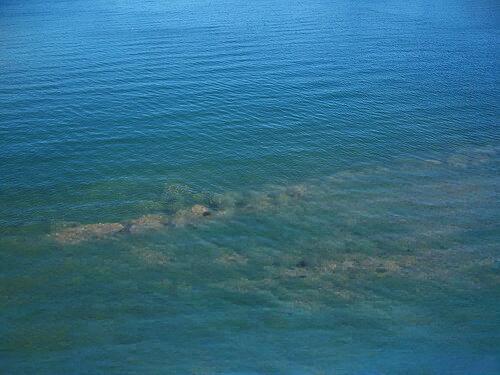} & \includegraphics[max width=.1\columnwidth, max height=2cm, valign=c]{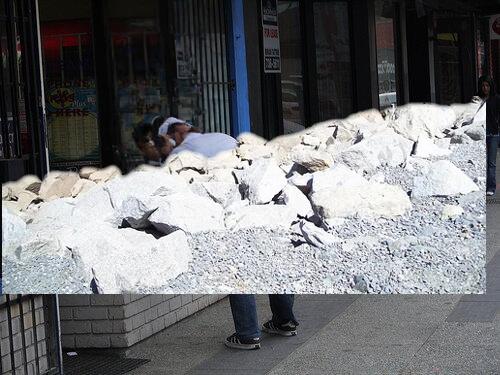} \includegraphics[max width=.1\columnwidth, max height=2cm, valign=c]{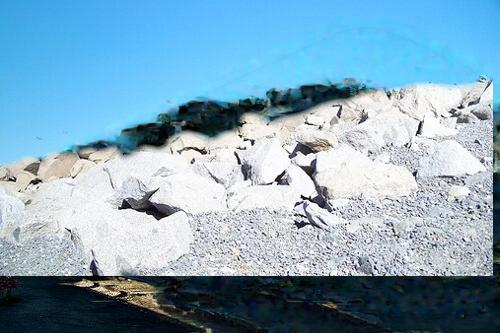} \includegraphics[max width=.1\columnwidth, max height=2cm, valign=c]{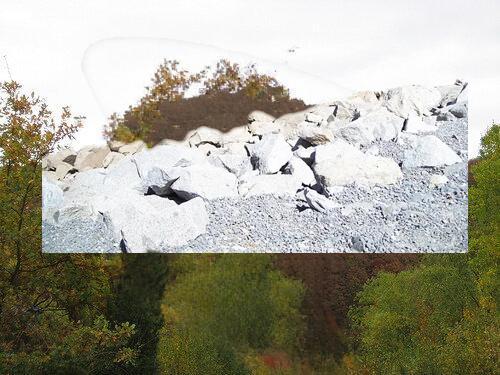} \includegraphics[max width=.1\columnwidth, max height=2cm, valign=c]{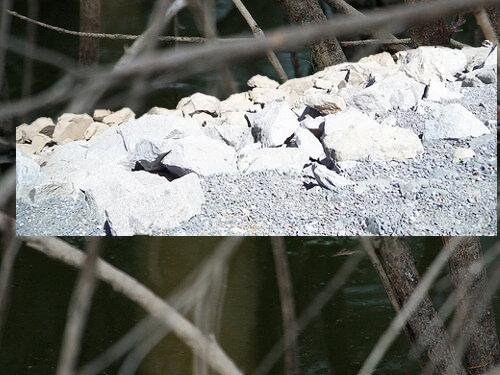} \includegraphics[max width=.1\columnwidth, max height=2cm, valign=c]{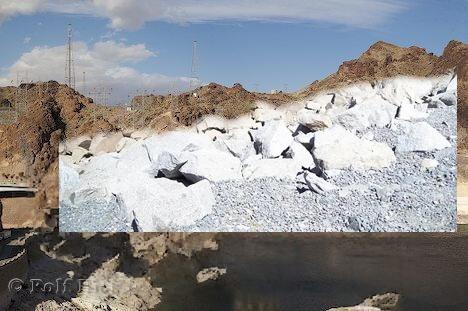} \includegraphics[max width=.1\columnwidth, max height=2cm, valign=c]{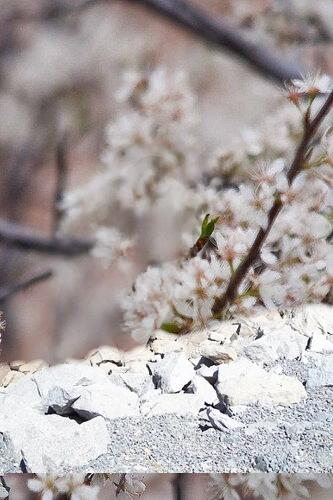} \\
			\bottomrule
		\end{tabular}
	}
\end{table*}
We show some example images of \schemename's recombinations for 14 random classes of ImageNet \cite{Deng2009} in \Cref{tbl:example-images}.
The recombined samples display substantial visual diversity, with each extracted foreground appearing in multiple, clearly different background contexts.
Foreground objects remain sharp and well‑preserved across recombinations, while backgrounds vary in texture, color, and scene type
Images show a broad range of spatial placements and scales for the same object, resulting in noticeably different overall layouts.

\FloatBarrier
\section{Multi-Instance Images}
\label{apdx:multiinstance}
Instead of a single object, ForAug treats \emph{all} instances of the label-class as the FG, and we produce a mask covering all instances matching the label-prompt.
This is enabled by open-vocabulary prompting in SAM~\cite{Kirillov2023}, which produces a mask covering all instances matching the prompt.
The filtering (Eq. 1) further penalizes BGs containing label information, including cases where not all instances were segmented.
Figure 1 (main paper) and \Cref{tbl:multiinstance-examples} show concrete examples.
To quantify multi-instance behavior, we stratify our segmentation evaluation by GT instance count $n$.
Box recall (fraction of bounding boxes covered by mask) declines gracefully: 98.2\% at n=1 to 74.2\% at n=3, and 66.8\% at n=4 (95th percentile of images; $2.7$ average instances) and box precision (fraction of mask inside bounding boxes) remains above 84\% across this range.
For Box recall, we use a 5\% overlap threshold to accommodate thin classes (e.g., snakes, fishing rods) whose objects occupy only a small fraction of their bounding boxes.
All in all, Multi-instance images pose no issue, as even a segmented subset preserves class semantics and the remaining instances are likely non-salient, avaiding label-ambiguity via the BG.

\begin{table*}[t!]
	\centering
	\caption{\textbf{Multi-instance samples.} \schemename segments \emph{all} instances of the label class jointly and recombines them together across different backgrounds.}
	\label{tbl:multiinstance-examples}
	\begin{tabular}{cccl}
		\toprule
		Class                                                                                                                 & \makecell{Original \\Image} & \makecell{Extracted\\Foreground} & \schemename's Recombinations \\
		\midrule
		\makecell{n01530575                                                                                                                        \\Brambling} &
		\includegraphics[max width=.1\columnwidth, max height=2cm, valign=c]{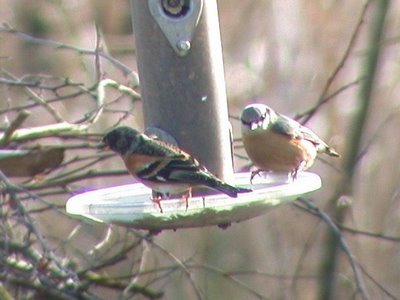}  &
		\includegraphics[max width=.1\columnwidth, max height=2cm, valign=c]{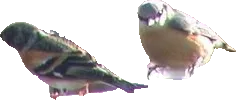}   &
		\includegraphics[max width=.1\columnwidth, max height=2cm, valign=c]{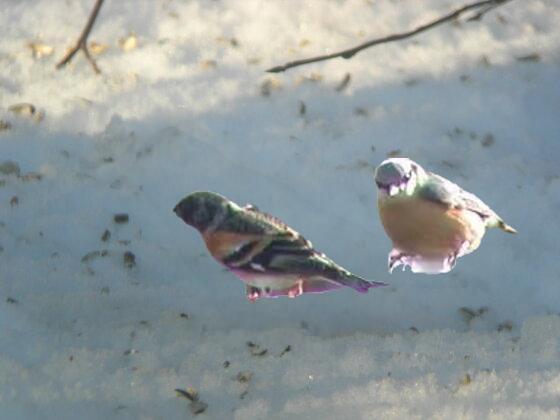}
		\includegraphics[max width=.1\columnwidth, max height=2cm, valign=c]{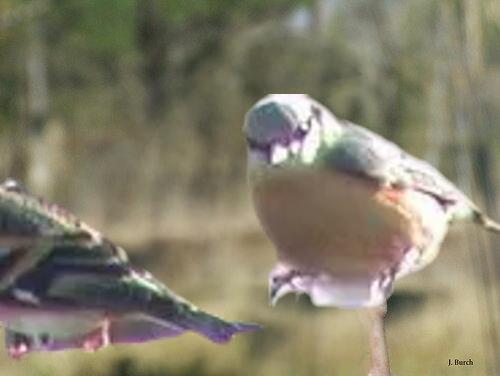}
		\includegraphics[max width=.1\columnwidth, max height=2cm, valign=c]{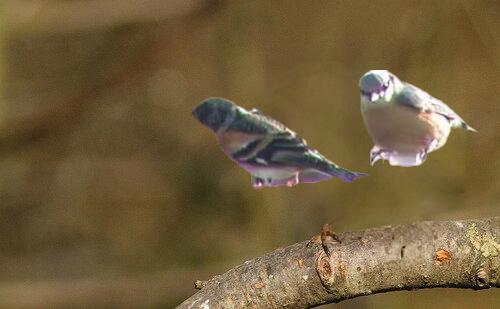}
		\includegraphics[max width=.1\columnwidth, max height=2cm, valign=c]{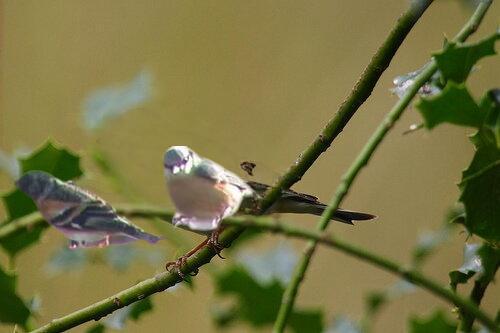}        \\ \\
		\makecell{n01882714                                                                                                                        \\Koala} &
		\includegraphics[max width=.1\columnwidth, max height=2cm, valign=c]{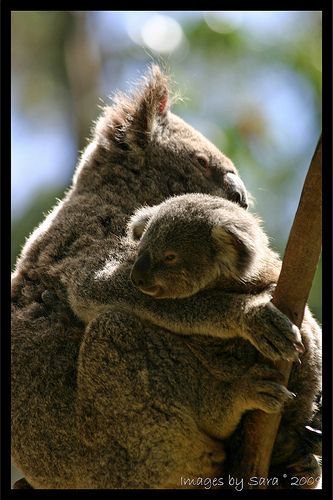} &
		\includegraphics[max width=.1\columnwidth, max height=2cm, valign=c]{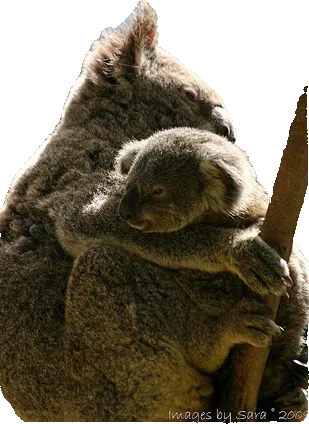}  &
		\includegraphics[max width=.1\columnwidth, max height=2cm, valign=c]{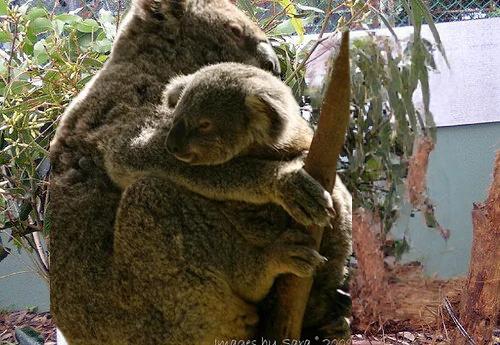}
		\includegraphics[max width=.1\columnwidth, max height=2cm, valign=c]{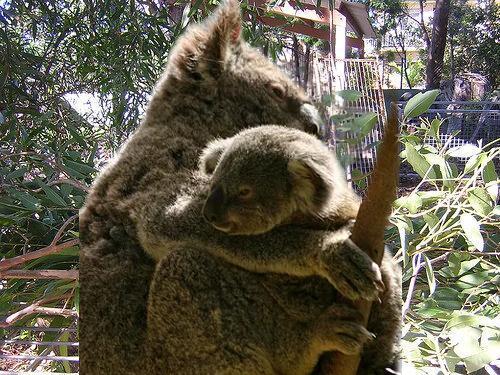}
		\includegraphics[max width=.1\columnwidth, max height=2cm, valign=c]{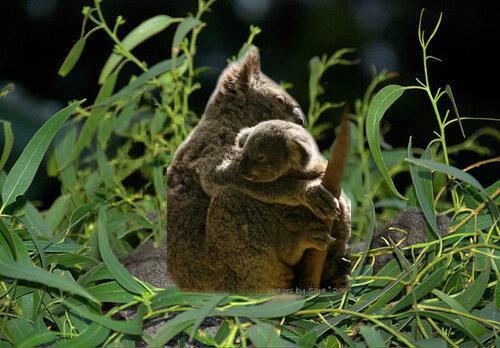}
		\includegraphics[max width=.1\columnwidth, max height=2cm, valign=c]{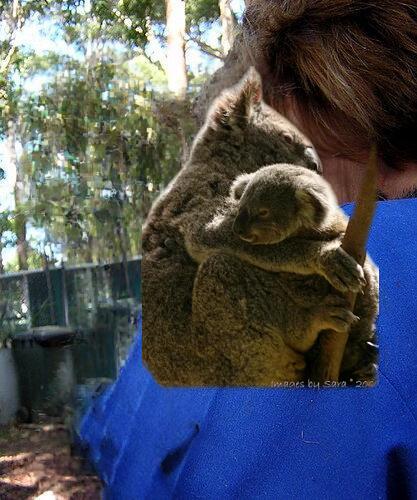}       \\ \\
		\makecell{n02101388                                                                                                                        \\Brittany Spaniel} &
		\includegraphics[max width=.1\columnwidth, max height=2cm, valign=c]{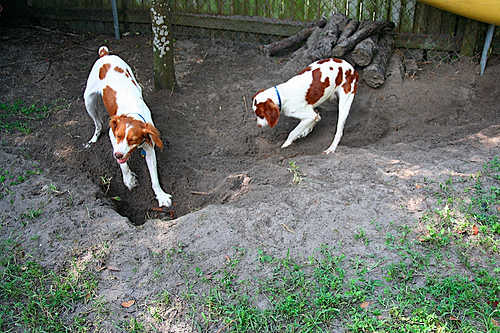}  &
		\includegraphics[max width=.1\columnwidth, max height=2cm, valign=c]{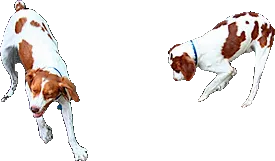}   &
		\includegraphics[max width=.1\columnwidth, max height=2cm, valign=c]{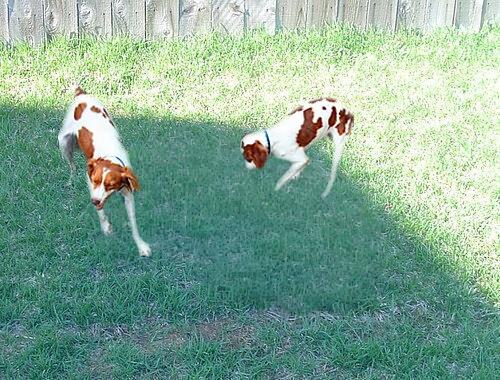}
		\includegraphics[max width=.1\columnwidth, max height=2cm, valign=c]{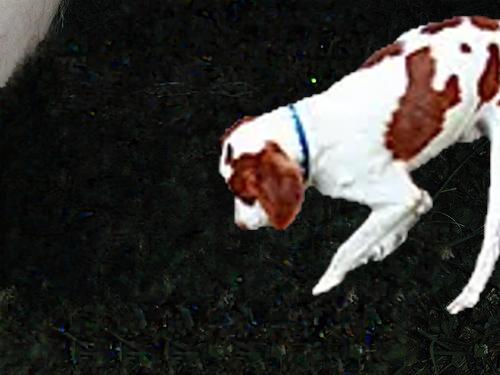}
		\includegraphics[max width=.1\columnwidth, max height=2cm, valign=c]{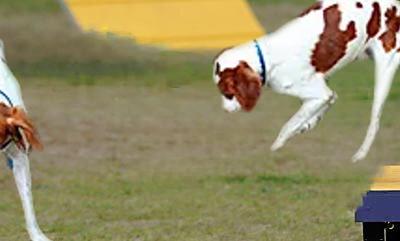}
		\includegraphics[max width=.1\columnwidth, max height=2cm, valign=c]{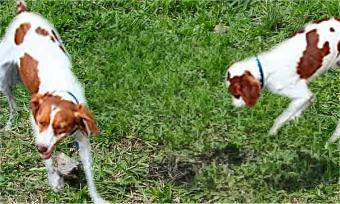}        \\ \\
		\makecell{n04204238                                                                                                                        \\Shopping Basket} &
		\includegraphics[max width=.1\columnwidth, max height=2cm, valign=c]{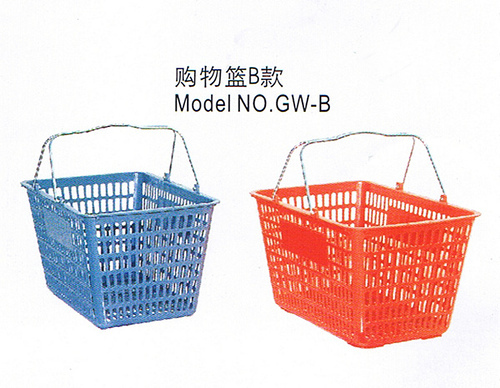}  &
		\includegraphics[max width=.1\columnwidth, max height=2cm, valign=c]{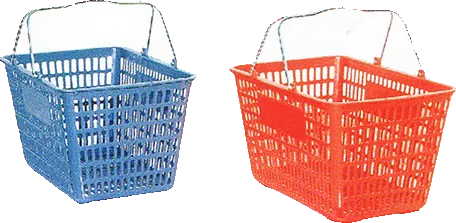}   &
		\includegraphics[max width=.1\columnwidth, max height=2cm, valign=c]{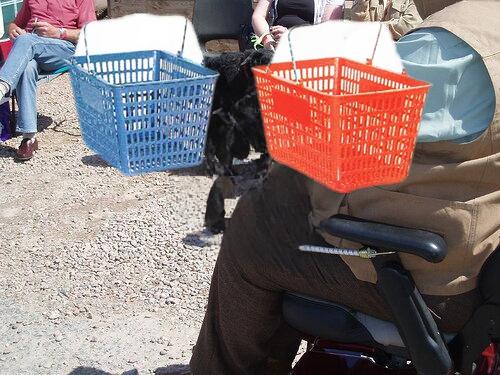}
		\includegraphics[max width=.1\columnwidth, max height=2cm, valign=c]{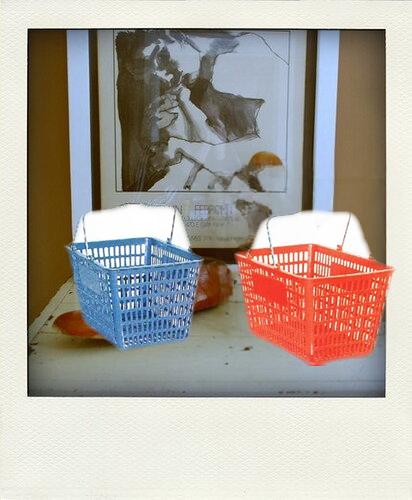}
		\includegraphics[max width=.1\columnwidth, max height=2cm, valign=c]{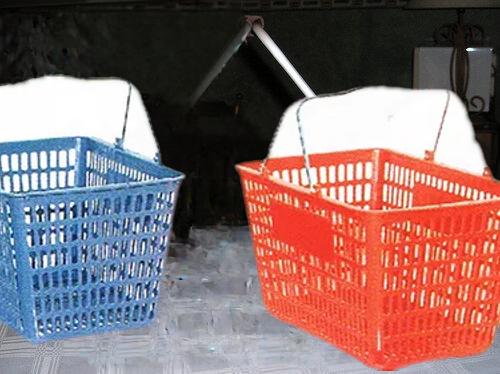}
		\includegraphics[max width=.1\columnwidth, max height=2cm, valign=c]{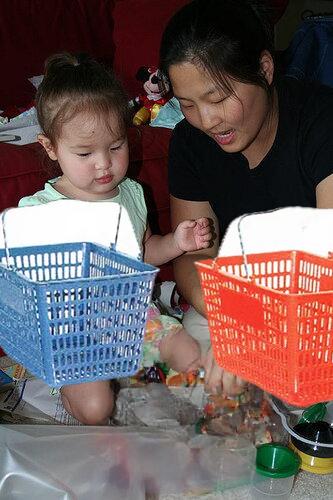}        \\ \\
		\makecell{n04592741                                                                                                                        \\Wing} &
		\includegraphics[max width=.1\columnwidth, max height=2cm, valign=c]{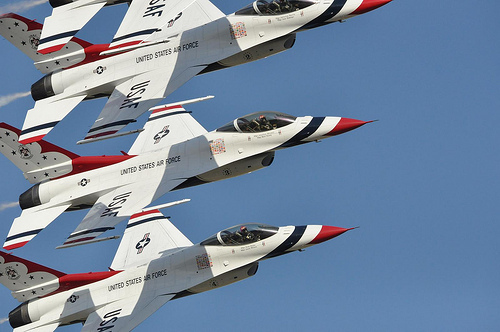} &
		\includegraphics[max width=.1\columnwidth, max height=2cm, valign=c]{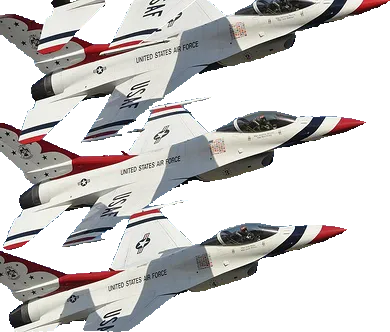}  &
		\includegraphics[max width=.1\columnwidth, max height=2cm, valign=c]{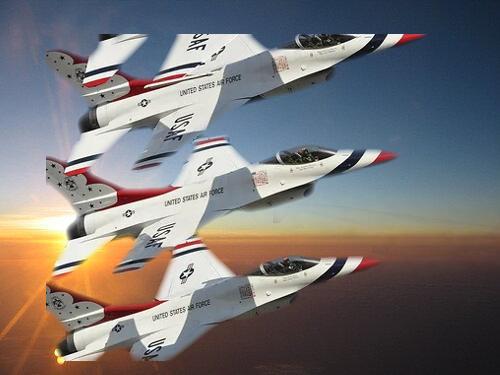}
		\includegraphics[max width=.1\columnwidth, max height=2cm, valign=c]{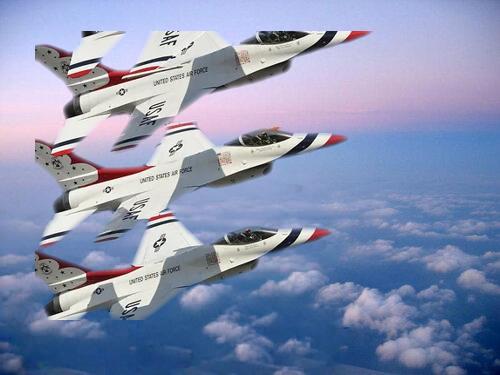}
		\includegraphics[max width=.1\columnwidth, max height=2cm, valign=c]{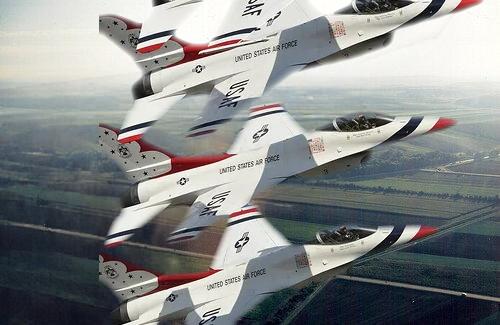}
		\includegraphics[max width=.1\columnwidth, max height=2cm, valign=c]{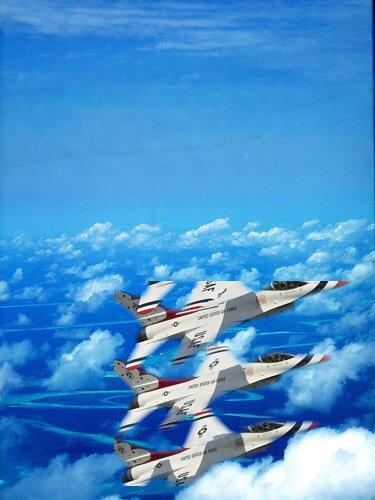}       \\ \\
		\makecell{n07753592                                                                                                                        \\Banana} &
		\includegraphics[max width=.1\columnwidth, max height=2cm, valign=c]{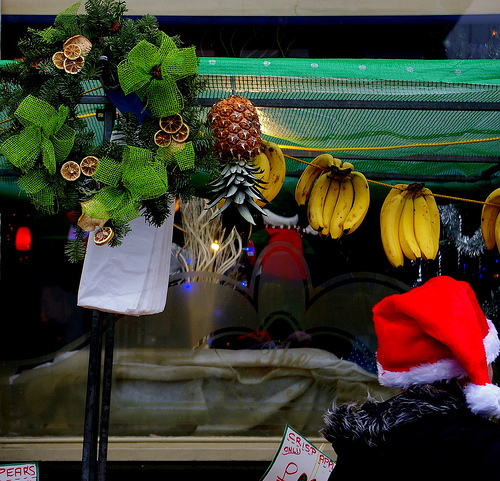}  &
		\includegraphics[max width=.1\columnwidth, max height=2cm, valign=c]{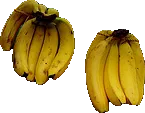}   &
		\includegraphics[max width=.1\columnwidth, max height=2cm, valign=c]{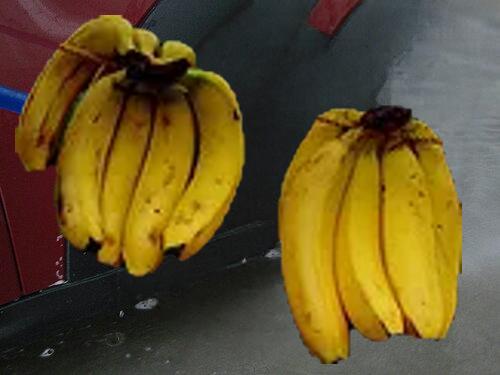}
		\includegraphics[max width=.1\columnwidth, max height=2cm, valign=c]{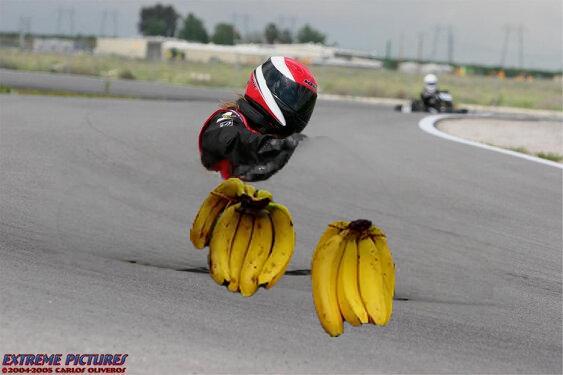}
		\includegraphics[max width=.1\columnwidth, max height=2cm, valign=c]{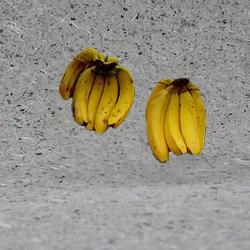}
		\includegraphics[max width=.1\columnwidth, max height=2cm, valign=c]{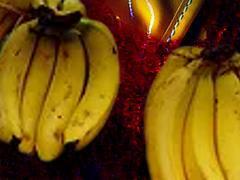}        \\
		\bottomrule
	\end{tabular}
\end{table*}

\FloatBarrier
\section{Infill Model Comparison}
\label{apdx:infill-models}
\begin{table*}[h!]
	\centering
	\caption{\textbf{Example infills.} Of LaMa and Attentive Eraser.}
	\label{tab:infill-examples}
	\resizebox{.9\textwidth}{!}{
		\begin{tabular}{cc@{\hskip 0.3in}cc}
			\toprule
			LaMa                                                                                                  & Att. Eraser                                                                                              & LaMa                                                                                                  & Att. Eraser                                                                                              \\
			\midrule
			\includegraphics[width=.23\columnwidth, valign=c]{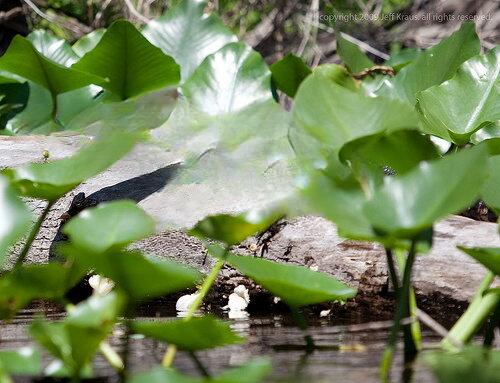} & \includegraphics[width=.23\columnwidth, valign=c]{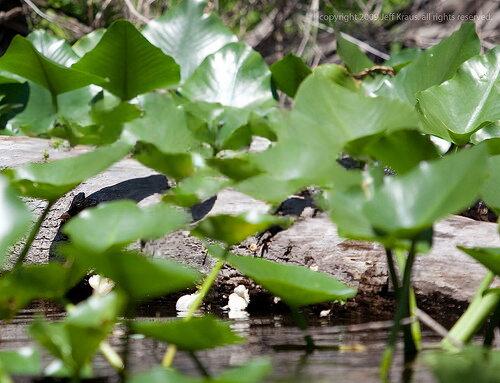} &
			\includegraphics[width=.23\columnwidth, valign=c]{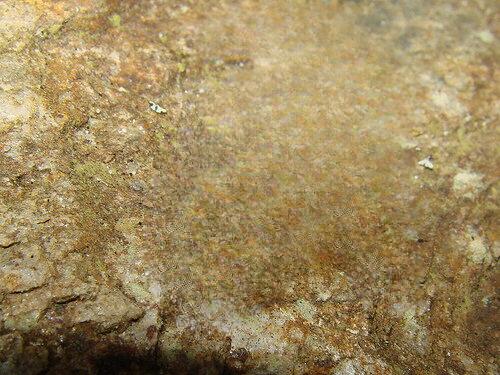} & \includegraphics[width=.23\columnwidth, valign=c]{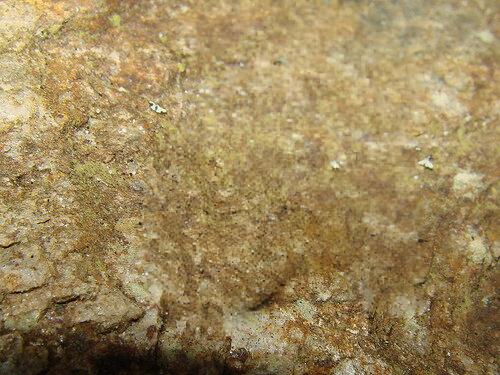}                                                                                                                                                                                                                    \\
			\includegraphics[width=.23\columnwidth, valign=c]{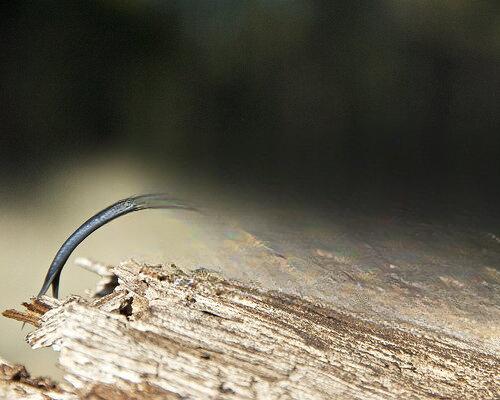} & \includegraphics[width=.23\columnwidth, valign=c]{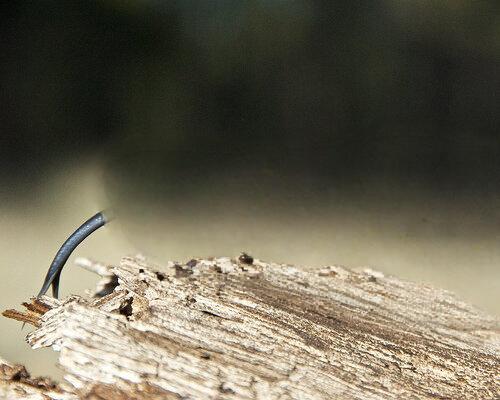} &
			\includegraphics[width=.23\columnwidth, valign=c]{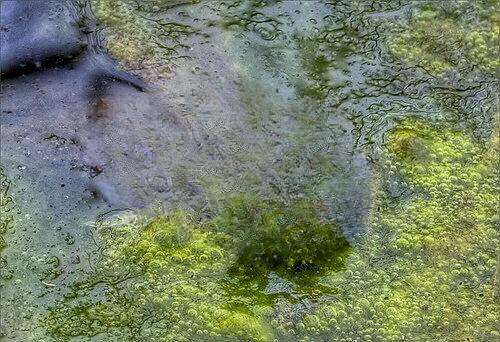} & \includegraphics[width=.23\columnwidth, valign=c]{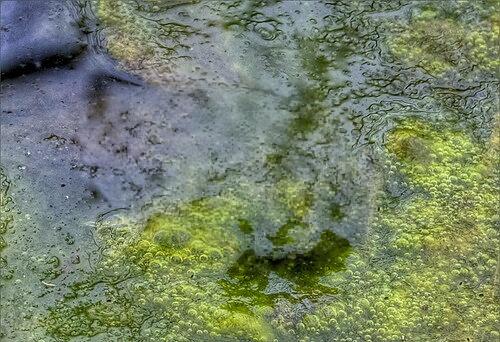}                                                                                                                                                                                                                    \\
			\includegraphics[width=.23\columnwidth, valign=c]{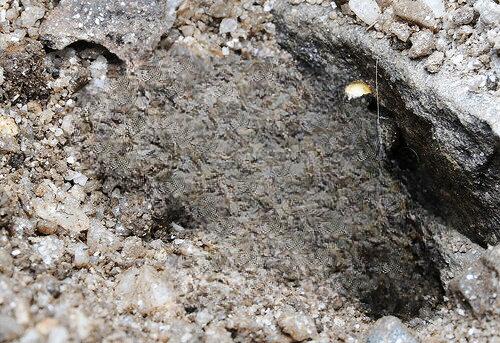} & \includegraphics[width=.23\columnwidth, valign=c]{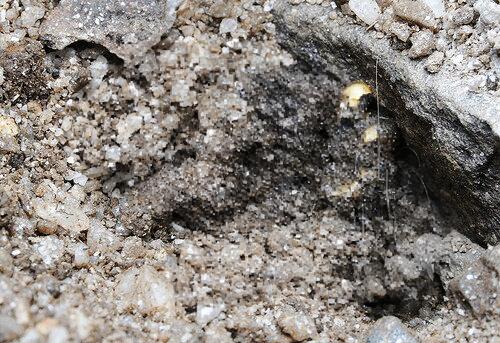} & \includegraphics[width=.23\columnwidth, valign=c]{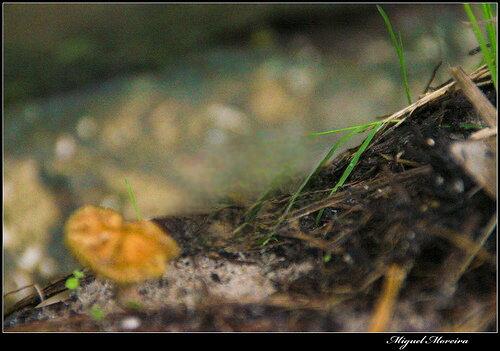} & \includegraphics[width=.23\columnwidth, valign=c]{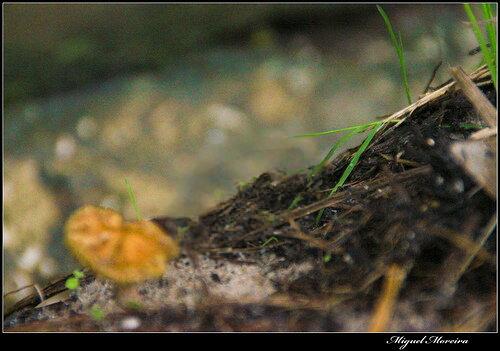} \\
			\includegraphics[width=.23\columnwidth, valign=c]{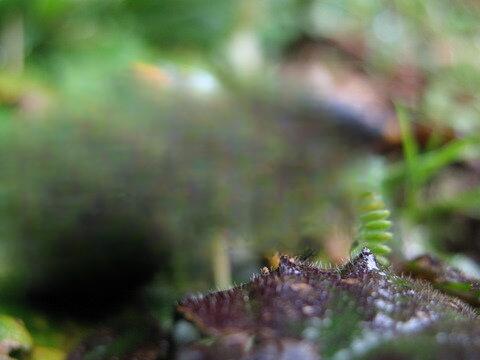} & \includegraphics[width=.23\columnwidth, valign=c]{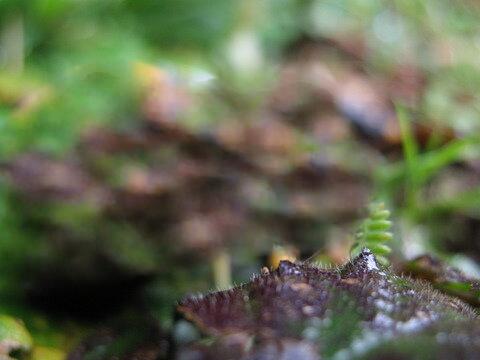} & \includegraphics[width=.23\columnwidth, valign=c]{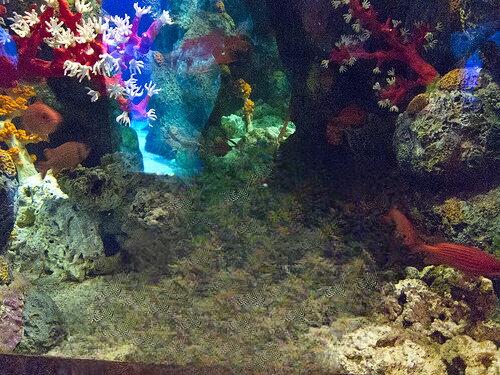} & \includegraphics[width=.23\columnwidth, valign=c]{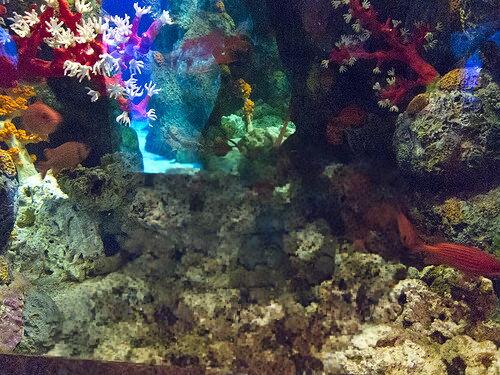} \\
			\includegraphics[width=.23\columnwidth, valign=c]{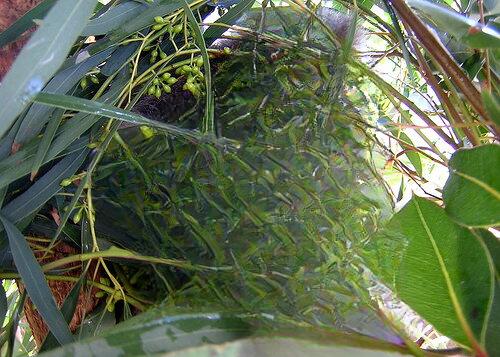} & \includegraphics[width=.23\columnwidth, valign=c]{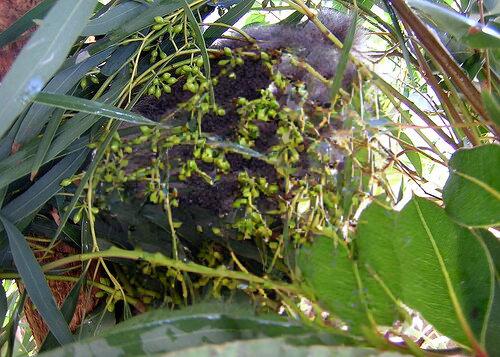} & \includegraphics[width=.23\columnwidth, valign=c]{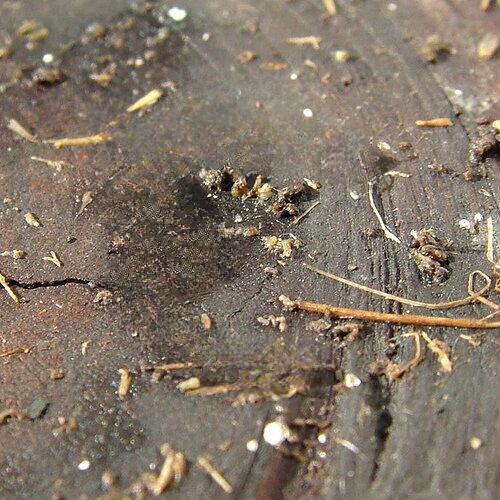} & \includegraphics[width=.23\columnwidth, valign=c]{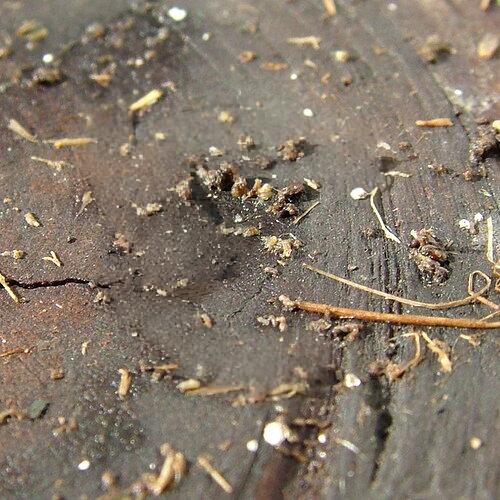} \\
			\includegraphics[width=.23\columnwidth, valign=c]{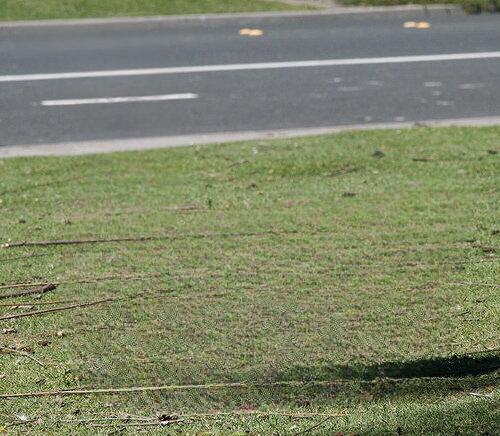} & \includegraphics[width=.23\columnwidth, valign=c]{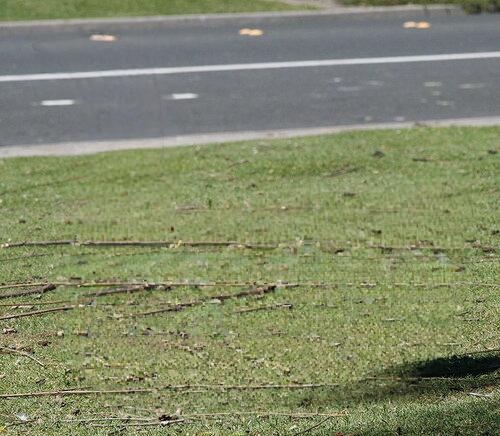} & \includegraphics[width=.23\columnwidth, valign=c]{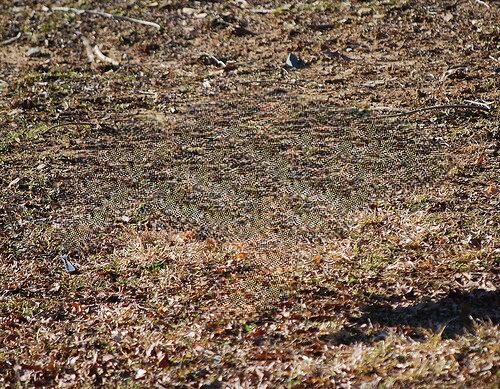} & \includegraphics[width=.23\columnwidth, valign=c]{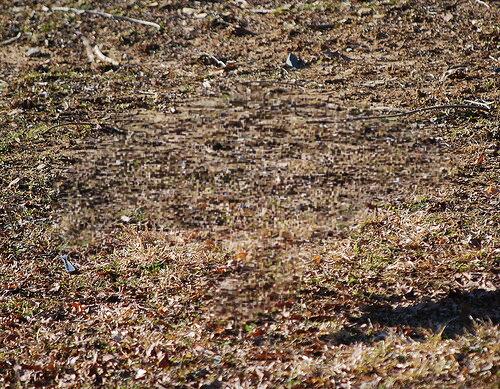} \\
			\bottomrule
		\end{tabular}
	}
\end{table*}
We visualize example infilled images for both LaMa \cite{Suvorov2022} and Attentive Eraser \cite{Sun2025} in \Cref{tab:infill-examples}.
The side‑by‑side examples show that both methods generally produce visually consistent infills, with many pairs appearing extremely similar at a glance.
We qualitatively find that Attentive Eraser yields slightly sharper textures or more coherent local structure, while LaMa sometimes produces smoother or more homogenized regions.
Across the table, fine‑detail areas such as foliage, bark, and ground textures reveal the most noticeable differences between the two methods.

\FloatBarrier
\newpage
\section{Image Infill Ratio}
\label{apdx:infill-ratio}
\begin{table*}[h!]
	\centering
	\caption{\textbf{Large infill-ratio examples.} Example infills with a large relative foreground area size that is infilled (infill ratio).}
	\label{tbl:high-rat}
	\resizebox{.8\textwidth}{!}{
		\begin{tabular}{ccc}
			\toprule
			Infill Ratio & LaMa                                                                                                             & Att. Eraser                                                                                                         \\
			\midrule
			83.7         & \raisebox{-50pt}{\includegraphics[width=.3\columnwidth]{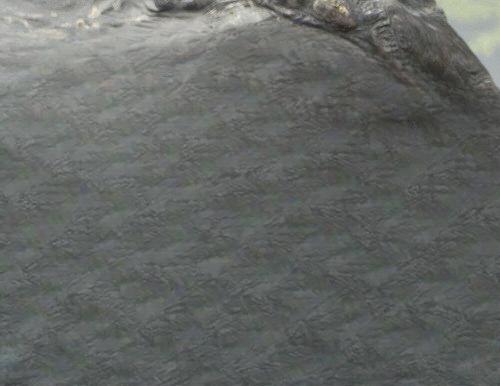}} & \raisebox{-50pt}{\includegraphics[width=.3\columnwidth]{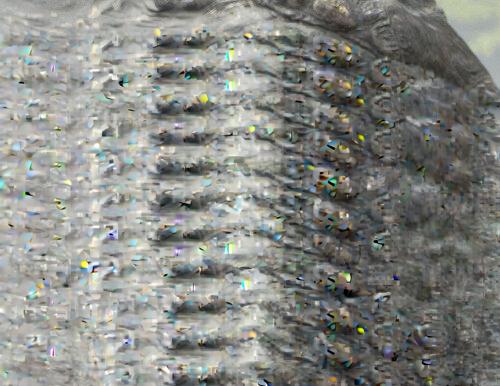}} \\ \\
			88.2         & \raisebox{-50pt}{\includegraphics[width=.3\columnwidth]{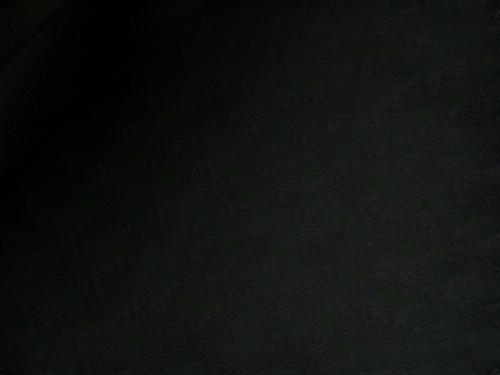}} & \raisebox{-50pt}{\includegraphics[width=.3\columnwidth]{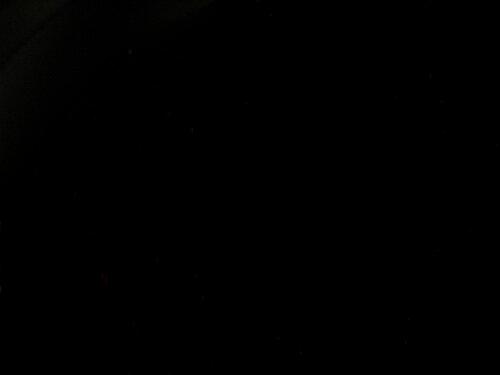}} \\ \\
			93.7         & \raisebox{-60pt}{\includegraphics[width=.3\columnwidth]{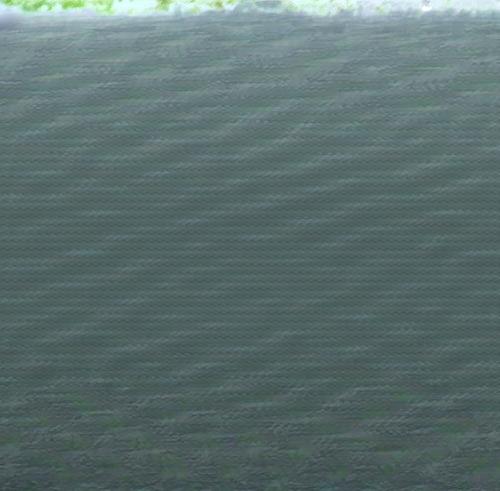}} & \raisebox{-60pt}{\includegraphics[width=.3\columnwidth]{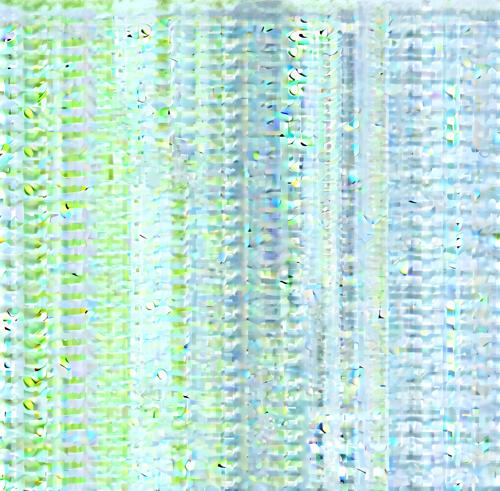}} \\ \\
			95.7         & \raisebox{-60pt}{\includegraphics[width=.3\columnwidth]{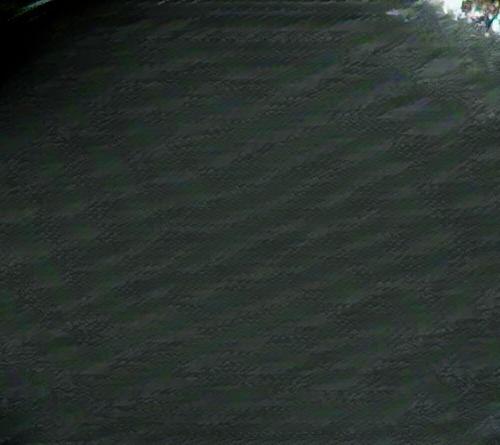}} & \raisebox{-60pt}{\includegraphics[width=.3\columnwidth]{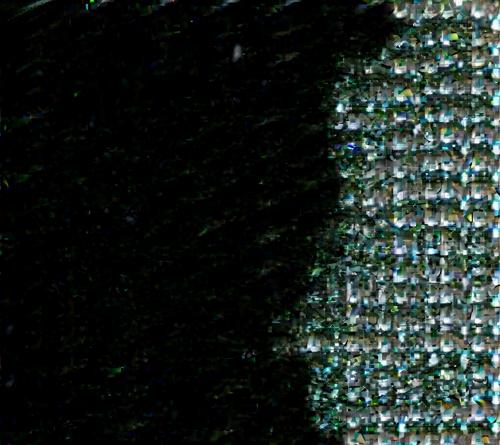}}
		\end{tabular}}
\end{table*}

\begin{figure}
	\centering
	\includegraphics[width=.9\textwidth]{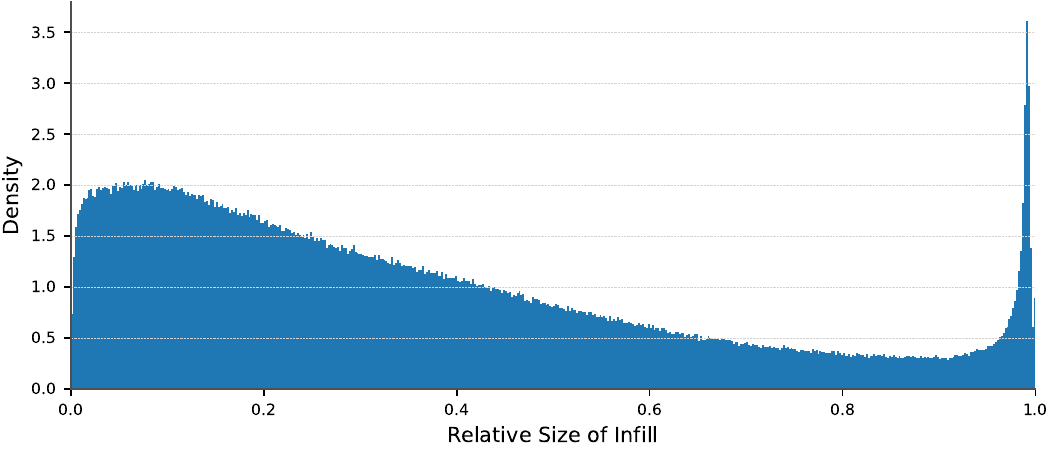}
	\caption{\textbf{Infill-ratio distribution.} We plot the distribution of the relative size of the detected foreground object that is infilled in our Segmentation step of ImageNet.
		While most images contain objects of smaller size, there is a peak where Grounded~SAM~\cite{Ren2024} detects almost the whole image as the foreground object. For examples of such large infills, see \Cref{tbl:high-rat}.
	}
	\label{fig:infill-distr}
\end{figure}

\Cref{tbl:high-rat} shows infills for images where Grounded SAM \cite{Ren2024} marks a high percentile of the image as the foreground object (Infill Ratio), that has to be erased by the infill models.
The examples show that when the infilled region becomes large, both methods begin to lose coherent global structure, with outputs dominated by repetitive or texture‑like patterns.
LaMa tends to produce smoother, more uniform surfaces, like we saw in \Cref{tab:infill-examples}, while Attentive Eraser often generates denser, more regular texture patterns.
Across the rows, increasing infill ratio corresponds to increasingly homogeneous results, with only faint hints of original scene cues remaining.
\Cref{fig:infill-distr} plots the distribution of infill ratios in \schemename.
While there is a smooth curve of the number of detections decreasing with the infill ratio until $\approx 90\%$, there is an additional peak at $\approx 100\%$ infill ratio.
We hypothesize that this peak is made up of failure cases of Grounded~SAM.

We filter out all backgrounds that have an infill ratio larger than our pruning threshold $t_\text{prune} = 0.8$, which translates to $10\%$ of backgrounds.
Per-class rates are not uniform, but no class is removed: $95$\% of classes retain $\geq 70$\% of their BGs, and the worst-affected class (n03841143: odometer) retains $30$\%.
Thus, each class retains a substantial BG-pool.